\documentclass[conference]{IEEEtran}

\usepackage[utf8]{inputenc}
\usepackage[cmex10]{amsmath}
\usepackage{amsthm}
\usepackage{amssymb}
\usepackage{mathrsfs}
\usepackage{graphicx}
\usepackage{float}
\usepackage{array}
\usepackage{epstopdf}
\usepackage{multirow}
\usepackage{cite}
\usepackage{comment}

\begin{document}

\title{Modern Augmented Reality: Applications, Trends, and Future Directions}

\author{Shervin Minaee$^{*}$,
Xiaodan Liang$^{**}$,
Shuicheng Yan$^{\dagger}$  
\\
$^{*}$Snap Inc\\ 
$^{**}$Sun Yat-sen University\\ 
$^{\dagger}$Sea AI Lab\\
}

\maketitle

\begin{abstract}
Augmented reality (AR) is one of the relatively old, yet trending areas in the intersection of computer vision and computer graphics with numerous applications in several areas, from gaming and entertainment, to education and healthcare. 
Although it has been around for nearly fifty years, it has seen a lot of interest by the research community in the recent years, mainly because of the huge success of deep learning models for various computer vision and AR applications, which made creating new generations of AR technologies possible.
This work tries to provide an overview of modern augmented reality, from both application-level and technical perspective. 
We first give an overview of main AR applications, grouped into more than ten categories.
We then give an overview of around 100 recent promising machine learning based works developed for AR systems, such as deep learning works for AR shopping (clothing, makeup), AR based image filters (such as Snapchat's lenses), AR animations, and more.
In the end we discuss about some of the current challenges in AR domain, and the future directions in this area.
\end{abstract}

\IEEEpeerreviewmaketitle

\section{Introduction}
\label{sec:intro}
Augmented reality (AR) is an interactive experience of real-world environments, where the objects of the real world are enhanced by computer-generated perceptual information, sometimes across multiple modalities, including visual, auditory, haptic, and somatosensory \cite{billinghurst2015survey}. It provides an enhanced version of the real physical world.
Augmented reality (AR) and virtual reality (VR) are closely related to each other, but in VR the users' perception of reality is completely based on virtual information.

Despite the huge popularity of AR in recent years, its history goes back to more than 50 years ago. 
Of course early AR applications were very basic, and AR technology has come a long way with a growing list of use cases in recent years. Here we provide a brief history augmented reality systems, from concepts to the new applications. 
In 1968, Ivan Sutherland, a Harvard professor and computer scientist, created the first head-mounted display called ‘The Sword of Damocles’. 
In 1974, a lab dedicated to artificial reality was created at the University of Connecticut, called "Videoplace".
The term "augmented reality" was later coined by Tom Caudell, a Boeing researcher. 
Later in 1992, a researcher (Louis Rosenburg) in the USAF Armstrong's Research Lab, created ‘Virtual Fixtures’, which was one of the first fully functional augmented reality systems. This system allowed military personnel to virtually control and guide machinery to perform tasks like training their US Air Force pilots on safer flying practices.
And in 1994, Julie Martin, a writer and producer, brought augmented reality to the entertainment industry for the first time with the theater production titled Dancing in Cyberspace.
In 1999, NASA created a hybrid synthetic vision system of their X-38 spacecraft. The system leveraged AR technology to assist in providing better navigation during their test flights.

AR systems started to get broader interests and more real-world applications around 2000. 
In 2000, Hirokazu Kato developed an open-source software library called the ARToolKit. This package helps other developers build augmented reality software programs.
In 2003, Sportvision enhanced the 1st \& Ten graphic to include the feature on the new Skycam system, providing viewers with an aerial shot of the field with graphics overlaid on top of it. 
In 2009, Esquire Magazine used augmented reality in print media for the first time in an attempt to make the pages come alive.
In 2013, Volkswagen debuted the MARTA app which primarily gave technicians step-by-step repair instructions within the service manual.
In 2014, Google unveiled its Google Glass devices, a pair of augmented reality glasses that users could wear for immersive experiences.
In 2016, Microsoft started shipping its version of wearable AR technology called the HoloLens.
In 2017, IKEA released its augmented reality app called IKEA Place that was a new experience in the retail industry.
Also during past few years, Snapchat has introduced several AR lenses in their apps, which have made image and video communications much more fun.

Hardware components for AR includes a processor, display, sensors and input devices. Modern mobile computing devices like smartphones and tablet computers contain these elements, making them suitable AR platforms. 
In terms of display, various technologies are used in AR rendering, including optical projection systems, monitors, handheld devices, and display systems, which are worn on the human body. A head-mounted display (HMD) is a display device worn on the forehead, such as a harness or helmet-mounted. 
AR displays can be rendered on devices resembling eyeglasses (such as Google Glass, and Snapchat's new Spectacles).

Some of the popular tools for developing augmented reality related solutions includes: ARKit developed by Apple and used by iOS developers to build mobile AR apps and games for iPhones, iPads, and other Apple devices, ARCore developed by Google and has many features that help integrate AR elements into the real environment, including motion tracking, surface detection, and lighting estimation (supports development in Android, iOS, Unreal, and Unity), SnapML and Lens Studio developed by Snap and used by the lens developers for Snapchat app, echoAR (a cloud platform for augmented reality and virtual reality), Unity, SparkAR, Vuforia, Wikitude, and ARToolKit.

In this work we provide a high level review of modern augmented reality from both application and technical perspectives. We first provide an overview of the main current applications of augmented reality, grouped into more than 10 categories. We then provide an overview of the recent machine learning based algorithms developed for various AR applications (such as clothing, make-up try on, face effects). Most of these works are based on deep learning models. We also mention the popular public benchmarks for each of those tasks, for cases where a public dataset is available.
After that, we provide a detailed section on the main challenges of AR systems, and some of the potential future directions in AR domain, for the young researchers in this area. 
The main AR applications discussed in this paper includes:
\begin{enumerate}
    \item Games
    \item Social Networks and Communications
    \item Education
    \item Healthcare
    \item Shopping
    \item Automotive Industry
    \item Television and Music Industry
    \item Art and Museum Galleries
    \item Constructions
    \item Advertisement and Financial Companies
    \item Other Areas (Archaeology, Industrial Manufacturing, Commerce, Literature, Fitness and Sport Activities, Military, and Human Computer Interaction)
\end{enumerate}

The structure of the rest of this paper is as follows:
In Section \ref{sec:applications}, we review some of the prominent AR applications, grouped into several categories. 
In Section \ref{sec:models}, we provide an overview of the prominent Machine/Deep learning based models developed for AR applications. 
In Section \ref{sec:future}, some of the challenges of the current AR systems, and some of the potential future directions in AR areas are discussed.
In the end, we conclude this paper in Section \ref{sec:conc}.

\section{Current Applications}
\label{sec:applications}
With the rising popularity of augmented reality in recent years, it has been used in more and more new applications everyday, which makes it hard to list all possible AR applications here. 
Instead, we try to cover the main applications of AR in today's world, grouped into several categories. We review their high-level applications in this section, and leave the technical/modeling part of those works for the next section.

\subsection{Games}
Gaming is bigger than it has ever been, driven by the growth of mobile gaming, and now makes up 20-26 percent of all media consumption hours. AR gaming is the integration of visual and audio content of the game with the user's environment in real time. Unlike virtual reality gaming, which usually requires a separate room or confined area to create an immersive environment, augmented reality gaming uses the existing environment and creates a playing field within it, which makes it  simpler for both users and developers. 
An augmented reality game often superimposes a pre-created environment on top of a user’s actual environment. 

\begin{figure}[h]
\centering
\includegraphics[width=0.80\linewidth]{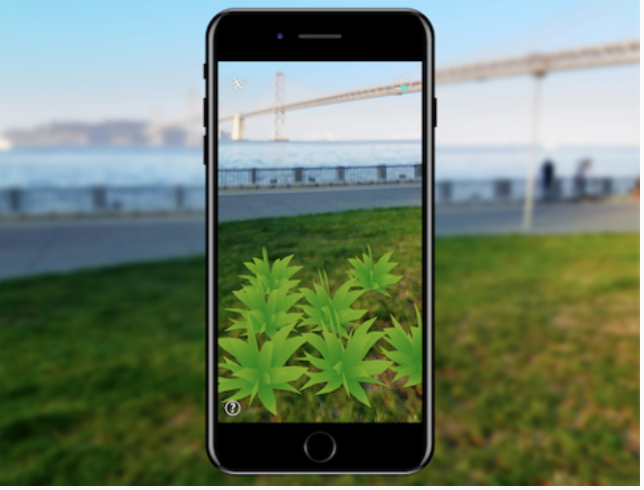}
\caption{A sample snapshot of Pokemon app, for finding the Pokemon from within the tall grass. Courtesy of Pokemon App.}
\label{fig:pokemon}
\end{figure}

Some of the prominent AR gaming apps includes \textbf{Pokémon GO} (which uses a smartphone’s camera, gyroscope, clock and GPS and to enable a location-based augmented reality environment) shown in Fig \ref{fig:pokemon}, \textbf{Jurassic World Alive} (which brings dinosaurs into the real world and players can head out in search of the prehistoric monsters and capture them), \textbf{Harry Potter: Wizards Unite} (that sets players to walk around in the real world and collect various wizarding items, battle with foes and deal with a calamity that has hit wizards and witches across the world), \textbf{The Walking Dead: Our World} (that the undead zombies from the popular television series out of the TV screen and into our surrounding environment).

\subsection{Social Networks and Communications}
\begin{figure}[h]
\centering
\includegraphics[width=0.80\linewidth]{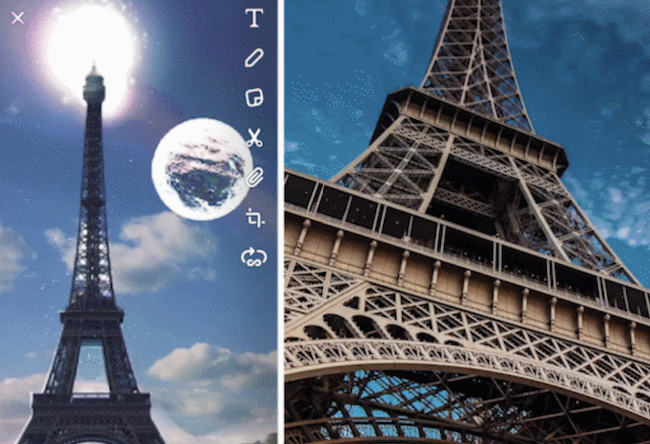}
\includegraphics[width=0.80\linewidth]{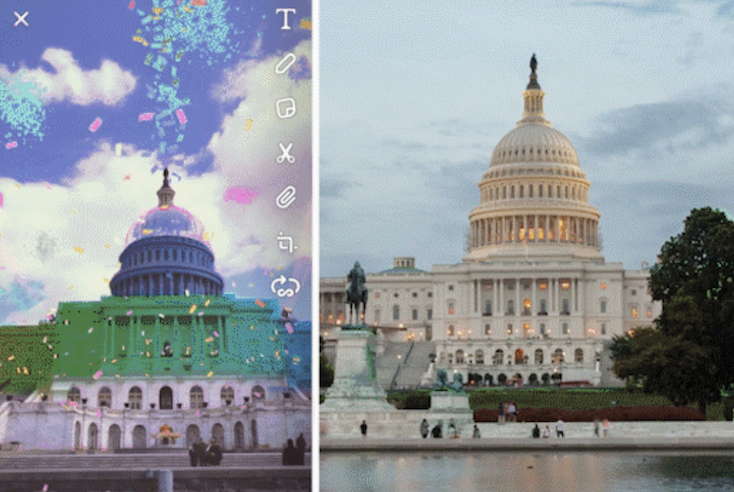}
\caption{The AR effect made by Snapchat on Eiffel Tower, and White House. Courtesy of TheVerge \cite{verge}.}
\label{fig:snapchat}
\end{figure}

Augmented Reality is one of the trending additions to the social networks and communication applications features. AR can make communications with friends and celebrities more entertaining. 
As an example, Snapchat provides various AR lenses for people, from simply adding hats/horns/eyeglasses to making popular landmarks move (some examples shown in Fig \ref{fig:snapchat})

The AR effects are also used in image and video communication tools,  such as Zoom, Microsoft Teams, and Google Meet (with the help of SnapCamera), in which people can augment their videos during a meeting by applying various AR effect.

\subsection{Education}
Augmented reality is great material for education and learning/training platforms. It  can be used to make the education and training platforms more engaging and fun. Children often enjoy learning new experiences and technology, so AR can motivate students to learn and make the classes more entertaining and engaging.
AR based platforms for education have been in huge demand after the COVID-19 pandemic, which shifted most of the education systems to the remote phase. 
Since AR has become more accessible and affordable, instead of buying physical supplies, AR may be more cost-effective for schools in the future.

As an example, in 4D Anatomy \cite{4danatomy}, students can explore more than 2,000 anatomical structures and discover 20 different dissection specimens of real anatomy. They can improve understanding anatomy by manipulating and observing virtual 3D objects from different angles.

\subsection{Healthcare}
Medical and healthcare industry are another place which augmented reality can be very effective and useful.
AR is already used in simulations for surgeries and diseases to enhance patient treatments. It is also used in education for patient and doctor. But their potential scope could go well beyond these.

One prominent AR based solution in healthcare is AccuVein \cite{accuvein}, which uses projection-based AR that scans a patient’s body and shows doctors the exact location of veins. This leads to improvement in the injection of the vein on the first stick by 3.5 times and reduces the need to call for assistance by 45\%.

\subsection{Shopping}
Shopping is perhaps one of the main areas in which AR can have a huge impact.
With the advent of e-commerce, some retail stores have already adopted the newest of AR technologies to enhance the customer’s shopping experience to get an edge over other stores. They have transformed the whole experience of a shopper from entering a store to opening the final product at home in unimaginable ways. 

AR application in shopping is very broad, from virtual clothing try-on (either on the app, or using the in-store magic mirror), virtual makeup try-on, to virtual in-store navigation. Some of the AI and machine learning applications in this space, includes techniques for clothing/make-up trying, object understanding, human parsing, object segmentation, size estimation, scene understanding, and many more.

\begin{figure}[h]
\centering
\includegraphics[width=0.9\linewidth]{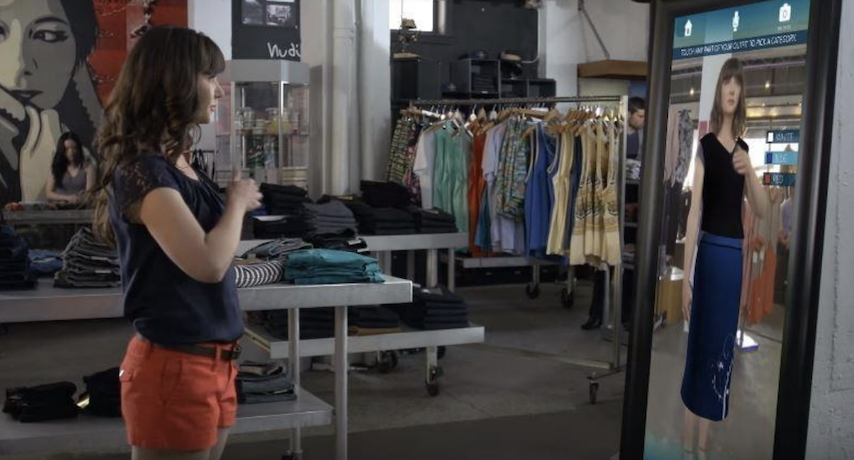}
\caption{A woman tries on virtual garments using virtual mirror in-store. Courtesy of \cite{loh2020virtual}.}
\label{fig:magic_mirror}
\end{figure}

\begin{figure}[h]
\centering
\includegraphics[width=0.7\linewidth]{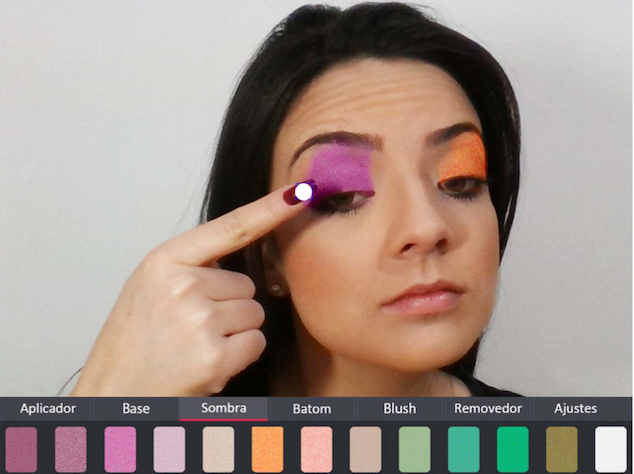}
\caption{User experience with a sample virtual makeup augmented reality system. Courtesy of \cite{borges2019virtual}.}
\label{fig:virtual_makeup}
\end{figure}

Some of the popular apps which are using AR for shopping includes, Home Depot (which expanded the functionality in its main mobile app to allow users to overlay Home Depot merchandise on any room in their home), IKEA place (this app take a picture of your living room, and automatically measures the space, then it provides recommendations on furniture that fits in the space), Wayfair, Target, Sephora (this app contains a Virtual Artist, that uses facial recognition technology to allow customers to digitally try on products), Nike (the app has Nike Fit feature that allows customers to find their true shoe size), Warby Parker (its app  allows customers to digitally try on glasses from the comfort of the customer’s home), Amazon, and many more.

\subsection{Automotive Industry}
Augmented reality in the automotive industry is in high demand, and car manufacturers are planning to incorporate AR into cars in near future. AR in automobiles is expected to have a value of more than \$600 billion by 2025.

As a prominent example, Nissan has developed Invisible-to-Visible (I2V) solution using AR and AI \cite{nissan}, which makes drivers aware of potential hazards like nearby objects, and redirect drivers’ focus to the road if they are not concentrated. AR can be very helpful for safety because it may decrease the number of accidents and drivers can drive comfortably.

\subsection{Television and Music Industry}
Another big AR applications are in TV and Music industries. It can help the producer to enrich their contents by providing more information about the show, program, music, and creators.

AR has  already been used in various TV programs for while. As an example, when you’re watching a show on the television, you may receive additional information about it, e.g. for a baseball match you receive match scores, player information, and related information. There could also be some pointers showing the position of some objects in sport games, such as balls, or players.

Also music has been transforming a lot recently, and music is more than just listening to some favorite tracks put together in playlists. AR can help us grab information like the artist bio, cover up videos, dance videos on the track and so much more. It can help us enhance live performance streaming events by telling us a story which couldn’t have been possible without AR.

\subsection{Art and Museum Galleries}
AR can make seeing an artwork or a museum much easier for people, and help people overcome location/distance barrier.  

\begin{figure}[h]
\centering
\includegraphics[width=0.9\linewidth]{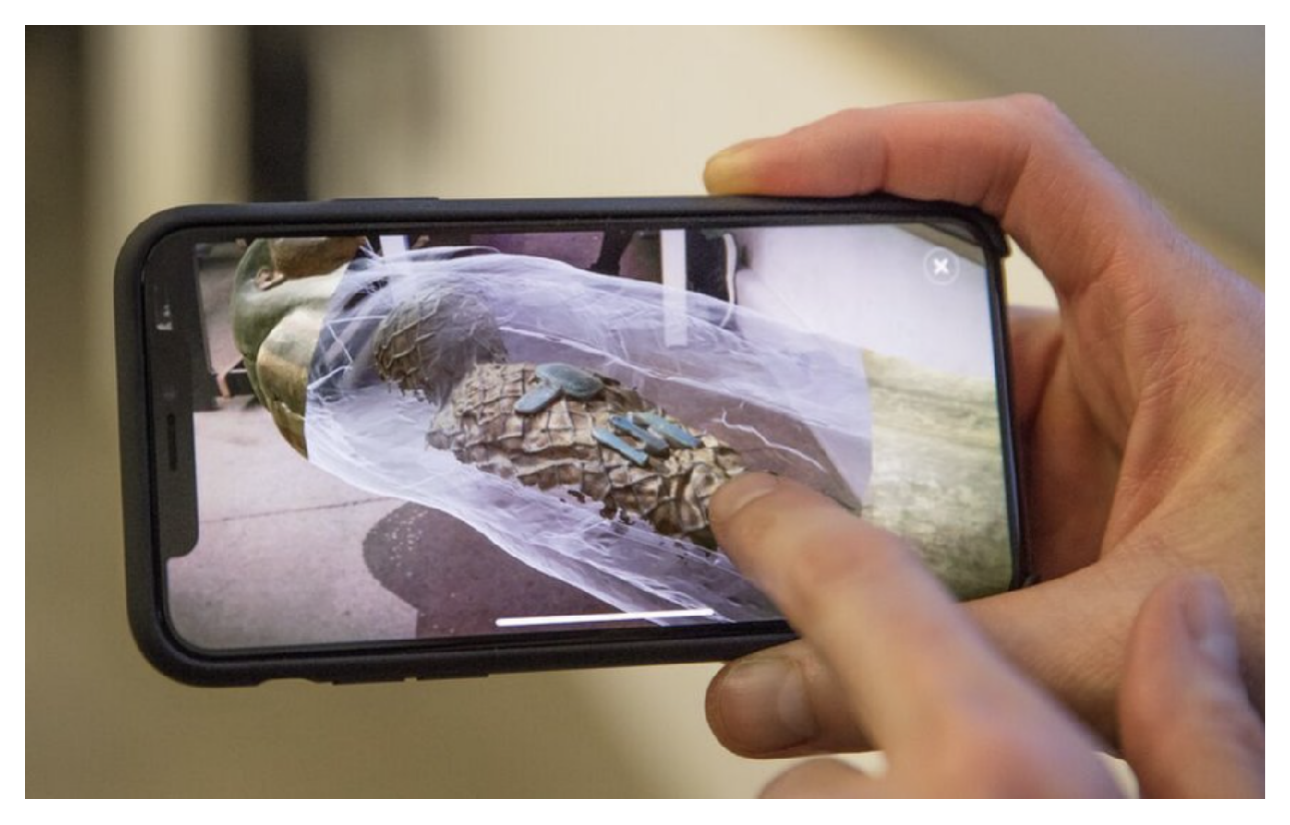}
\caption{An example of AR experience with Civilizations app. Courtesy of \cite{civilization}.}
\label{fig:Civilizations}
\end{figure}

We’re seeing more art galleries incorporating AR experiences. In December 2017, the first ever AR-powered art exhibition by Perez Art Museum Miami (PAMM), was released. 
Another popular app along this use case is the Civilizations app by BBC. The app creators gathered more than 280 artifacts from famous museums and galleries and turned them into 3D models. This app allows exploring artifacts in exhibitions and learning their history and specific details.
One example is shown in Fig \ref{fig:Civilizations}.

In addition to the museums, many artists have come ahead with AR mobile apps that let users around the world view their artwork the way it is meant to be seen. 
This can help artists to better promote their artworks, and make it accessible to more people around the world.

\subsection{Constructions}
AR has several applications in construction areas, and has been already used by many of the biggest construction firms around the world. 
Its applications ranges from simple use cases such as safety training of the works, to more advanced use-cases such as team collaboration, real-time project information, and project planning and building modeling. 
With the help of AR technology an empty shell of a building floor can come to life with the location, style and size of windows and doors, pipes, and HVAC systems. Using an AR headset, the worker sees these details as if they were right in front of them; they can compare what they see to the building plan to ensure everything is in order. 

AR can also be used to showcase 3D models and even provide tours, giving clients a solid idea of what a building would look like before it’s built. If an owner wants to show the client what a new installation would look like on-site, AR can also bring that vision to life.

\subsection{Advertisement and Financial Companies}
Since advertisement remains one of the biggest source of revenue for tech companies, many of them are using AR to produce a more engaging  and informative ad, to lure customer to buy different products. AR allows brands to interact with your customers by giving them a ‘free taste’ of the product before making a purchase.

Augmented reality trends in banking aim to help consumers keep track of their finances better. AR in banking offers a rich visualization of their data and other services. As an example, Wells Fargo designed and built an AR system for consumers to interact with bank tellers within a virtual space placed over reality. Moreover, it comes with gamification like AR games and puzzles.

In addition to the financial companies, insurance companies are also adopting AR. Through the use of AR, insurance companies can better communicate and explain their service to their customers, and help them.
As an example, Allianz uses AR to make its customers aware of possible dangers within their homes. Using their smartphone, they can see such hazards. These range from an overheating toaster to crashed upstairs bathroom floors due to sink flooding, and much more.

\subsection{Other Areas}
It is obvious that the AR applications are not limited to the above items, and AR can be useful in many more areas. 
In addition to the areas listed above, AR has applications in Archaeology (to augment archaeological features onto the modern landscape), industrial manufacturing, Commerce, Literature (as an example AR was blended with poetry by ni ka from Sekai Camera in Tokyo, Japan), fitness and sport activities, and human computer interaction.

\section{Popular Deep Learning Based Models}
\label{sec:models}

\begin{figure*}[h]
\centering
\includegraphics[width=0.95\linewidth]{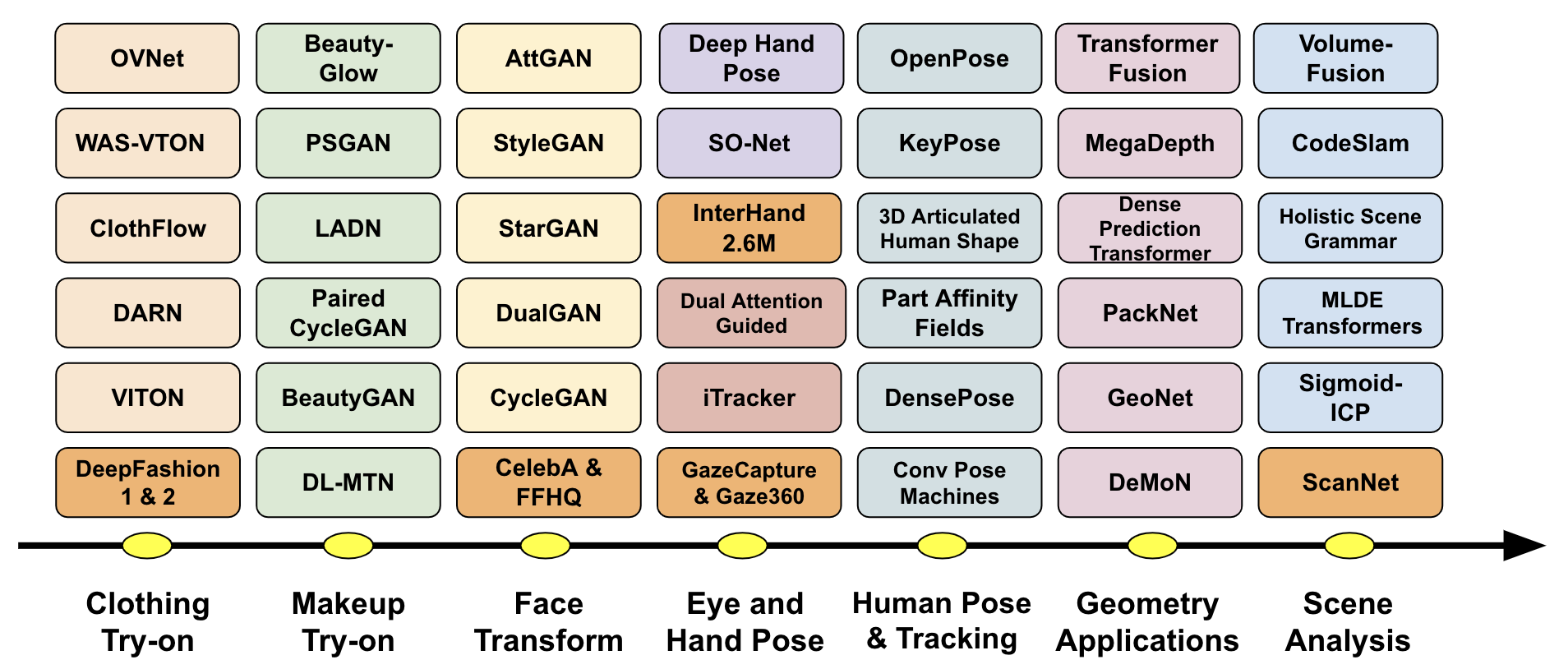}
\caption{An illustration of some of the most representative deep learning works related to various AR applications. Those blocks shown in dark orange refer to potential dataset to use for the corresponding task.}
\label{fig:representatives}
\end{figure*}

In this section, we are going to review some of the recent prominent machine/deep learning algorithms developed for various AR applications. 
Many of the deep learning based models for AR applications are focused on: 
\begin{itemize}
    \item AR for Shopping (clothing try-on, makeup try-on)
    \item AR for Face/Body Transformations
    \item Tracking and Pose Estimation for AR Applications
    \item AR for 3D human Reconstruction 
    \item Geometry Applications
    \item Scene Understanding and Reconstruction
\end{itemize}


Fig \ref{fig:representatives} provides an illustration of a subset of the representative works, which are going to be discussed in the following sections.

\subsection{Models for Clothing Shopping and Try On}
In this section we will provide an overview of some of the recent works for clothing retrieval and try-on. We first cover some of the prominent works for clothing retrieval/matching, and then discuss about the models developed for clothing try-on.

\begin{figure}[h]
\centering
\includegraphics[width=0.9\linewidth]{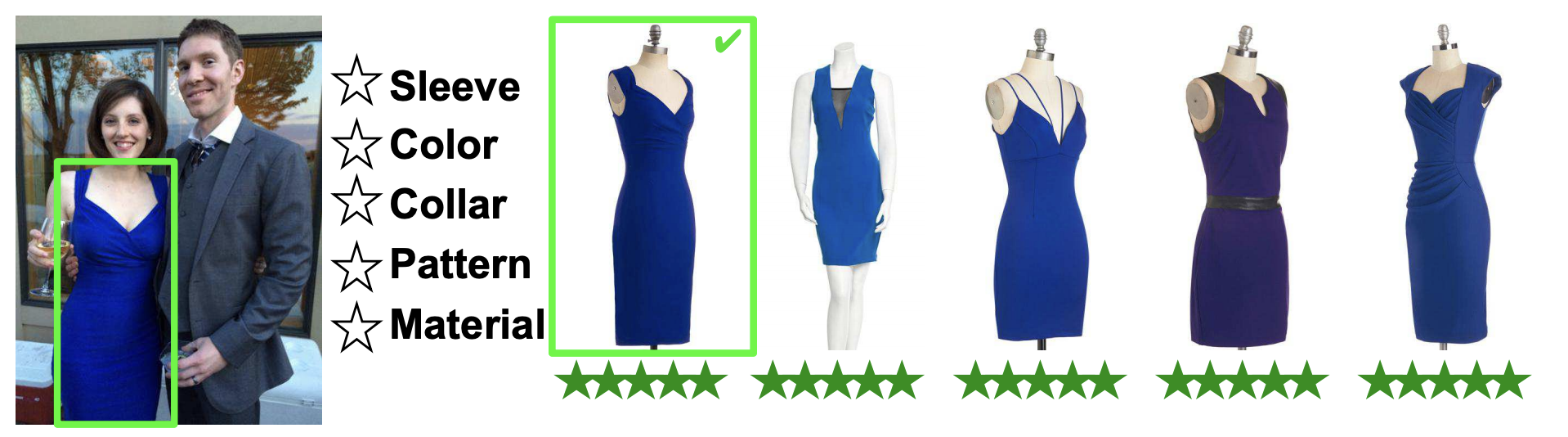}
\caption{An example of clothing matching. Courtesy of \cite{hadi2015buy}.}
\label{fig:clothing_matching}
\end{figure}

Matching a real-world clothing/garment to the same item in an online shopping website could be the first step in finding a desired garment (one example shown in Fig \ref{fig:clothing_matching}.
This is an extremely challenging task due to visual differences between
street photos (pictures of people wearing clothing in everyday uncontrolled settings) and online shop photos (pictures of clothing items on models, mannequins, or in isolation, captured by professionals in more controlled settings.

\begin{figure}[h]
\centering
\includegraphics[width=0.9\linewidth]{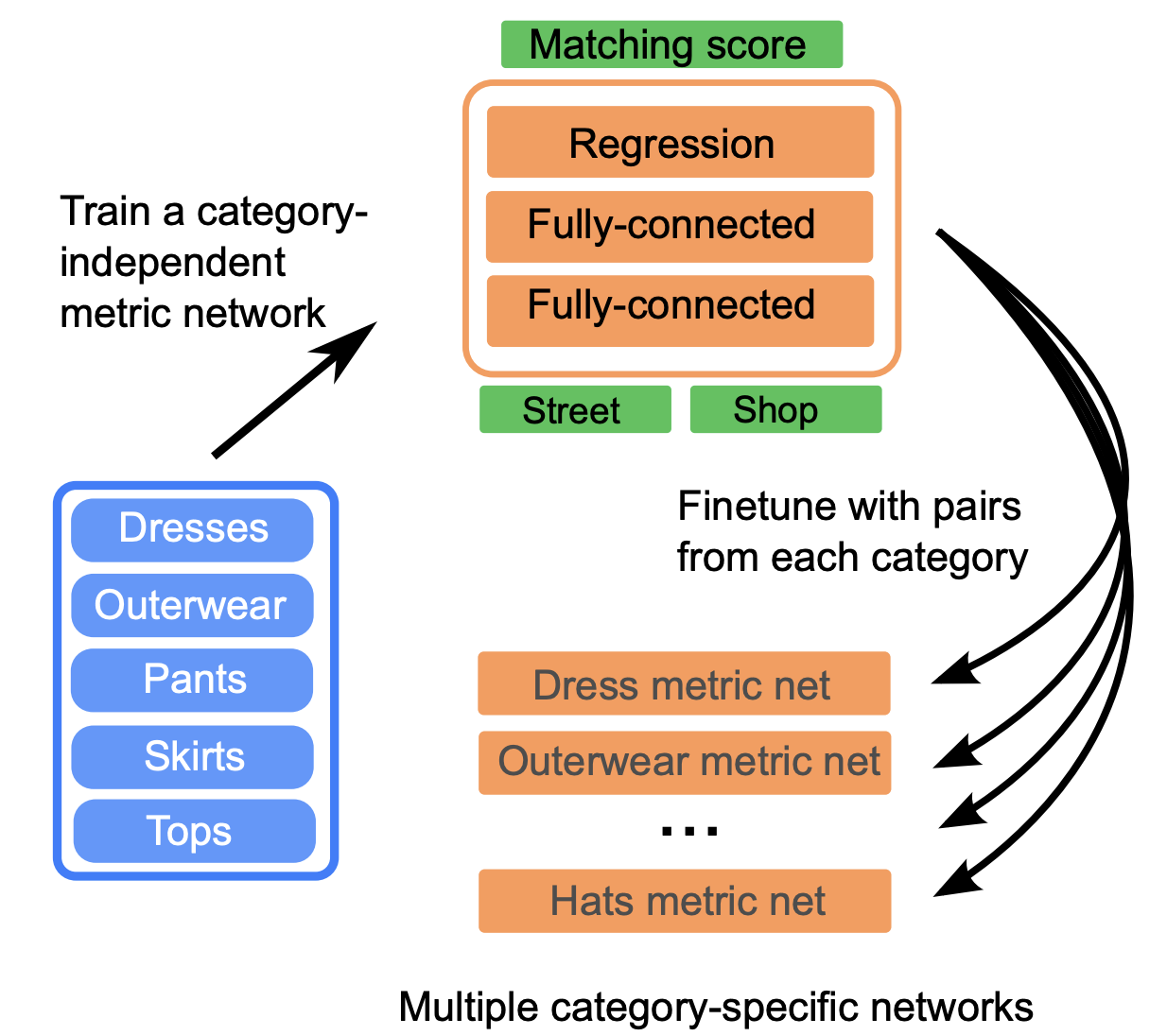}
\caption{The overview of the training, followed by fine-tuning procedure for training category-specific similarity for each category in \cite{hadi2015buy}. Courtesy of \cite{hadi2015buy}.}
\label{fig:clothing_kiapour}
\end{figure}

In \cite{hadi2015buy}, Kiapour et al. collected a dataset for this application containing 404,683 shop photos collected from 25 different online retailers and 20,357 street photos, providing a total of
39,479 clothing item matches between street and shop photos, and developed three different methods for Exact Street to
Shop retrieval, including two deep learning baseline methods, and a method to learn a similarity measure between
the street and shop domains. The overview of their proposed model is shown in Fig \ref{fig:clothing_kiapour}.

\begin{figure}[h]
\centering
\includegraphics[width=0.8\linewidth]{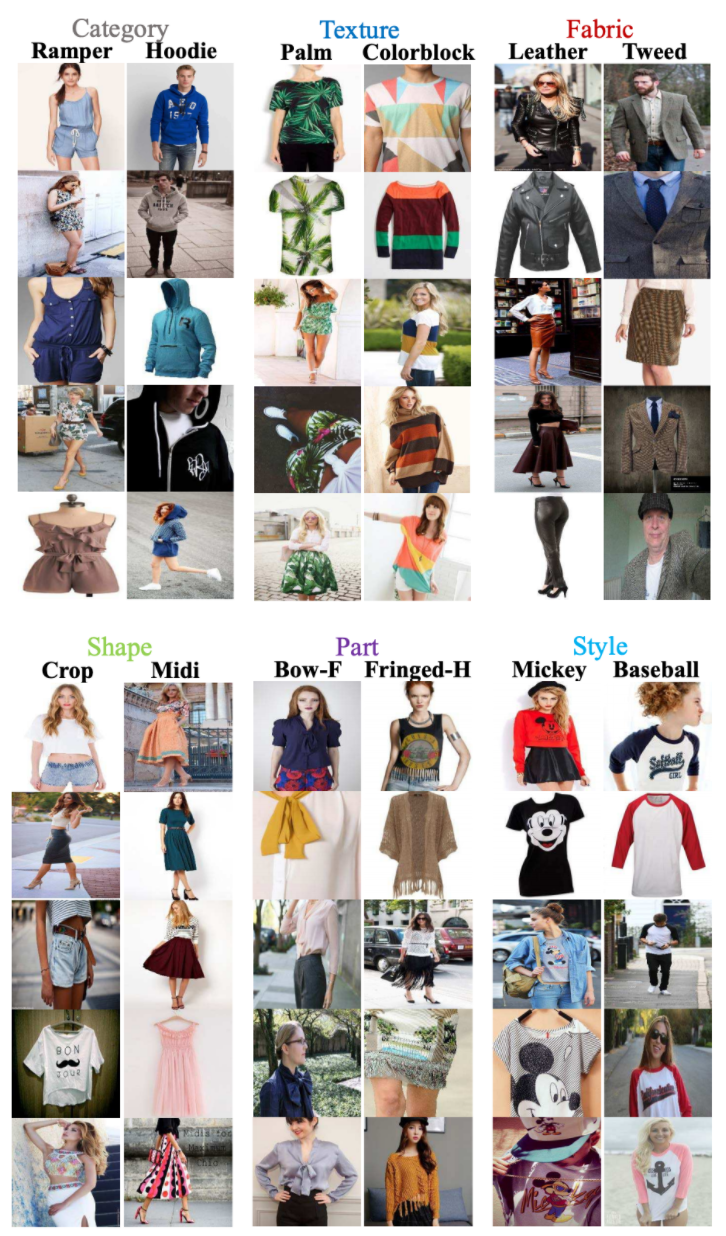}
\caption{Example images of different categories and attributes in DeepFashion. The attributes form five groups: texture, fabric, shape, part, and style. Courtesy of \cite{liu2016deepfashion}.}
\label{fig:DeepFashion1}
\end{figure}

\begin{figure}[h]
\centering
\includegraphics[width=0.9\linewidth]{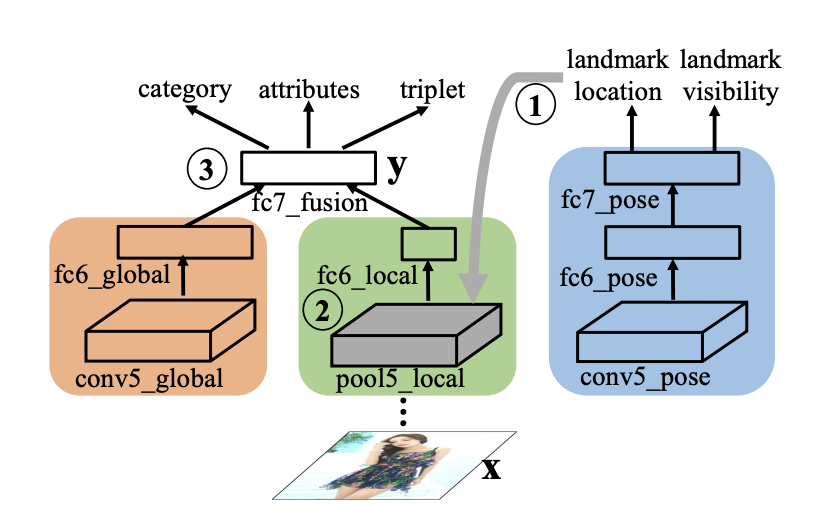}
\caption{Pipeline of FashionNet, which consists of global appearance branch (in orange), local appearance branch (in green) and pose branch (in blue). Shared convolution layers are omitted for clarity. Courtesy of \cite{liu2016deepfashion}.}
\label{fig:fashinNet}
\end{figure}

In \cite{liu2016deepfashion}, Liu et al. introduced DeepFashion1, a large-scale clothes dataset with comprehensive annotations. It contains over 800,000 images, which are richly annotated with massive attributes, clothing landmarks, and correspondence of images taken under different scenarios including store, street snapshot, and consumer. 
Fig \ref{fig:DeepFashion1}, shows some of the sample images from this dataset.
To demonstrate the advantages of DeepFashion, they proposed a
new deep model, namely FashionNet, which learns clothing
features by jointly predicting clothing attributes and landmarks. FashionNet pipeline is shown in Fig \ref{fig:fashinNet}.

\begin{figure}[h]
\centering
\includegraphics[width=0.9\linewidth]{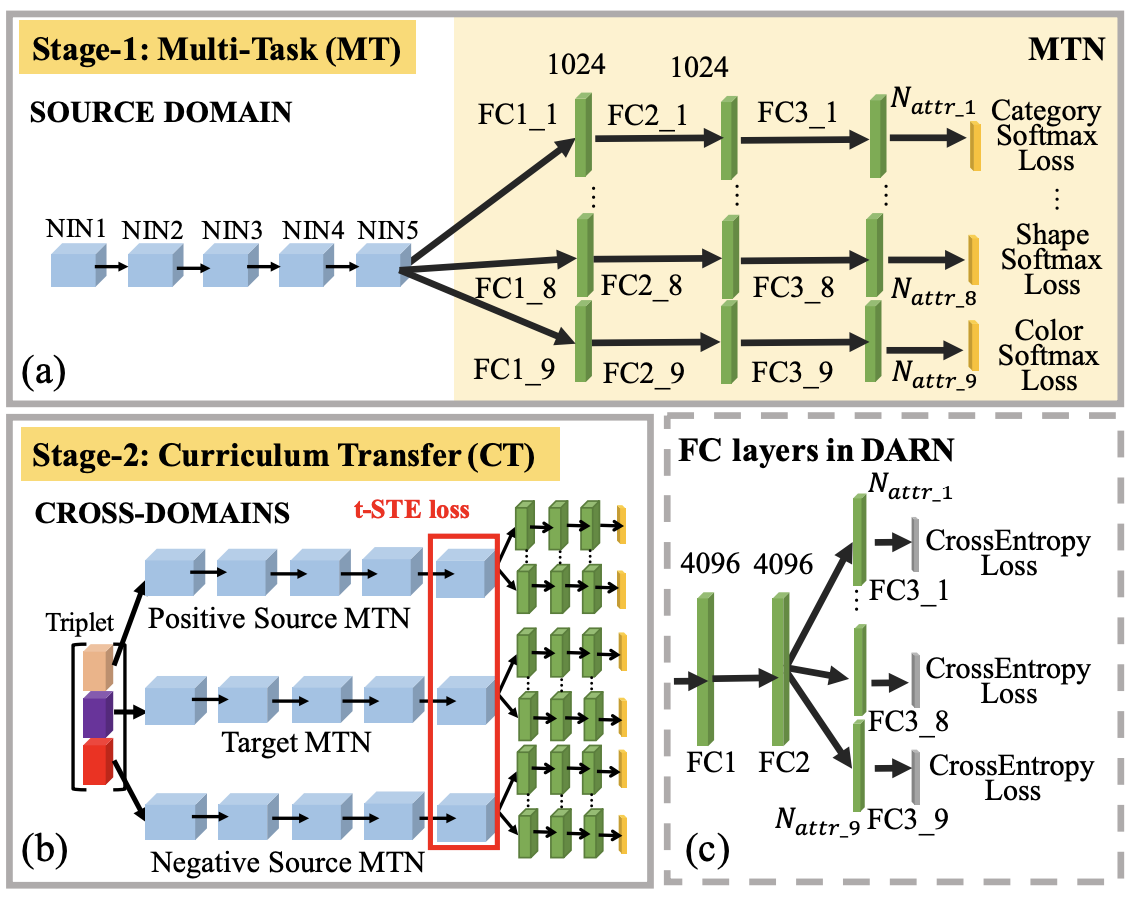}
\caption{(a),(b) show the MTCT network design. (c) illustrates the FC layers of DARN. Courtesy of \cite{dong2017multi}}
\label{fig:MTCT}
\end{figure}

In \cite{dong2017multi}, Dong et al. developed a deep learning framework capable of model transfer learning from well-controlled shop clothing images collected from web retailers to in-the-wild images from the street. They formulated a novel MultiTask Curriculum Transfer (MTCT) deep learning method to explore multiple sources of different types of web annotations with multi-labelled fine-grained attributes.
The architecture of the proposed framework is shown in Fig 
\ref{fig:MTCT}.

\begin{figure}[h]
\centering
\includegraphics[width=0.9\linewidth]{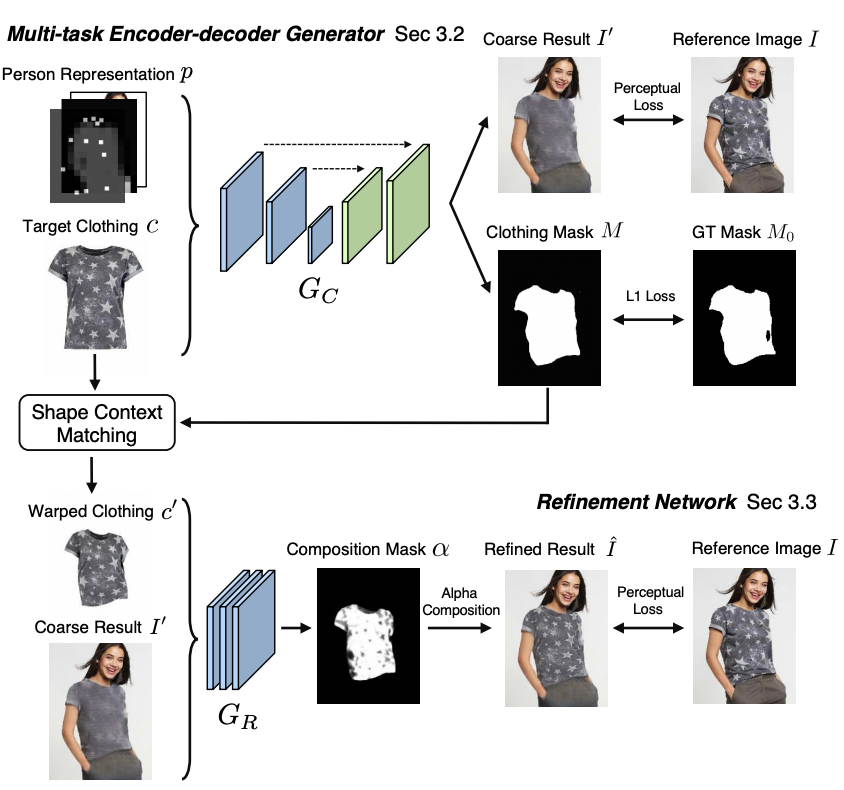}
\caption{An overview of VITON. VITON consists of two
stages: (a) an encoder-decoder generator stage,
and (b) a refinement stage. Courtesy of \cite{han2018viton}}
\label{fig:viton_arch}
\end{figure}

In \cite{han2018viton}, Han et al. proposed an image-based Virtual Try-On Network (VITON) without using 3D information in any form, which seamlessly transfers a desired clothing item onto the corresponding region of a person using a coarse-to-fine strategy.
Conditioned upon a new clothing-agnostic yet descriptive person representation, this framework first generates a
coarse synthesized image with the target clothing item overlaid on that same person in the same pose. 
They further enhance the initial blurry clothing area with a refinement
network. This network is trained to learn how much detail
to utilize from the target clothing item, and where to apply
to the person in order to synthesize a photo-realistic image
in which the target item deforms naturally with clear visual patterns.
The architecture of this framework is shown in Fig \ref{fig:viton_arch}.

\begin{figure}[h]
\centering
\includegraphics[width=0.9\linewidth]{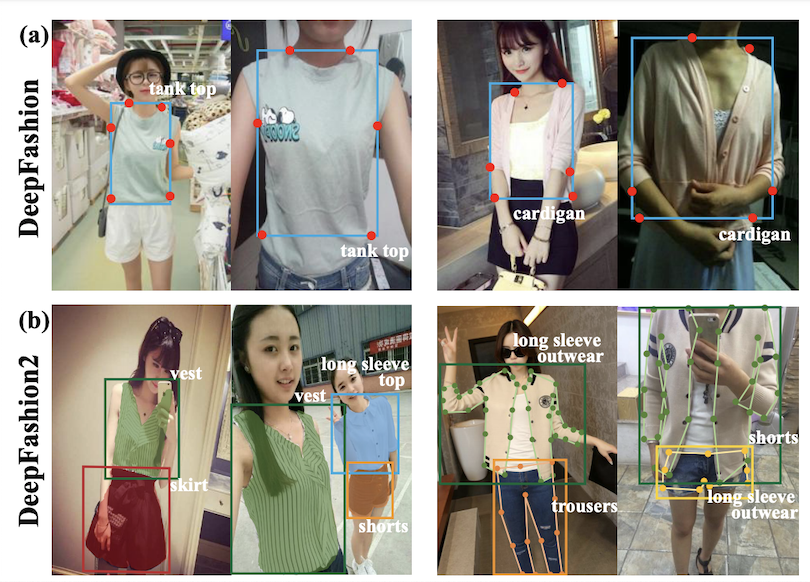}
\caption{Comparisons between (a) DeepFashion and (b) DeepFashion2. (a) only has single item per image, which is annotated with 4-8 sparse landmarks. The bounding boxes are estimated from the labeled landmarks, making them noisy. In (b), each image has minimum single item while maximum 7 items. Each item is manually labeled with bounding box, mask, dense landmarks
(20 per item on average), and commercial-customer image pairs. Courtesy of \cite{ge2019deepfashion2}}
\label{fig:DeepFashion2}
\end{figure}

In \cite{ge2019deepfashion2}, Ge et al. presented DeepFashion2, which is new dataset for clothing. It is a versatile benchmark of four tasks including clothes detection, pose estimation, segmentation, and retrieval. It has 801K clothing items where
each item has rich annotations such as style, scale, viewpoint, occlusion, bounding box, dense landmarks (e.g. 39
for ‘long sleeve outwear’ and 15 for ‘vest’), and masks.
There are also 873K Commercial-Consumer clothes pairs
The comparison between DeepFashion and DeepFashion2 for some images is shown in Fig \ref{fig:DeepFashion2}. 

\begin{figure}[h]
\centering
\includegraphics[width=0.9\linewidth]{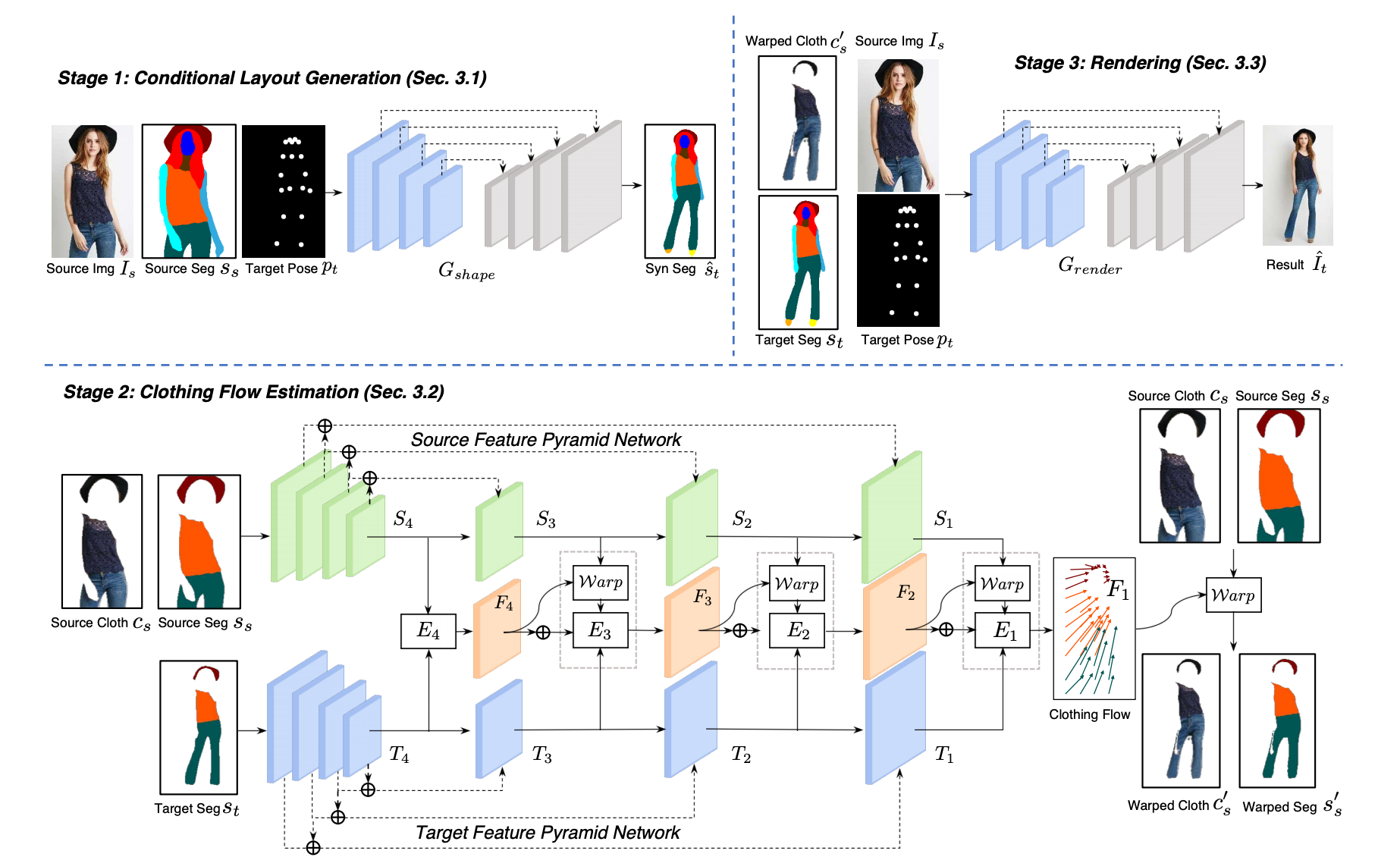}
\caption{Overview of the proposed ClothFlow architecture. Courtesy of \cite{han2019clothflow}}
\label{fig:ClothFlow}
\end{figure}

In  \cite{han2019clothflow}, Han et al. presented ClothFlow, an appearance-flow-based generative model to synthesize clothed persons for pose-guided person image generation and virtual try-on. By estimating a dense flow between source and target clothing regions, ClothFlow effectively models the geometric changes
and naturally transfers the appearance to synthesize novel
images.
They achieve this with a three stage framework: 1) Conditioned on a target pose, they first estimate a person semantic layout to provide richer guidance to the generation process. 2) Built on two feature pyramid networks, a cascaded flow estimation network then accurately estimates the appearance matching between corresponding clothing regions. The resulting dense flow warps
the source image to flexibly account for deformations. 3)
Finally, a generative network takes the warped clothing regions as inputs and renders the target view. 
The architecture of ClothFlow framework is shown in Fig \ref{fig:ClothFlow}.

\begin{figure}[h]
\centering
\includegraphics[width=0.8\linewidth]{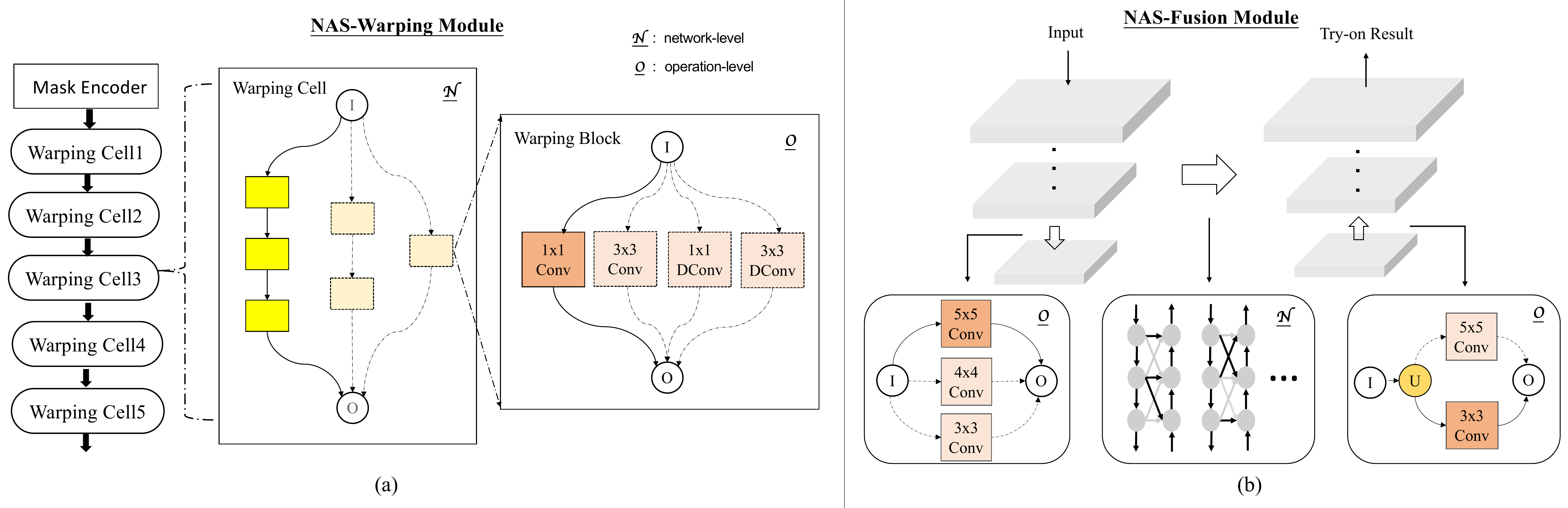}
\caption{The search space for the NAS-Warping Module and NAS-Fusion Module of WAS-VTON. Courtesy of \cite{xie2021wasvton}.}
\label{fig:WAS-VTON-search-space}
\end{figure}

\begin{figure}[h]
\centering
\includegraphics[width=0.8\linewidth]{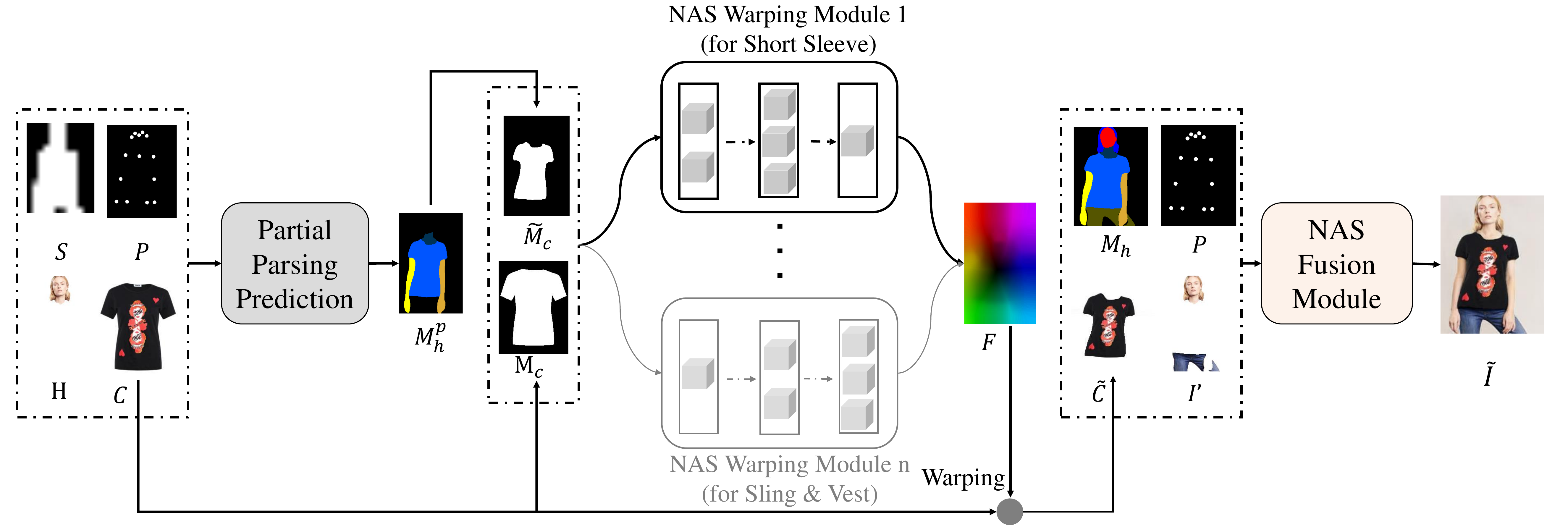}
\caption{The overall framework of WAS-VTON. Courtesy of \cite{xie2021wasvton}.}
\label{fig:WAS-VTON-framework}
\end{figure}

In \cite{xie2021wasvton}, Xie et al. proposed WAS-VTON that employs the Neural Architecture Search (NAS) to explore the garment-category-specific warping network and the optimal garment-person fusion network for the virtual try-on task. To meet this end, WAS-VTON introduces NAS-Warping Module and NAS-Fusion Module, each of which is composed of a network-level (i.e., with different network architecture) and an operation-level (i.e., with different convolution operations) search space. Specifically, the search space of NAS-Warping Module covers various sub-networks with different warping ability which is defined by the number of warping blocks within each warping cell, while the search space of NAS-Fusion Module consists of various sub-networks with skip connections between different scale features. Furthermore, to support two searchable modules, WAS-VTON introduces Partial Parsing Prediction to estimate the semantic labels of the replaced region in the try-on result.
Finally, WAS-VTON applies the one-shot framework in \cite{guo2020single} to separately search the category-specific network for garment warping, and search the optimal network with particular skip connection for garment-person fusion.
The search space of each module and the overall framework of WAS-VTON are shown in Fig \ref{fig:WAS-VTON-search-space} and Fig \ref{fig:WAS-VTON-framework}, respectively.

\begin{figure}[h]
\centering
\includegraphics[width=0.8\linewidth]{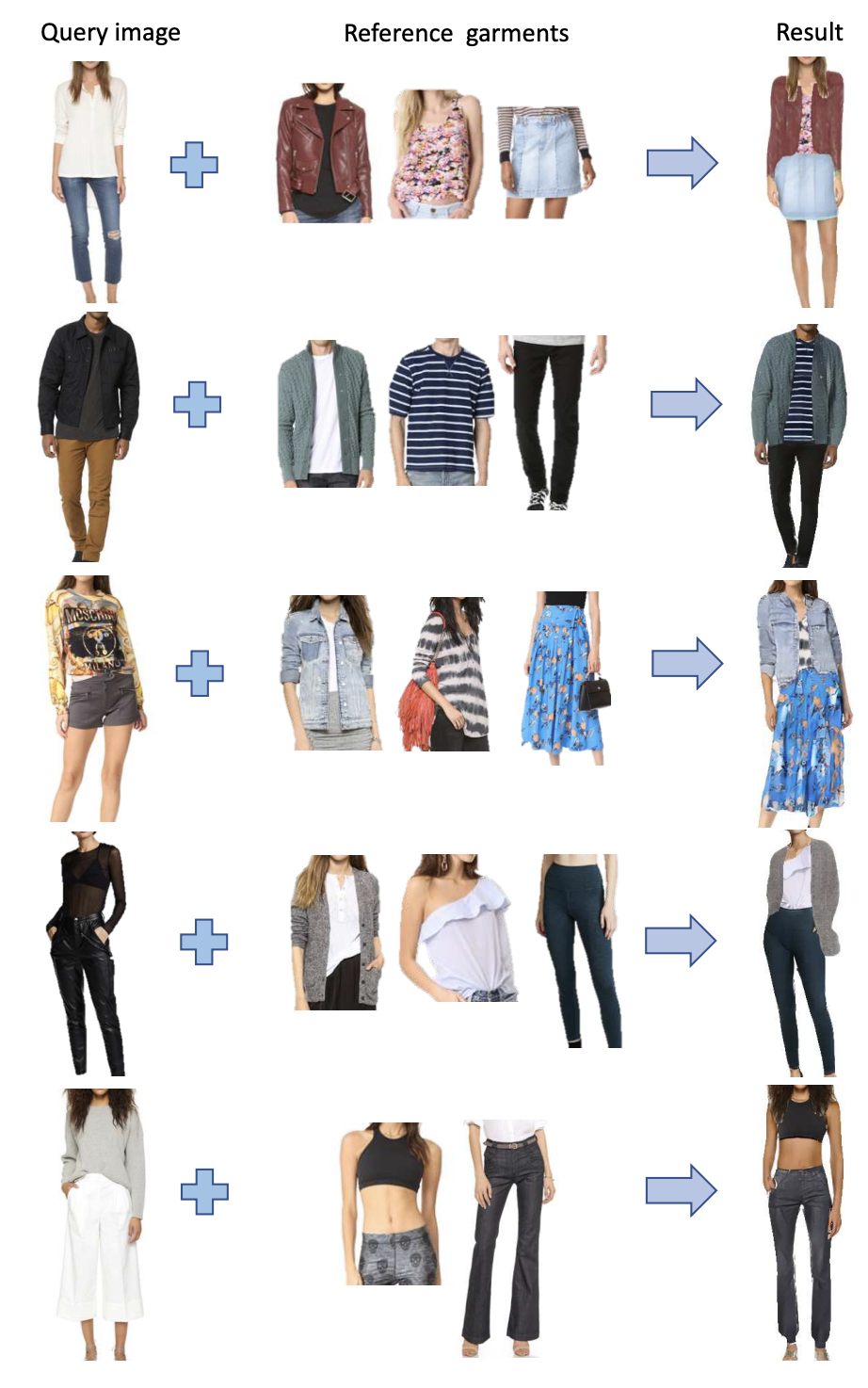}
\caption{The O-VITON algorithm is designed to synthesize images that show how a person in a query image is expected to look with garments selected from multiple reference images. Courtesy of \cite{yang2020towards}}
\label{fig:O-VITON}
\end{figure}

\begin{figure*}[h]
\centering
\includegraphics[width=0.9\linewidth]{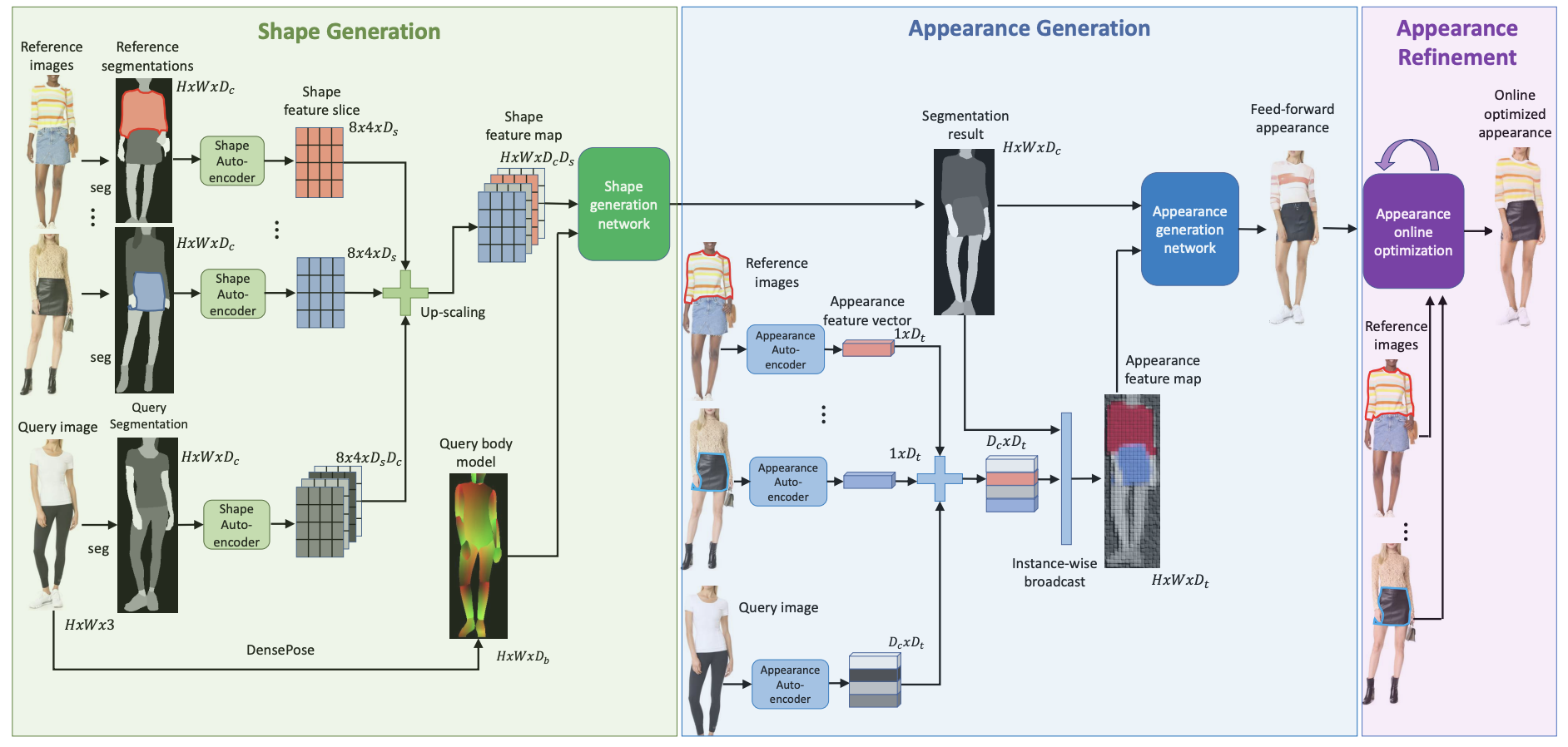}
\caption{The model architecture of O-VITON framework. Courtesy of \cite{yang2020towards}}
\label{fig:O-VITON-arch}
\end{figure*}

In \cite{neuberger2020image}, Neuberger et al. presented a new image-based virtual try-on
approach (Outfit-VITON) that helps visualize how a composition of clothing items selected from various reference
images form a cohesive outfit on a person in a query image.
Fig \ref{fig:O-VITON} illustrates the high-level ideal of this framework. 
The O-VITON framework has three main steps. The first shape generation step generates a new segmentation map representing the
combined shape of the human body in the query image and the shape feature map of the selected garments, using a shape auto-encoder.
The second appearance generation step feed-forwards an appearance feature map together with the segmentation result to generate an a
photo-realistic outfit. An online optimization step then refines the appearance of this output to create the final outfit.
This is shown in Fig \ref{fig:O-VITON-arch}.

\begin{figure}[h]
\centering
\includegraphics[width=0.9\linewidth]{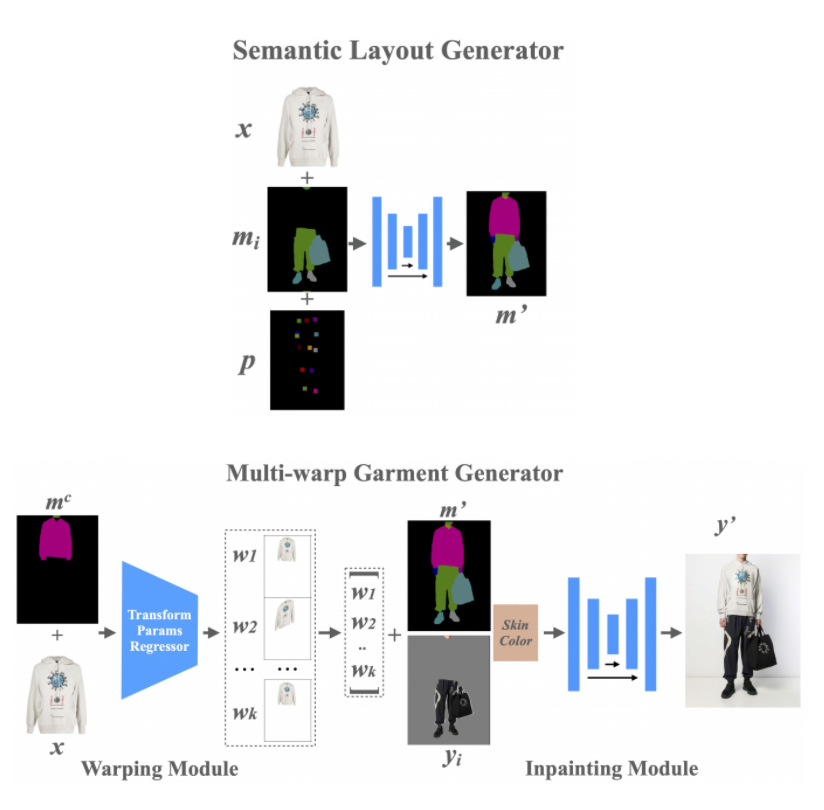}
\caption{The model architecture of Outfit Visualization Net. Courtesy of \cite{li2021toward}}
\label{fig:OVNet}
\end{figure}

In \cite{li2021toward}, Li et al. proposed Outfit
Visualization Net (OVNet) to capture these important details (e.g. buttons, shading, textures, realistic hemlines, and interactions between garments) and produce high quality multiple-garment virtual try-on images. OVNet consists of
1) a semantic layout generator and 2) an image generation pipeline using multiple coordinated warps. We train
the warper to output multiple warps using a cascade loss,
which refines each successive warp to focus on poorly generated regions of a previous warp and yields consistent improvements in detail. In addition, they introduce a method
for matching outfits with the most suitable model and produce significant improvements for both our and other previous try-on methods.
The high-level architecture of OVNet is shown in Fig \ref{fig:OVNet}.

\begin{figure}[h]
\centering
\includegraphics[width=1.0\linewidth]{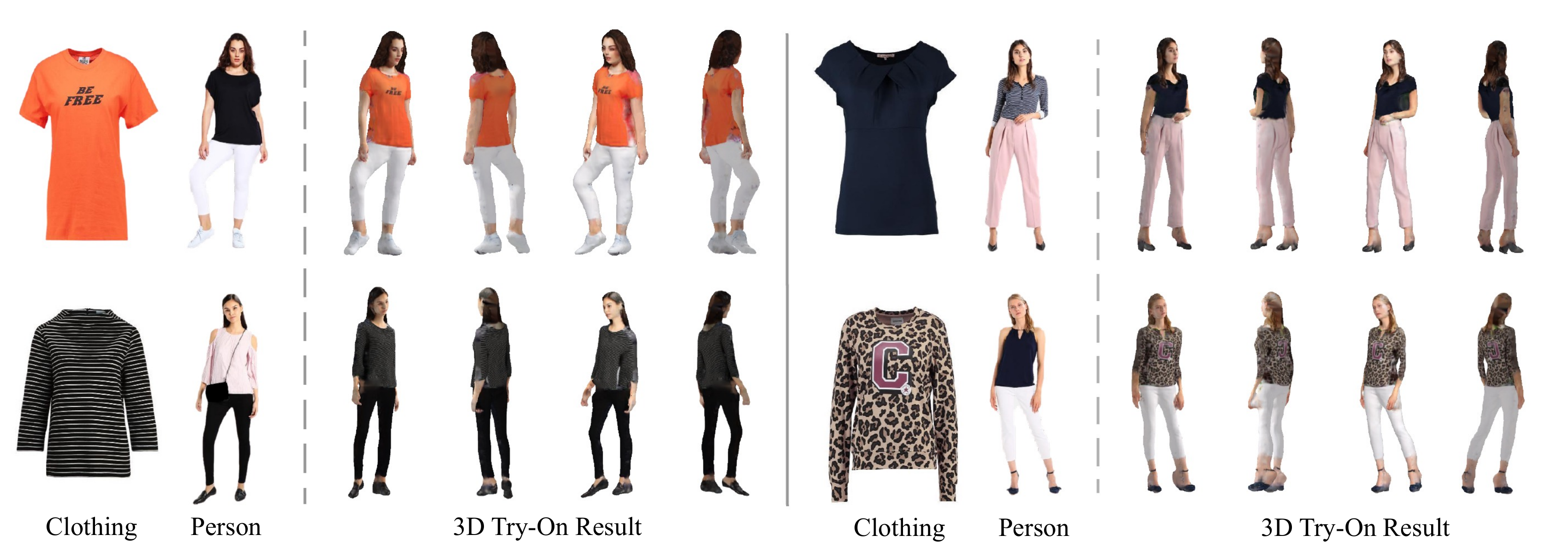}
\caption{3D try-on results from M3D-VTON. Courtesy of \cite{Zhao2021M3DVTONAM}}
\label{fig:m3d-vton_teaser}
\end{figure}

\begin{figure}[h]
\centering
\includegraphics[width=1.0\linewidth]{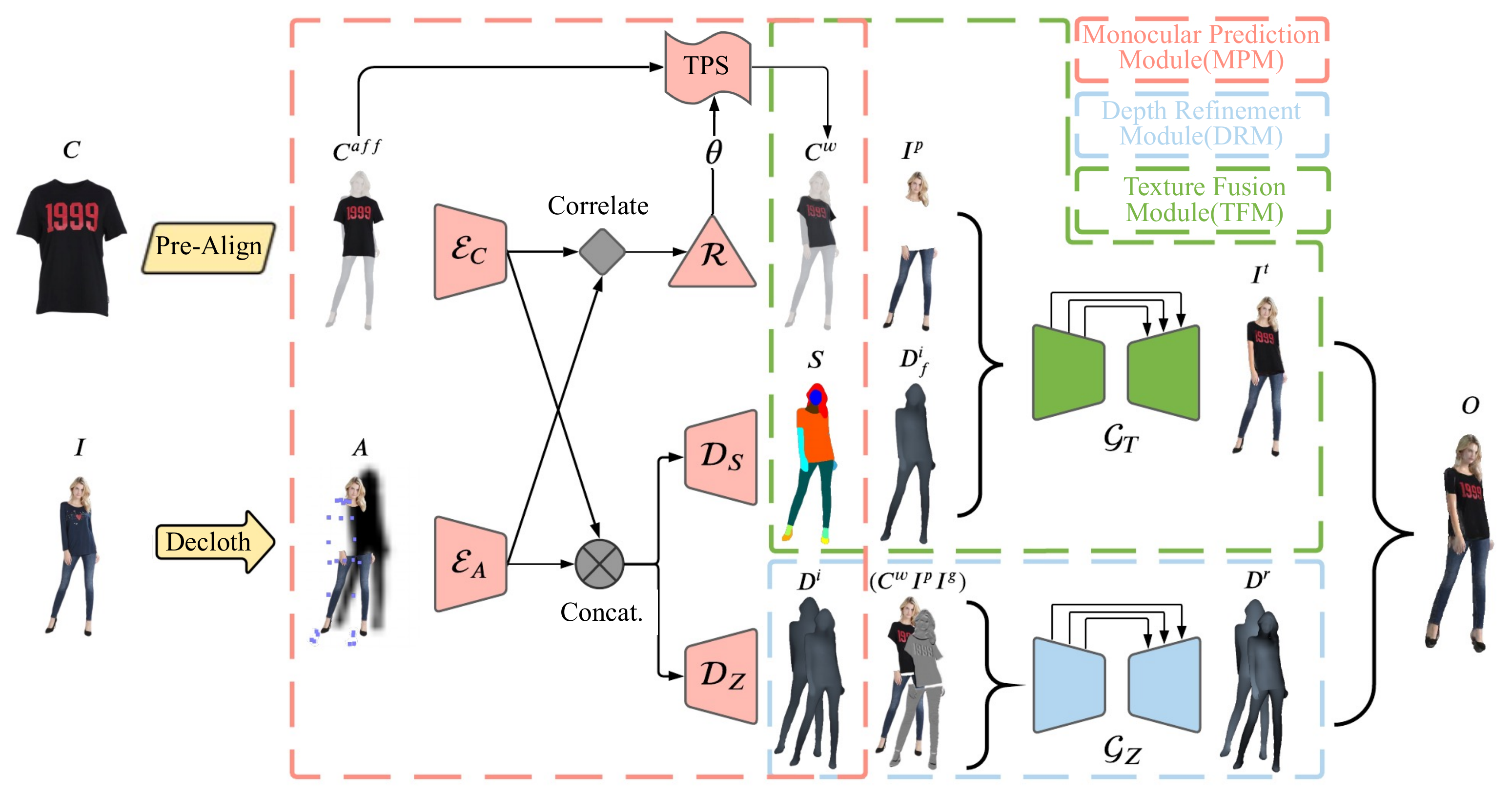}
\caption{Overview of M3D-VTON architecture. Courtesy of \cite{Zhao2021M3DVTONAM}}
\label{fig:m3d-vton_arch}
\end{figure}

In \cite{Zhao2021M3DVTONAM}, Zhao et al. presented a Monocular-to-3D Virtual Try-On Network (M3D-VTON), which also takes as input an in-shop clothing image and a person image, but output the 3D try-on mesh instead of image in 2D space (Fig \ref{fig:m3d-vton_teaser}).  By efficiently exploiting the merits of 2D non-rigid deformation and 3D non-parametric body estimation, this work successfully recovers high-fidelity 3D try-on result, being as well much faster than pure 3D methods. Fig \ref{fig:m3d-vton_arch} depicts its overall architecture that contains three modules, namely the preparatory Monocular Prediction Module (MPM), the Texture Fusion Module (TFM) to generate try-on texture and the Depth Refinement Module (DRM) to estimate the fine-detail body depths. M3D-VTON highlights the first attempt to bridge the 2D try-on and 3D human reconstruction, leading to a effective solution of 3D try-on problem in a novel view.

Some of the other promising works for AR shopping for clothing includes, SwapNet \cite{raj2018swapnet}, Learning‐based animation of clothing for virtual try‐on \cite{santesteban2019learning}, GarNet: A two-stream network for fast and accurate 3d cloth draping \cite{gundogdu2019garnet}, 360-degree textures of people in clothing from a single image \cite{lazova2019360}, M2e-try on net: Fashion from model to everyone \cite{wu2019m2e},  Fw-gan: Flow-navigated warping gan for video virtual try-on \cite{dong2019fw},  LA-VITON: a network for looking-attractive virtual try-on \cite{jae2019viton}, Fashion++: Minimal edits for outfit improvement \cite{hsiao2019fashion}, TailorNet: Predicting Clothing in 3D as a Function of Human Pose, Shape and Garment Style \cite{patel2020tailornet}, ViBE: Dressing for diverse body shapes \cite{hsiao2020vibe}, Cloth Interactive Transformer for Virtual Try-On \cite{ren2021cloth}, VITON-HD: High-Resolution Virtual Try-On via Misalignment-Aware Normalization \cite{choi2021viton}, Parser-Free Virtual Try-on via Distilling Appearance Flows \cite{ge2021parser}, and Complementary Transfering Network (CT-Net)  \cite{yang2021ct}.


\subsection{Models for Makeup Try On}
There have been several deep learning based frameworks proposed for make-up try-on. Here we provide an overview of some of the most popular ones. 

\begin{figure}[h]
\centering
\includegraphics[width=0.9\linewidth]{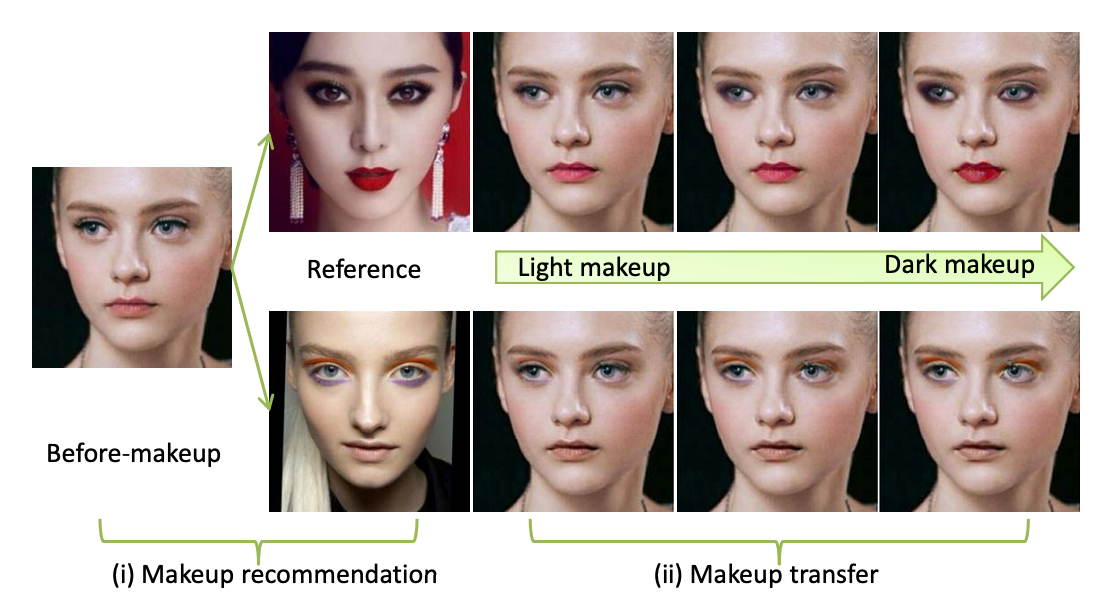}
\caption{The makeup recommendation and synthesis for an example image by DL-MTN. Courtesy of \cite{liu2016makeup}}
\label{fig:DLMTN_rec}
\end{figure}

In \cite{liu2016makeup}, Liu et al. proposed a novel Deep Localized Makeup Transfer Network to automatically recommend the most suitable makeup for a female and synthesize the makeup on her face. Given a before-makeup face, her most suitable makeup is
determined automatically. Then, both the before makeup and the reference faces are fed into the proposed Deep Transfer Network to generate the after-makeup face. The makeup recommendation for one sample image is shown in Fig \ref{fig:DLMTN_rec}.

\begin{figure}[h]
\centering
\includegraphics[width=0.99\linewidth]{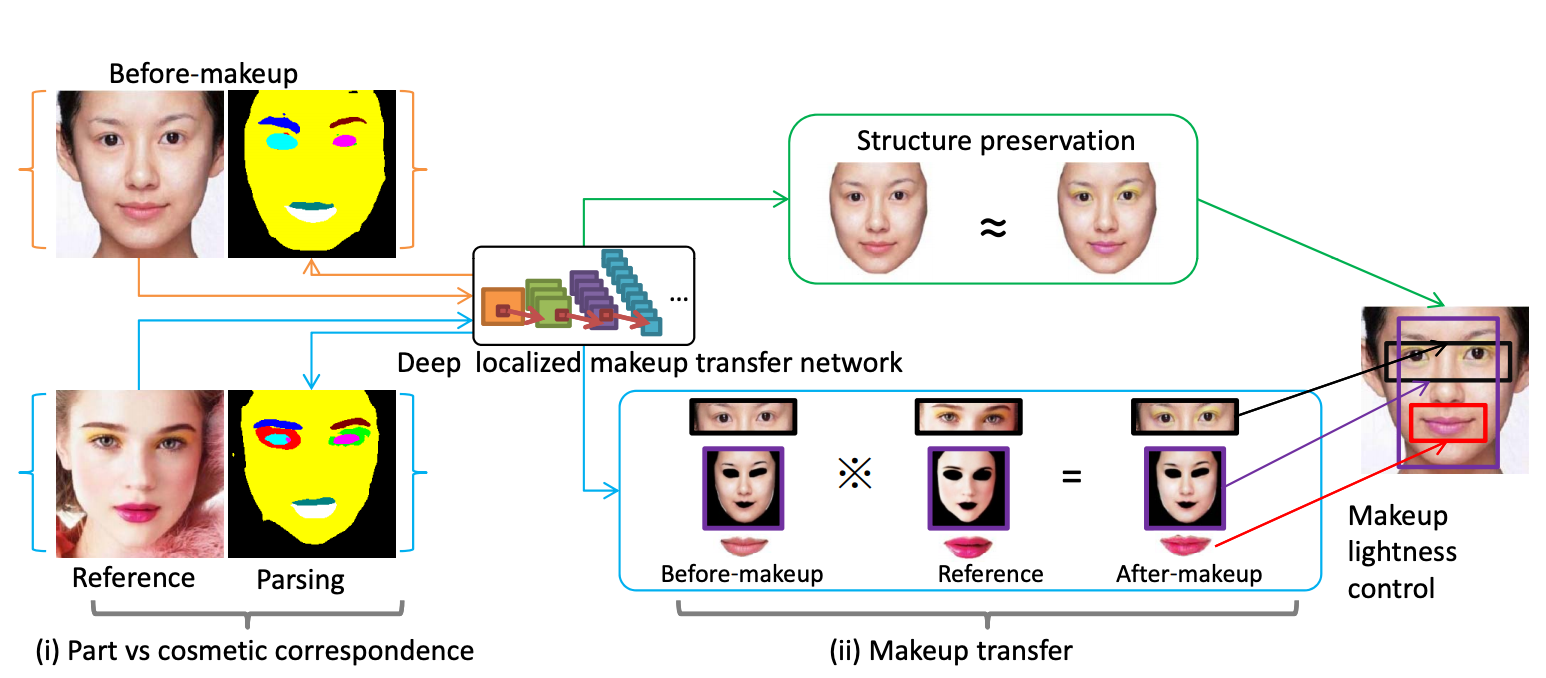}
\caption{The model architecture of Deep Localized Makeup Transfer Network. Courtesy of \cite{liu2016makeup}}
\label{fig:DLMTN_arch}
\end{figure}

The proposed Deep Localized Makeup Transfer Network contains two sequential steps. (i) the correspondences between the facial part (in the before-makeup face) and the cosmetic (in the reference face) are built based on the face parsing network. (ii) Eye shadow, foundation and lip gloss are locally transferred with a global smoothness regularization. 
The high-level architecture of this work is shown in Fig \ref{fig:DLMTN_arch}.

\begin{figure}[h]
\centering
\includegraphics[width=0.99\linewidth]{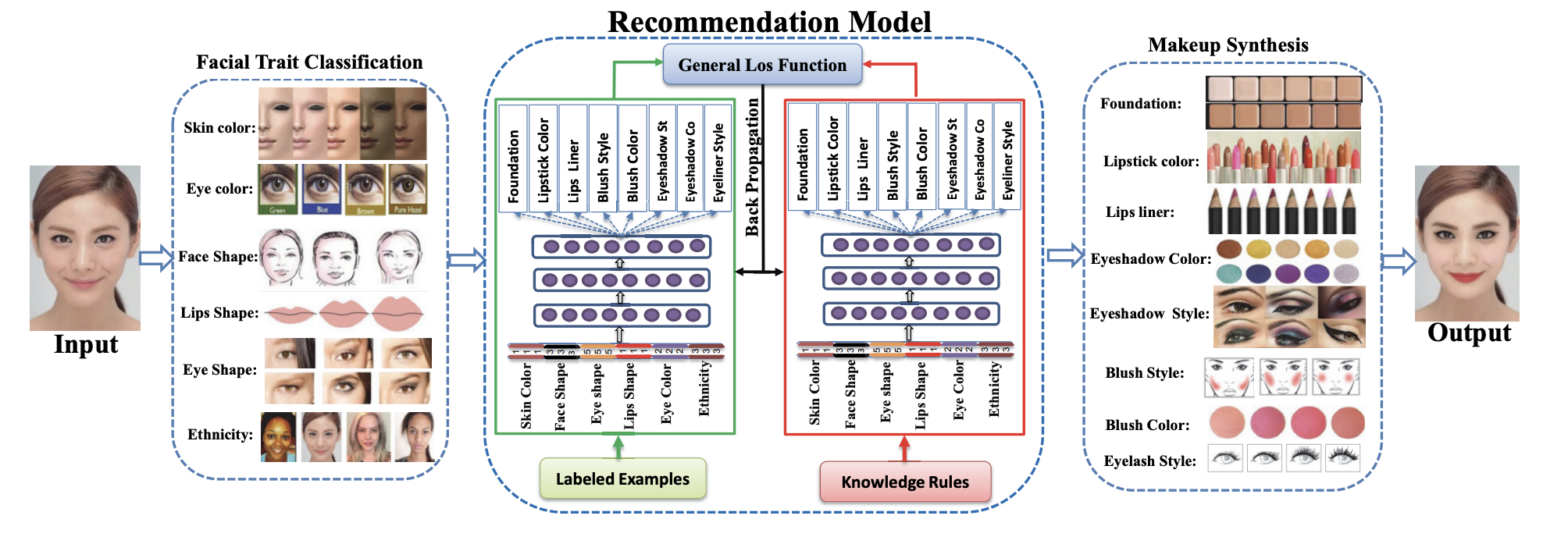}
\caption{The model architecture of the Examples-Rules Guided DNN for makeup recommendation. Courtesy of \cite{alashkar2017examples}}
\label{fig:alash_arch}
\end{figure}

In \cite{alashkar2017examples}, Alashkar et al. developed a fully automatic makeup recommendation system and proposed a novel examples-rules guided deep neural network approach.
The framework consists of three stages. First, makeup-related facial traits are classified into structured coding. Second, these facial traits are fed into examples-rules guided deep neural recommendation model which makes use of the pairwise of Before-After images and
the makeup artist knowledge jointly. Finally, to visualize the
recommended makeup style, an automatic makeup synthesis
system is developed as well.
Fig \ref{fig:alash_arch} illustrates the block-diagram of the proposed model by this work.

\begin{figure}[h]
\centering
\includegraphics[width=0.8\linewidth]{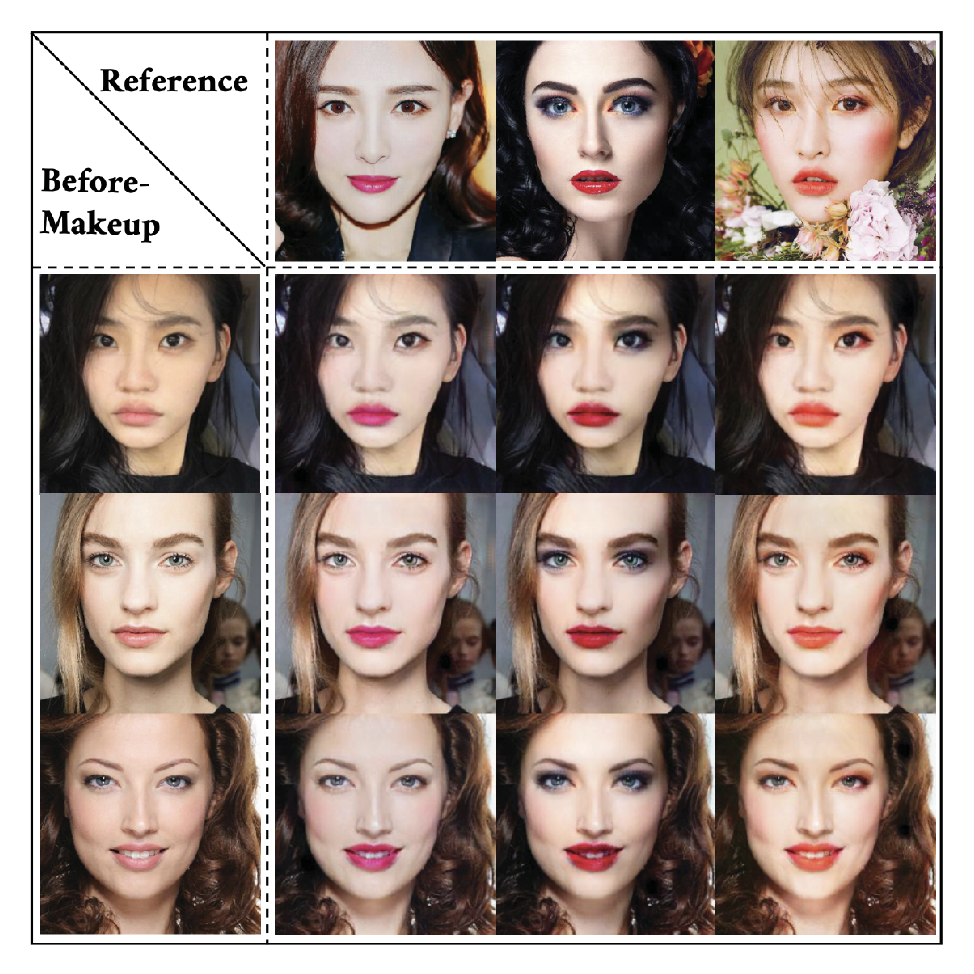}
\caption{Example results of our BeautyGAN model for makeup transfer. Courtesy of \cite{li2018beautygan}}
\label{fig:BeautyGAN_res}
\end{figure}

In \cite{li2018beautygan}, Li proposed an instance-level facial makeup transfer with generative adversarial network, called BeautyGAN. Some of the sample result generated by this framework are shown in Fig \ref{fig:BeautyGAN_res}.
They first transfer the non-makeup face to the makeup domain with a couple of discriminators that distinguish generated images from domains’ real samples. On the basis of domain-level transfer, they achieve instance-level transfer by adopting a pixel-level histogram loss calculated on different facial regions. To preserve face identity and eliminate artifacts, they also incorporate a perceptual loss and a cycle consistency loss in the overall objective function. The overall architecture of this framework is shown in Fig \ref{fig:BeautyGAN_res}.

\begin{figure}[h]
\centering
\includegraphics[width=0.8\linewidth]{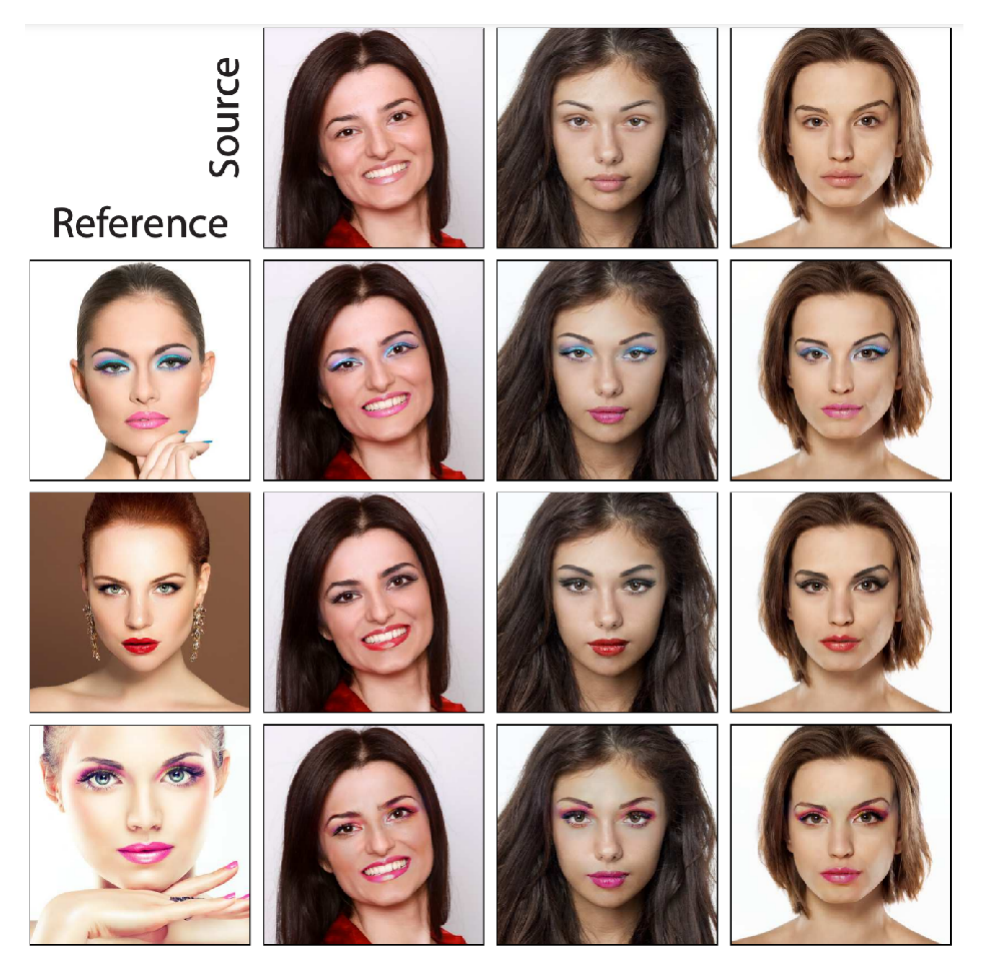}
\caption{Three source photos are each modified to
match makeup styles of three reference photos to
produce nine different outputs. Courtesy of \cite{chang2018pairedcyclegan}}
\label{fig:PairedCycleGAN}
\end{figure}

\begin{figure}[h]
\centering
\includegraphics[width=0.99\linewidth]{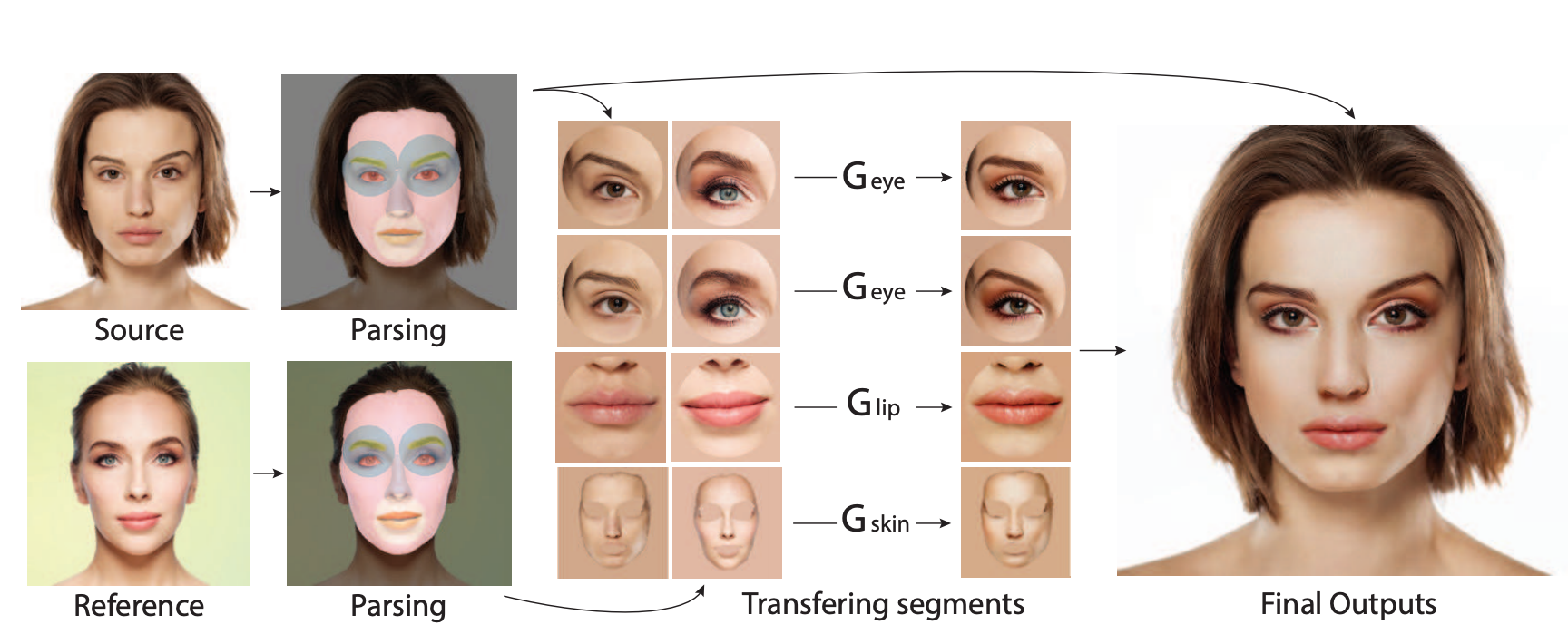}
\caption{Illustration of generator per segment in PairedCycleGAN. Courtesy of \cite{chang2018pairedcyclegan}}
\label{fig:PairedCycleGAN_gen}
\end{figure}

In \cite{chang2018pairedcyclegan}, Chang et al.  introduced an automatic method for editing a portrait photo so that the subject appears to be wearing makeup in the style of another person in a reference photo. The proposed unsupervised learning approach relies on cycle-GAN. Different from the image domain transfer problem, this style transfer problem involves two asymmetric
functions: a forward function encodes example-based style
transfer, whereas a backward function removes the style. They
constructed two coupled networks to implement these functions – one that transfers makeup style and a second that
can remove makeup – such that the output of their successive application to an input photo will match the input.
Fig \ref{fig:PairedCycleGAN} shows some of the sample makeup transferred images by this model.
For each image, they applied face parsing algorithm to segment out each facial component. And they trained three generators and discriminators separately for eyes, lip and skin considering the unique characteristics of each regions. This is shown in Fig \ref{fig:PairedCycleGAN_gen}.

\begin{figure}[h]
\centering
\includegraphics[width=0.75\linewidth]{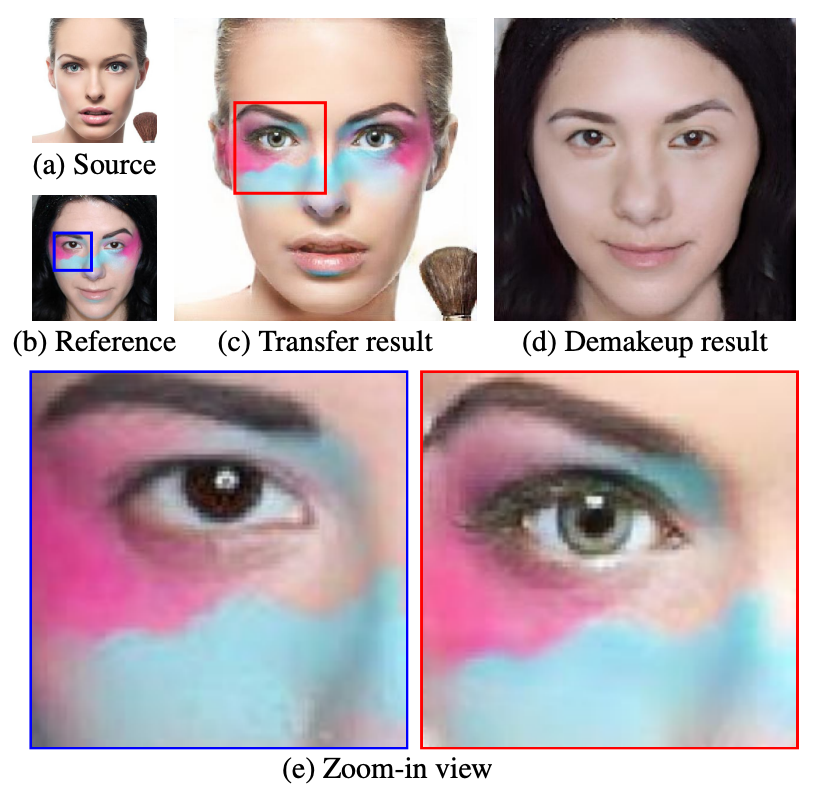}
\caption{Facial makeup and de-makeup with dramatic makeup
style using LADN framework. Courtesy of \cite{gu2019ladn}}
\label{fig:LADN_res}
\end{figure}

\begin{figure}[h]
\centering
\includegraphics[width=0.99\linewidth]{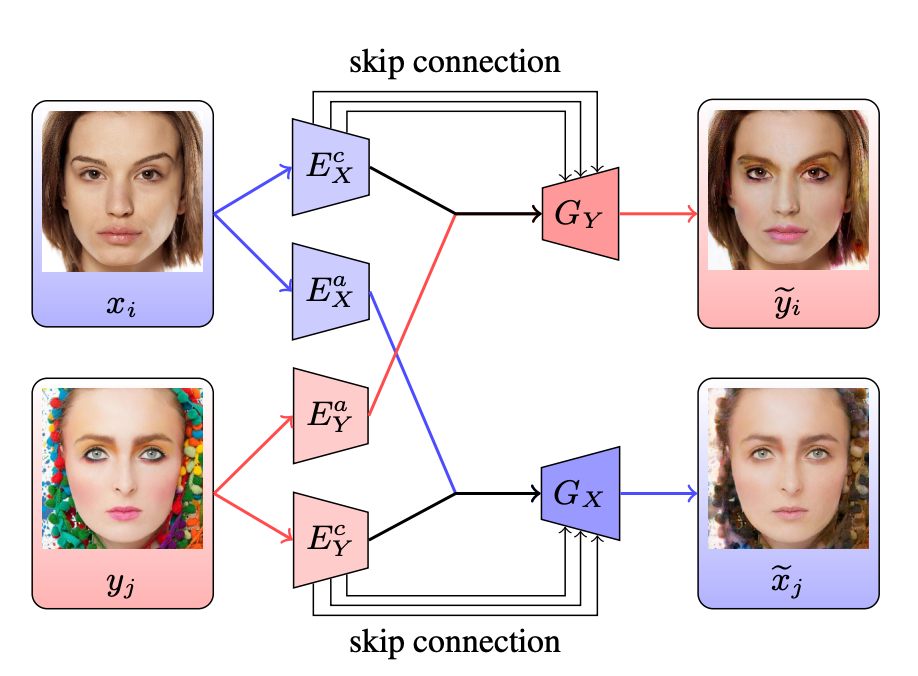}
\caption{The generator part of LADN framework.  The outputs of $E^c$ and $E^a$ are C and A, which are concatenated at the bottleneck and fed into generators. Skip connections are added between $E^c$  and G to capture more details in generated results. Courtesy of \cite{gu2019ladn}}
\label{fig:LADN_gen}
\end{figure}

In \cite{gu2019ladn}, Gu et al. proposed a local adversarial disentangling network (LADN) for facial makeup and de-makeup. Central to their method are multiple and overlapping local adversarial discriminators in a content-style disentangling network for achieving local detail transfer between facial images, with the use of asymmetric loss functions for dramatic makeup
styles with high-frequency details. 
Fig \ref{fig:LADN_res} shows the result of this framework on two sample images.
The high-level structure of the generator part of LADN is shown in Fig \ref{fig:LADN_gen}.

\begin{figure}[h]
\centering
\includegraphics[width=0.99\linewidth]{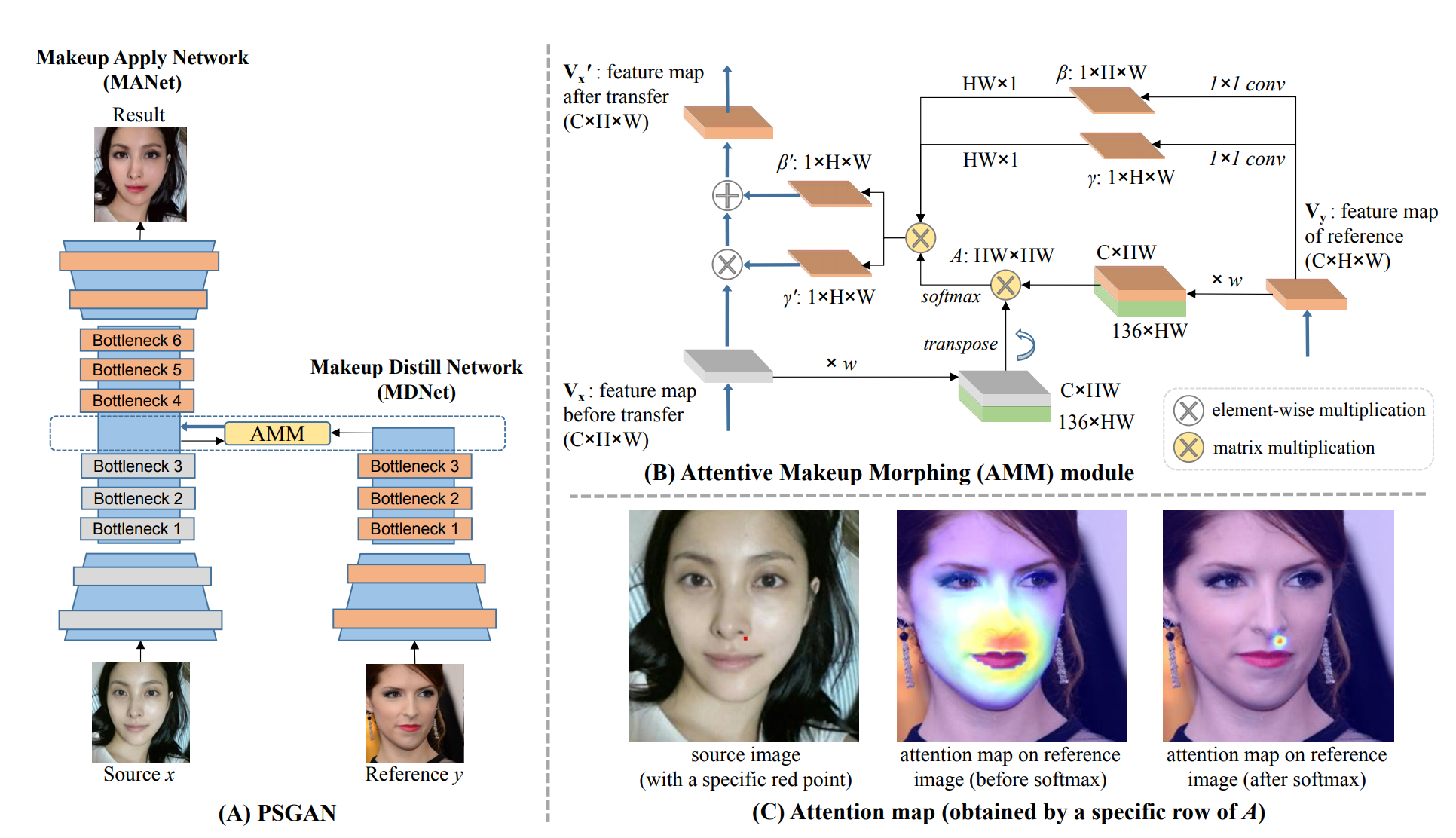}
\caption{Illustration of PSGAN framework. Courtesy of \cite{jiang2020psgan}}
\label{fig:PSGAN_arch}
\end{figure}

In \cite{jiang2020psgan}, Jiang et al. tried to address the issues with previous texisting methods for facial makeup transfer, which transferring between images with large pose and expression differences, and also not being able to realize customizable transfer that allows a controllable shade of makeup or specifies the part to transfer, which limits their applications.
They proposed Pose and expression robust Spatial-aware GAN (PSGAN). It first utilizes Makeup Distill Network to disentangle the makeup of the reference image as two spatial-aware makeup matrices. Then, Attentive Makeup Morphing module is introduced to specify how the makeup of a pixel in the source image is morphed from the reference image.
The model architecture of PSGAN framework is shown in Fig \ref{fig:PSGAN_arch}.

\begin{figure}[h]
\centering
\includegraphics[width=0.99\linewidth]{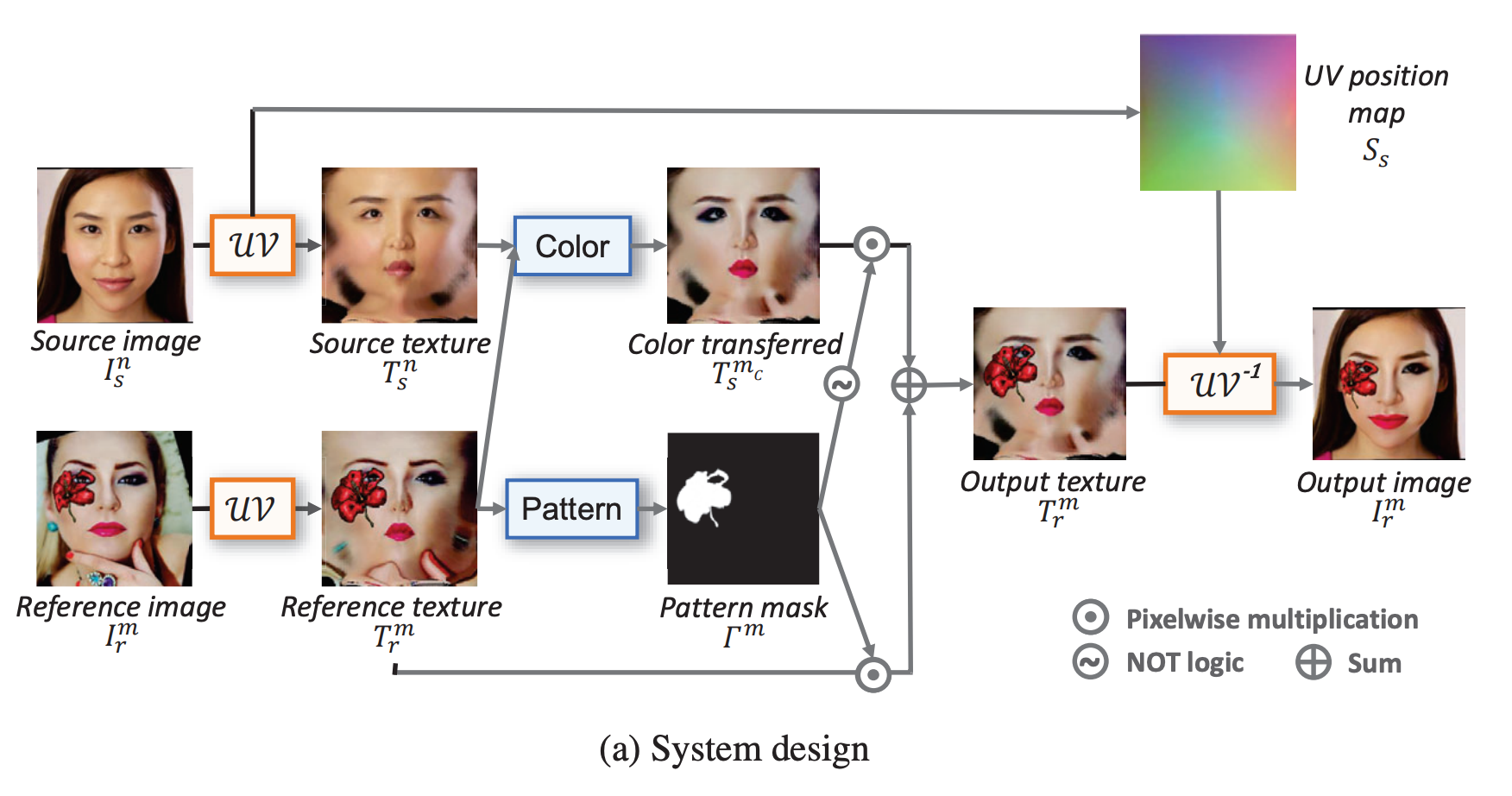}
\caption{The high-level architecture of . Courtesy of \cite{nguyen2021lipstick}}
\label{fig:Nguyen_arch}
\end{figure}

In \cite{nguyen2021lipstick}, Nguyen et al. proposed a holistic makeup transfer framework that can handle all the mentioned makeup components. It consists of an improved color transfer branch and a novel pattern transfer branch to learn all makeup properties, including color, shape, texture, and location. To train and evaluate such a system, we also introduce new makeup datasets for real and synthetic extreme makeup.
Fig \ref{fig:Nguyen_arch} shows the high level architecture of the proposed framework.

Some of the other promising works for virtual makeup try-on includes: makeup removal via bidirectional tunable de-makeup network \cite{cao2019makeup}, face beautification: Beyond makeup transfer \cite{liu2019face}, BeautyGlow \cite{chen2019beautyglow}, face beautification via dynamic skin smoothing, guided feathering, and texture restoration \cite{velusamy2020fabsoften}, and weakly supervised color aware GAN for controllable makeup transfer \cite{kips2020gan}.

\subsection{Models for Face/Body Transformations}
Face style transfer or (face transformation) is another active research area, with huge applications in social media such Snapchat Lenses, Instagram Filters, TikTok lenses/effects. 
Although the algorithm used by those companies is not known, there are several research works which have developed algorithms for applying various effects on faces. 
Since acquiring paired training data for face transformation is not very easy in most cases, we are going to mostly focused on algorithms which would work in an unpaired fashion here. 

\begin{figure}[h]
\centering
\includegraphics[width=0.99\linewidth]{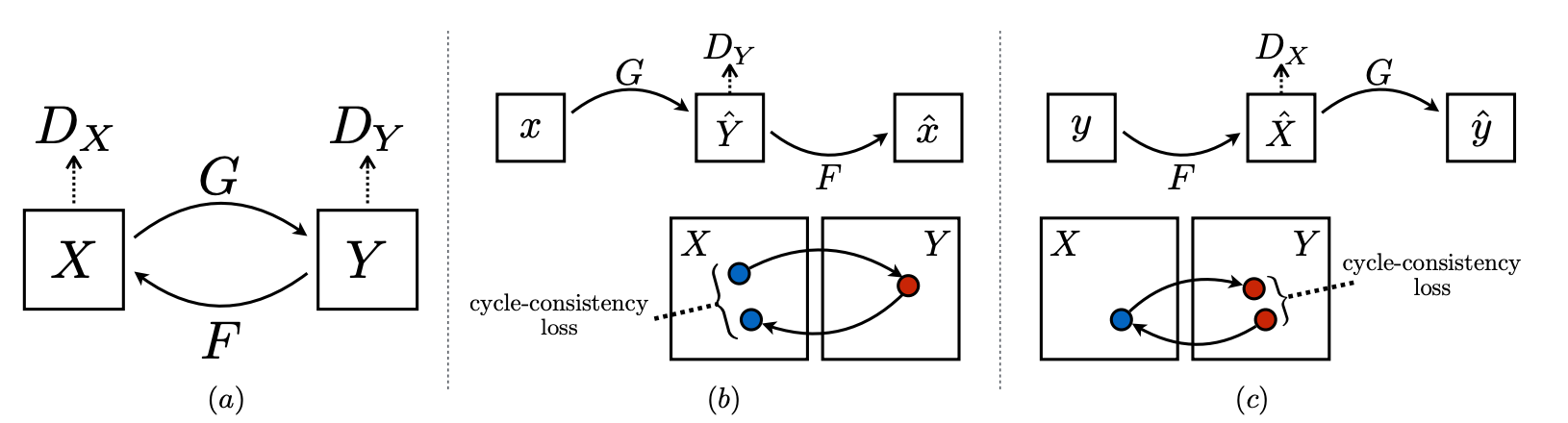}
\caption{CycleGAN model contains two mapping functions $G : X\rightarrow Y$ and $F : Y\rightarrow X$, and associated adversarial discriminators $D_Y$ and $D_X$. $D_Y$
encourages G to translate X into outputs indistinguishable from domain Y , and vice versa for $D_X$, F, and X. To further regularize the mappings, they introduced two “cycle consistency losses” that capture the intuition that if they translate from one domain to the other and back again we should arrive where they started. Courtesy of \cite{zhu2017unpaired}}
\label{fig:cycle_arch}
\end{figure}

\begin{figure}[h]
\centering
\includegraphics[width=0.99\linewidth]{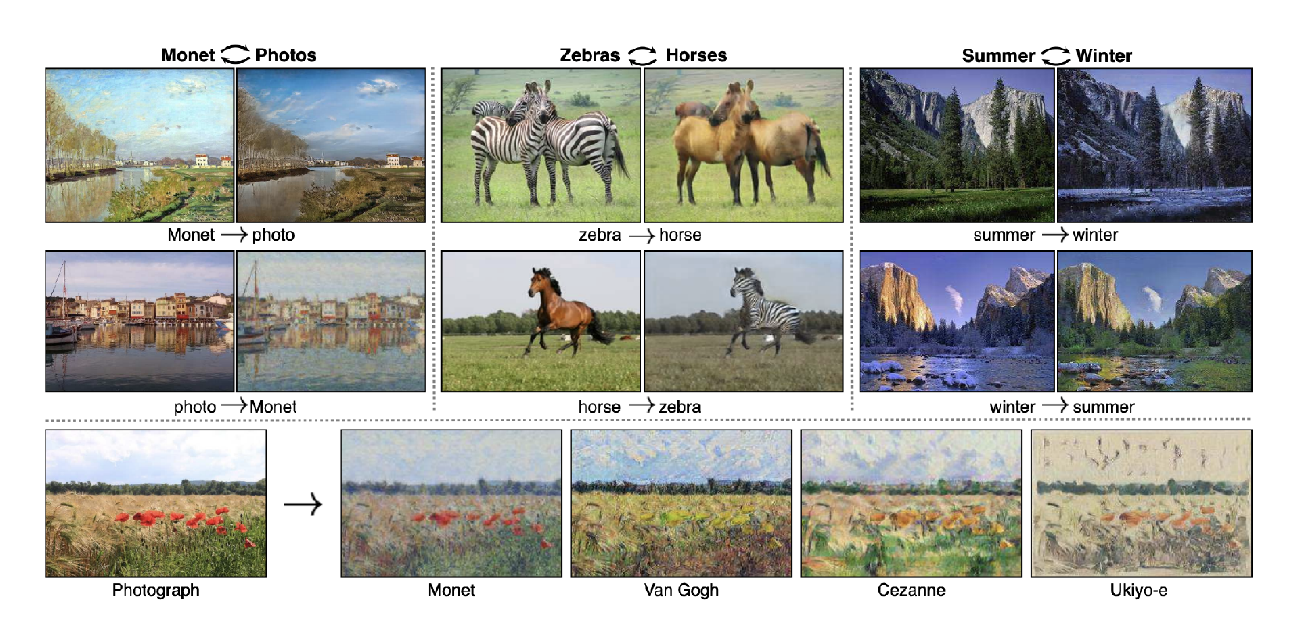}
\caption{The sample images transferred via CycleGAN model. Courtesy of \cite{zhu2017unpaired}}
\label{fig:cycle_samples}
\end{figure}

In \cite{zhu2017unpaired}, Zhu et al. presented an approach for learning to translate an image from a source domain X to a target domain Y in the absence of paired examples. Their goal is to learn a mapping $G : X \rightarrow Y$ such that the distribution of images from G(X) is indistinguishable from the distribution Y using an adversarial loss. Because this mapping is highly under-constrained, they coupled it with an inverse mapping $F : Y \rightarrow X$ and introduce a cycle consistency loss to push F(G(X)) $\approx$ X (and vice versa). 
The high-level idea of CycleGAN framework is shown in Fig \ref{fig:cycle_arch}.
Some of the sample images generated via CycleGAN model are shown in Fig \ref{fig:cycle_samples}.

In \cite{yi2017dualgan}, Yi et al. developed dual-GAN mechanism, which enables image translators to be trained from two sets of unlabeled images from two domains. In their architecture, the primal GAN learns to translate images from domain U to those in domain V , while the dual GAN learns to invert the task. The closed loop made by the primal and dual tasks allows images from either domain to be translated and then reconstructed. Hence a loss function that accounts for the reconstruction error of images can be used to train the translators. On high-level, dual-GAN and CycleGAN share a lot of similarities.

\begin{figure}[h]
\centering
\includegraphics[width=0.99\linewidth]{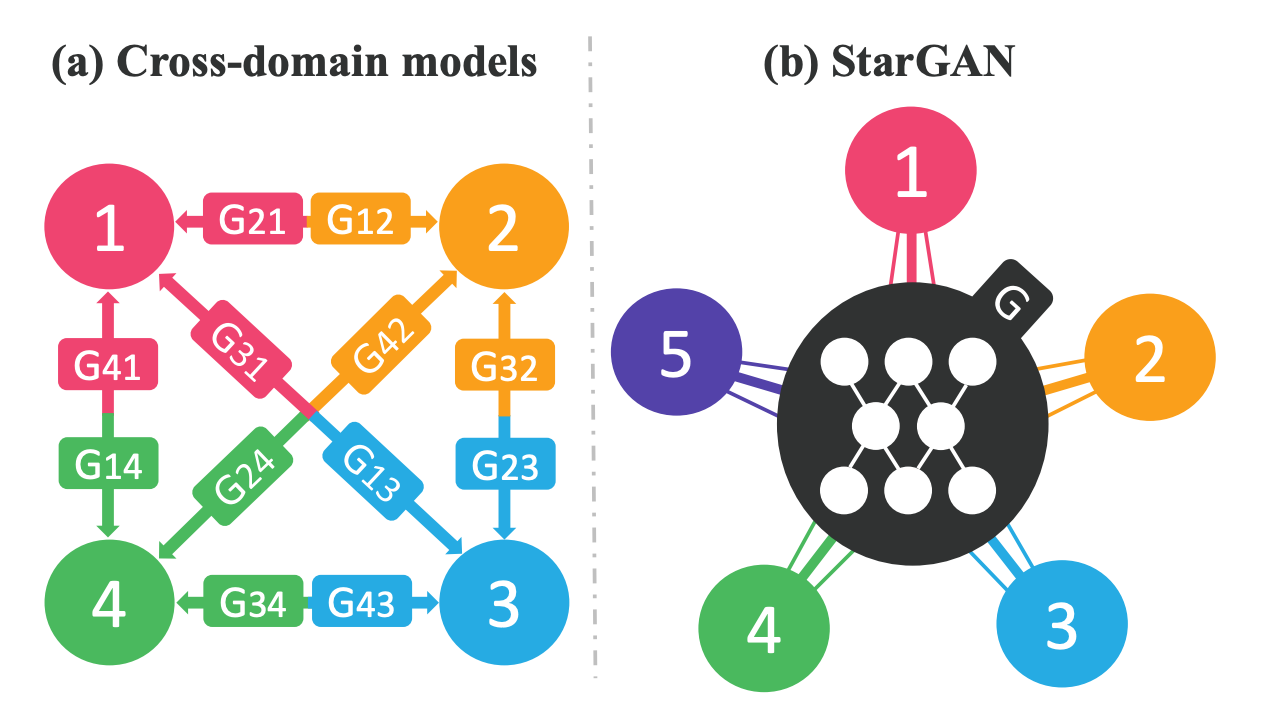}
\caption{Comparison between cross-domain models and the proposed StarGAN model. Courtesy of \cite{choi2018stargan}}
\label{fig:StarGAN_comp}
\end{figure}

\begin{figure}[h]
\centering
\includegraphics[width=0.9\linewidth]{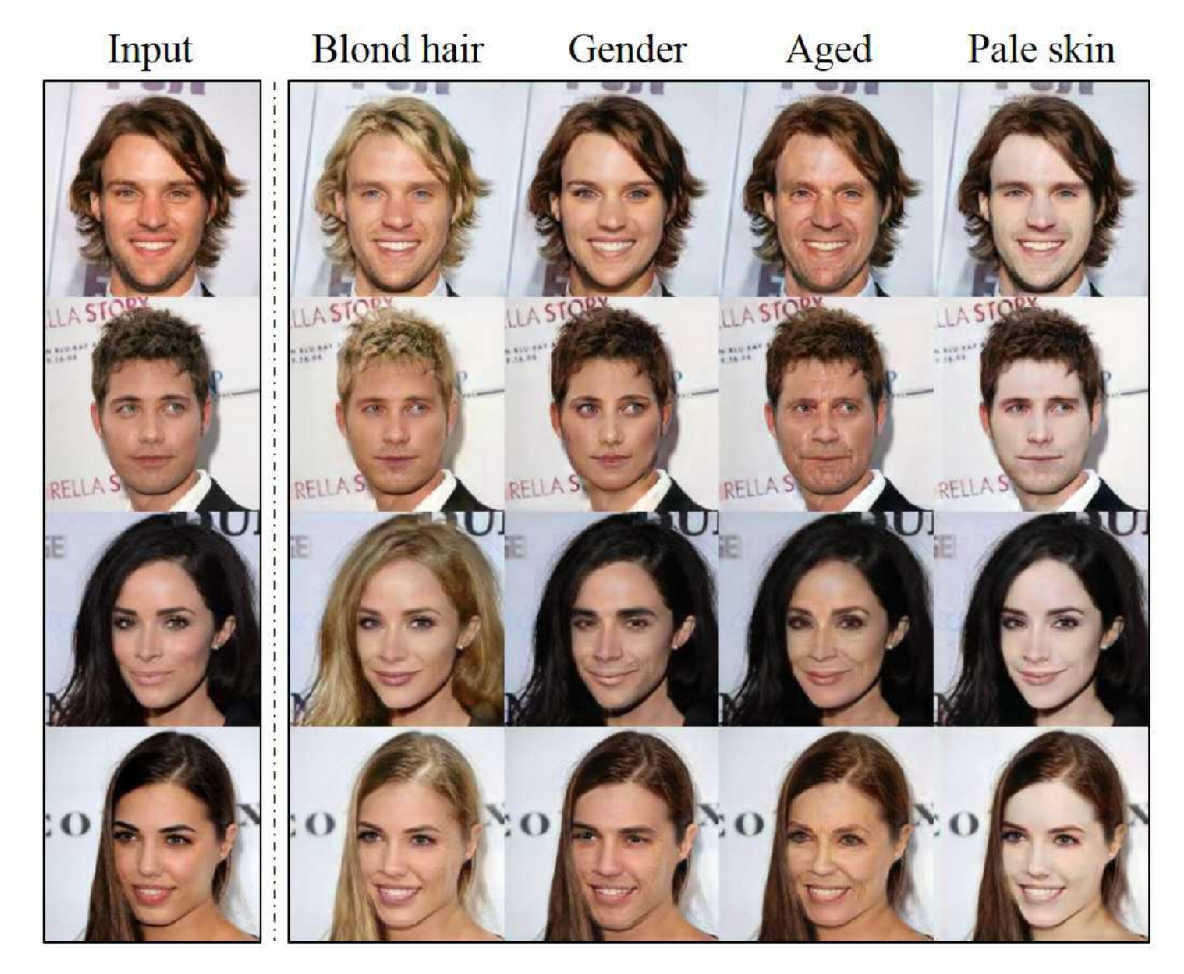}
\caption{The first column shows input images, while the remaining columns are images generated by StarGAN. Courtesy of \cite{choi2018stargan}}
\label{fig:StarGAN_samp}
\end{figure}

In \cite{choi2018stargan}, Choi et al. proposed StarGAN, a novel and scalable approach that can perform image-to-image translations for multiple domains using only a single model.
Such a unified model architecture of StarGAN allows simultaneous training of multiple datasets with different domains within a single network. This leads to StarGAN’s superior quality of translated images compared to existing models as well as the novel capability of flexibly translating an input image to any desired target domain. 
This is specially useful when the number of domains are large, as shown in Fig \ref{fig:StarGAN_comp}.
Some of the generated models via StarGAN model for different human emotions are shown in Fig \ref{fig:StarGAN_samp}.

\begin{figure}[h]
\centering
\includegraphics[width=0.99\linewidth]{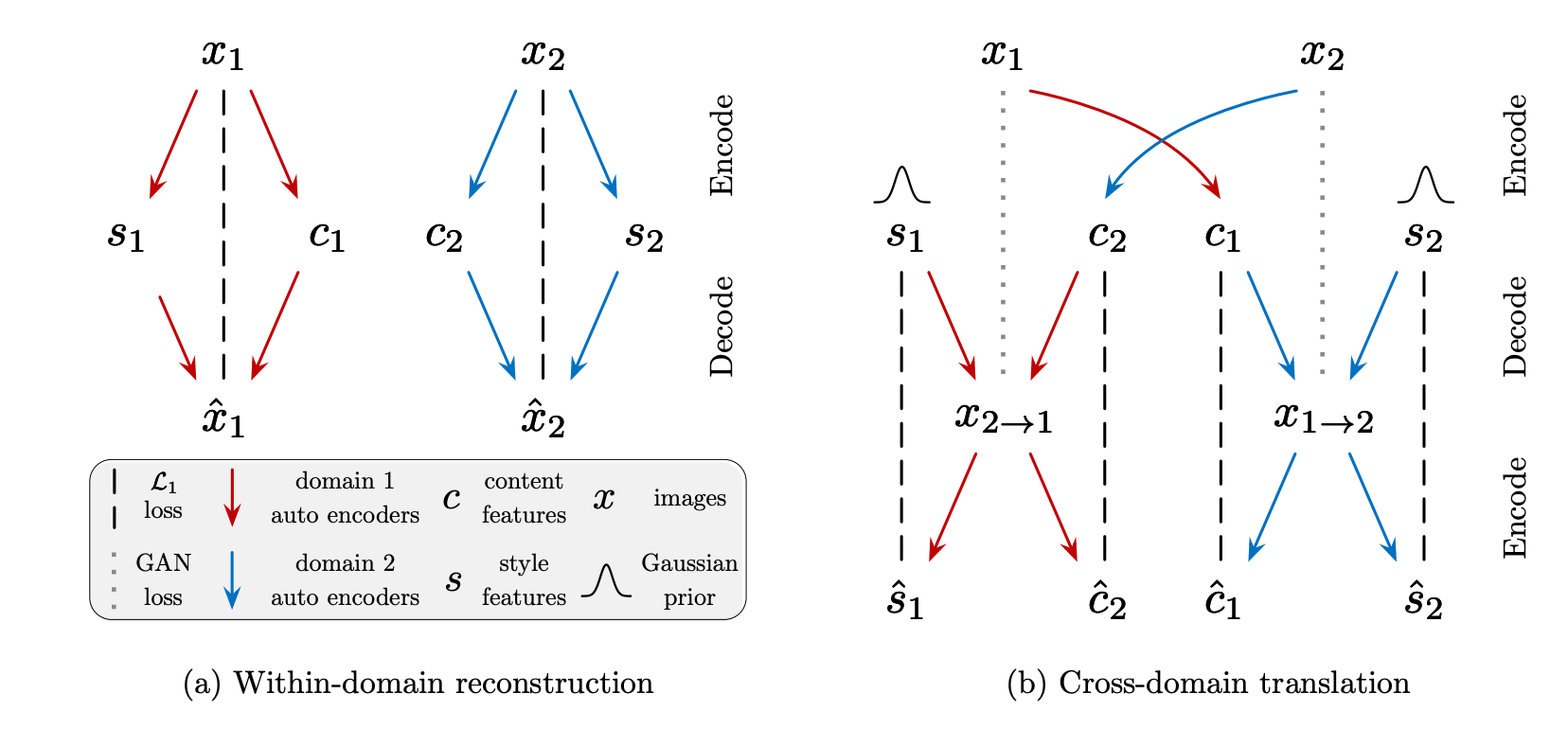}
\caption{The overview of MUNIT framework. Their image-to-image translation model consists of two autoencoders (denoted by red and blue arrows respectively), one for each domain. The latent
code of each auto-encoder is composed of a content code c and a style code s. They train the model with adversarial objectives (dotted lines) that ensure the translated images
to be indistinguishable from real images in the target domain, as well as bidirectional reconstruction objectives (dashed lines) that reconstruct both images and latent codes. Courtesy of \cite{huang2018multimodal}}
\label{fig:MUNIT_arch}
\end{figure}

In \cite{huang2018multimodal}, Huang et al. proposed a Multimodal Unsupervised Image-to-image Translation (MUNIT) framework. They assumed that the image representation can be decomposed into a content code that is domain-invariant,
and a style code that captures domain-specific properties. To translate an image to another domain, they recombine its content code with a random style code sampled from the style space of the target domain. 
The high-level overview of this framework is shown in Fig \ref{fig:MUNIT_arch}.

\begin{figure}[h]
\centering
\includegraphics[width=0.99\linewidth]{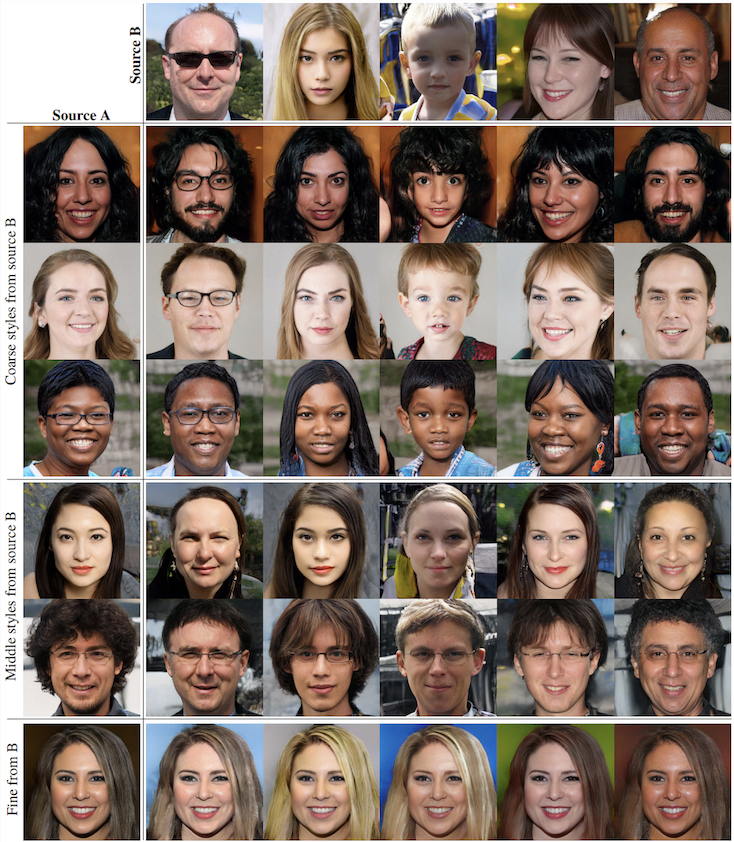}
\caption{Two sets of images were generated from their respective latent codes (sources A and B); the rest of the images were generated by copying a specified subset of styles from source B and taking the rest from source A. Courtesy of \cite{karras2019style}}
\label{fig:stylegan_sam}
\end{figure}

In \cite{karras2019style}, Karras et al. proposed an alternative generator architecture for generative adversarial networks (which is also called StyleGAN), borrowing from style transfer
literature. The new architecture leads to an automatically
learned, unsupervised separation of high-level attributes
(e.g., pose and identity when trained on human faces) and
stochastic variation in the generated images (e.g., freckles,
hair), and it enables intuitive, scale-specific control of the
synthesis. When it was proposed, the new generator improved the state-of-the-art in terms of traditional distribution quality metrics, led to demonstrably better interpolation properties, and also better disentangles the latent factors of variation.
StyleGAN model opened up the door for many of deep learning based (realistic) AR effects on human face and body images.
Some of the sample images generated by StyleGAN model are shown in Fig \ref{fig:stylegan_sam}.

\begin{figure}[h]
\centering
\includegraphics[width=0.99\linewidth]{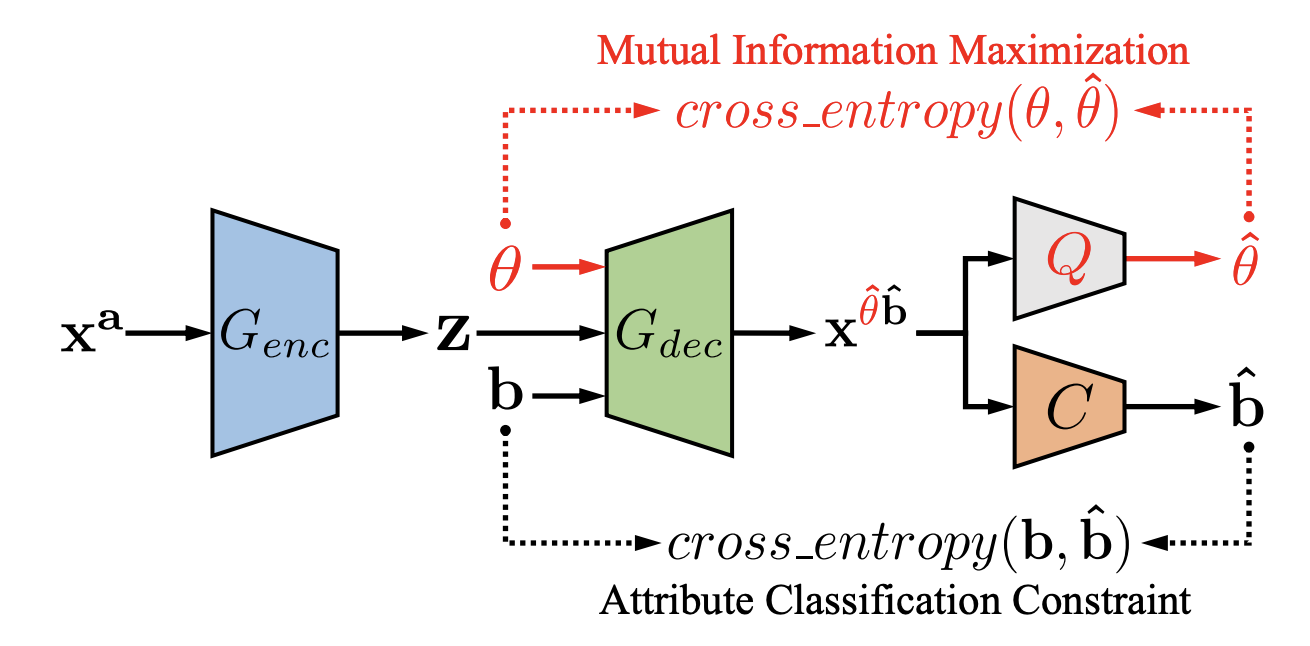}
\caption{Illustration of AttGAN extension for attribute style manipulation. Courtesy of \cite{he2019attgan}}
\label{fig:AttGAN_arch}
\end{figure}

\begin{figure}[h]
\centering
\includegraphics[width=0.9\linewidth]{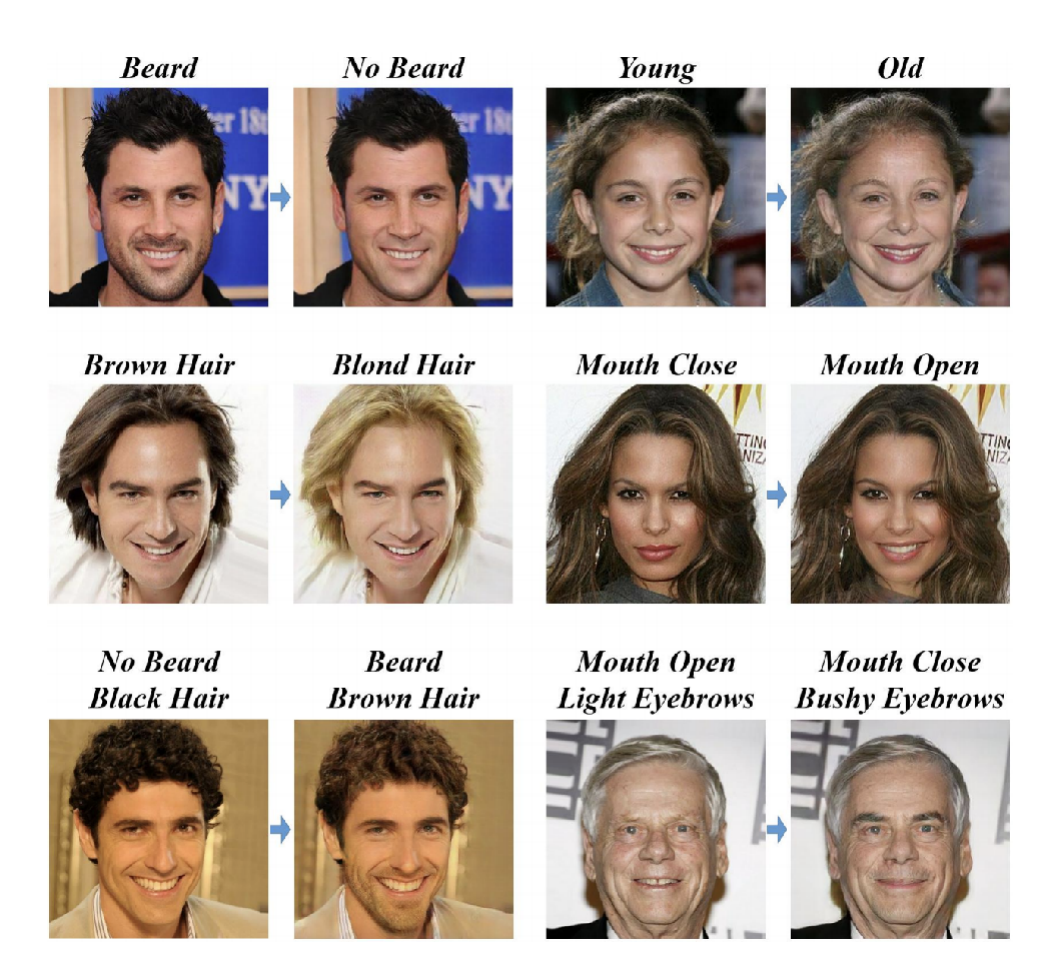}
\caption{Facial attribute editing results from our AttGAN. Courtesy of \cite{he2019attgan}}
\label{fig:AttGAN_samp}
\end{figure}

In \cite{he2019attgan}, He et al. developed AttGAN, which applies an attribute classification constraint to the generated image to just guarantee the correct change of desired attributes, i.e., to “change what you want”. Meanwhile, the reconstruction learning is introduced to preserve attribute-excluding details, in other words, to “only change what you want”. Besides, the adversarial learning is employed for visually realistic editing. These three components cooperate with each other forming an effective framework for high quality facial attribute editing. 
Fig \ref{fig:AttGAN_arch} shows the high-level architecture of AttGAN framework.
Some of the sample results of this model are shown in Fig \ref{fig:AttGAN_samp}.

In \cite{choi2020stargan}, Choi et al. proposed StarGAN v2, a single framework that tackles the following properties and shows significantly improved results over the baselines. 
On one hand it tries to have a good diversity among the generated images
and on the other hand it tries to achieve scalability over multiple domains.

\begin{figure}[h]
\centering
\includegraphics[width=0.9\linewidth]{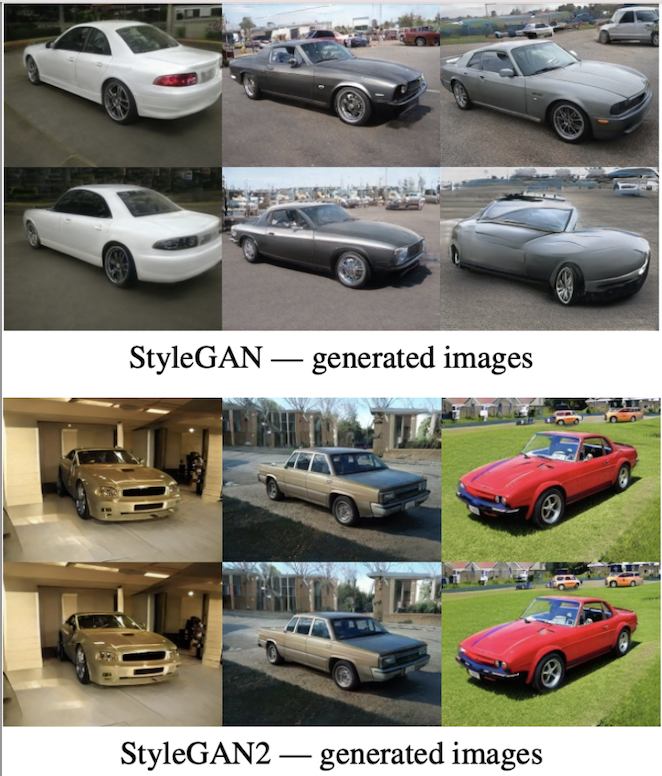}
\caption{Example images and their projected and re-synthesized counterparts. For each configuration, top row shows the target images and bottom row shows the synthesis of the corresponding projected latent vector and noise inputs. With the baseline StyleGAN, projection
often finds a reasonably close match for generated images, but especially the backgrounds differ from the originals. The images generated using StyleGAN2 can be projected almost perfectly back into generator inputs, while projected real images (from the training set) show clear differences to the originals, as expected. Courtesy of \cite{karras2020analyzing}}
\label{fig:styleganv2}
\end{figure}

In \cite{karras2020analyzing}, Karras et al.  proposed StyleGAN-v2, which introduces changes in StyleGAN's both model architecture and training methods to address some of the previous issues.
In particular, they redesigned the generator normalization, revisit progressive growing, and regularize the generator to
encourage good conditioning in the mapping from latent
codes to images.
In addition to improving image quality, this path length regularizer yields the additional benefit that the generator becomes significantly easier to invert. This makes it possible to reliably attribute a generated image to a particular network.
Some of the example images and their projected and re-synthesized counterparts with StyleGAN and StyleGAN2 are shown in Fig \ref{fig:styleganv2}.

\begin{figure}[h]
\centering
\includegraphics[width=0.9\linewidth]{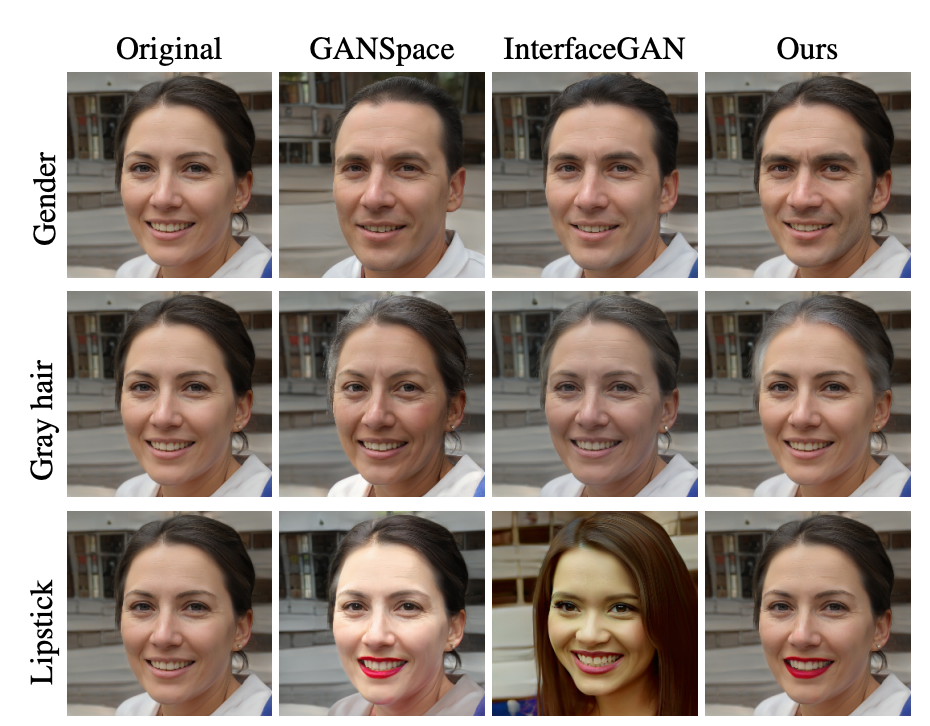}
\caption{Comparison with state-of-the-art methods using the same amount of manipulation. Courtesy of \cite{wu2021stylespace}}
\label{fig:StyleSpace}
\end{figure}

In \cite{wu2021stylespace}, Wu et al. explored and analyzed the latent style space of StyleGAN2, a state-of-the-art architecture for image generation, using models pre-trained on several different datasets.
They first showed that StyleSpace, the space of channel-wise
style parameters, is significantly more disentangled than
the other intermediate latent spaces explored by previous
works. They also described a method for discovering a large
collection of style channels, each of which is shown to control a distinct visual attribute in a highly localized and disentangled manner.  Furthermore, they proposed a simple method for identifying style channels that control a specific attribute, using a pre-trained classifier or a small number of example images.
The comparison of StyleSpace with some of the other frameworks are shown in Fig \ref{fig:StyleSpace}.

\begin{figure}[h]
\centering
\includegraphics[width=0.99\linewidth]{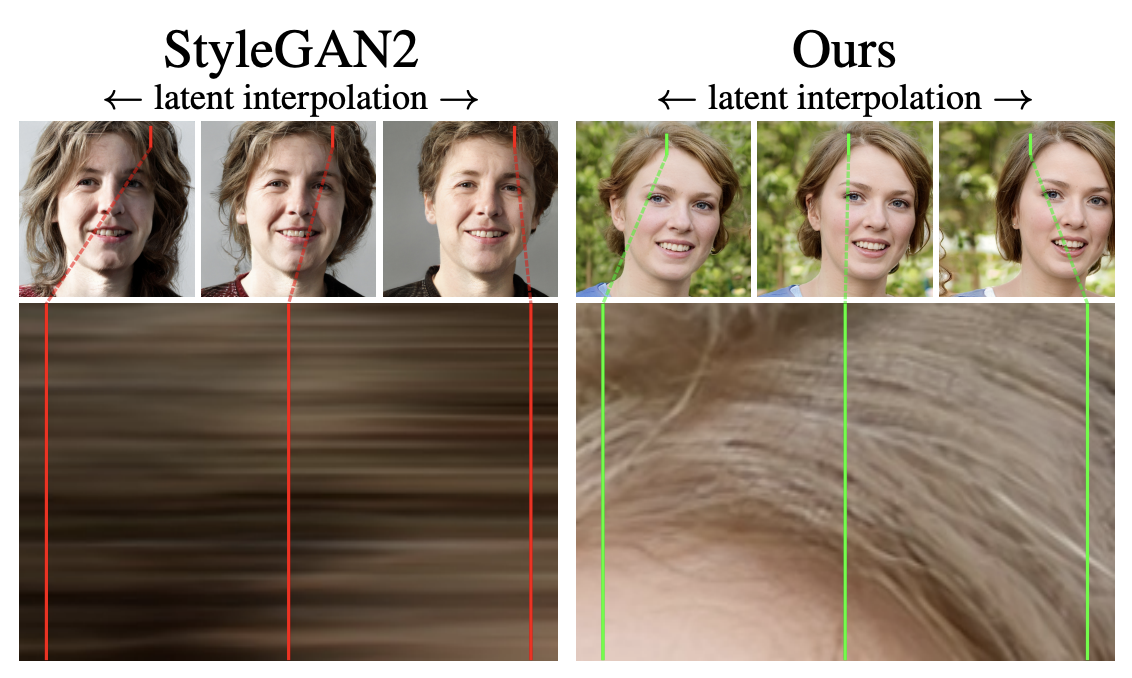}
\caption{Examples of “texture sticking”. From a latent space interpolation (top row), they extract a short vertical segment of pixels from each generated image and stack them
horizontally (bottom). The desired result is hairs moving in animation, creating a time-varying field. With StyleGAN2 the hairs mostly stick to the same coordinates, creating horizontal streaks instead. Courtesy of \cite{karras2021alias}}
\label{fig:aliasfree}
\end{figure}

In \cite{karras2021alias}, Karras et al. discussed that despite their hierarchical convolutional nature, the synthesis process of typical generative adversarial networks depends on absolute pixel coordinates in an unhealthy manner. This manifests itself as, e.g., detail appearing to
be glued to image coordinates instead of the surfaces of depicted objects. They  traced the root cause to careless signal processing that causes aliasing in the generator
network.
Interpreting all signals in the network as continuous, they derive generally applicable, small architectural changes that guarantee that unwanted information cannot leak into the hierarchical synthesis process.  The resulting networks match the FID of StyleGAN2 but differ dramatically in their internal representations, and they are fully equivariant to translation and rotation even at subpixel scales.
Some of the examples of “texture sticking” usign this model and also StyleGAN2 are shown in  Fig \ref{fig:aliasfree}.

\subsection{Tracking and Pose Estimation for AR}
Augmented reality has revolutionized the gaming industry, and there have been several AR based games which have been developed in the past decade, such as Pokemon Go, Jurassic  World  Aliev, The Walking  Dead: Our  World, and many more. 
There are various algorithms which are the core of AR based games, such as tracking, scene understanding, and reconstruction.
In this part, we focus on the tracking frameworks, which involve algorithms for tracking a target object/environment via cameras and sensors, and estimating viewpoint poses.
Although vision is not the only modality used for tracking in AR applications, given the scope of this paper, we mainly focus on vision based tracking frameworks.

\subsubsection{Eye Tracking and Gaze Estimation}
\begin{figure}[h]
\centering
\includegraphics[width=0.99\linewidth]{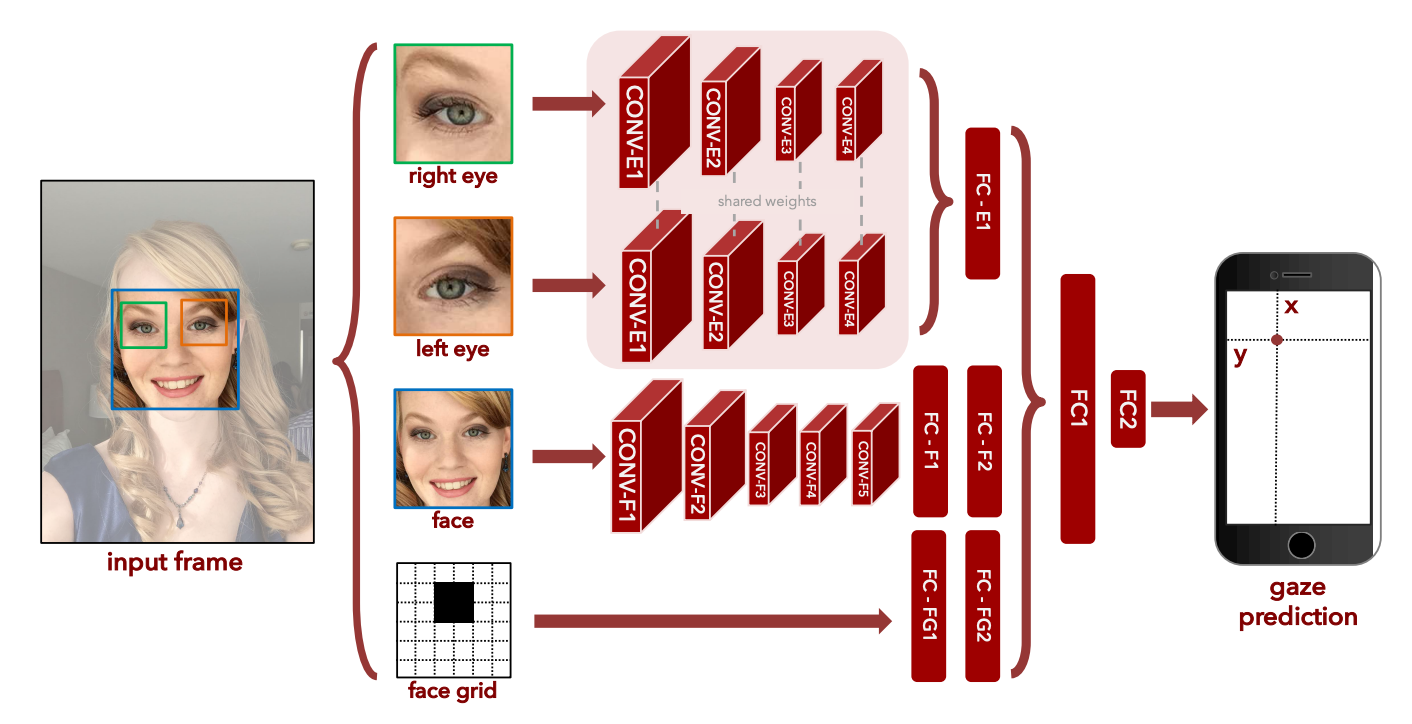}
\caption{An overview of iTracker framework. Courtesy of \cite{krafka2016eye}}
\label{fig:iTracker}
\end{figure}

In \cite{krafka2016eye}, Krafka et al. introduced GazeCapture, the first large-scale dataset for eye tracking, containing data from over 1450 people consisting of almost 2.5M frames. Using GazeCapture, they trained iTracker, a convolutional neural network for eye tracking, which achieved a significant reduction in error over previous approaches while running in real time (10–15fps) on a modern mobile device. Their model achieved a prediction error of 1.71cm and 2.53cm without calibration on mobile phones and tablets respectively.
An overview of iTracker is shown in  Fig \ref{fig:iTracker}.

In \cite{zhang2017s}, Zhang et al. proposed an appearance-based method for eye gaze estimation that, in contrast to a long-standing line of work in computer vision, only takes the full face image as input. This method encodes the face image using a convolutional neural network with spatial weights applied on the feature maps to flexibly suppress or enhance information in different facial regions.

\begin{figure}[h]
\centering
\includegraphics[width=0.99\linewidth]{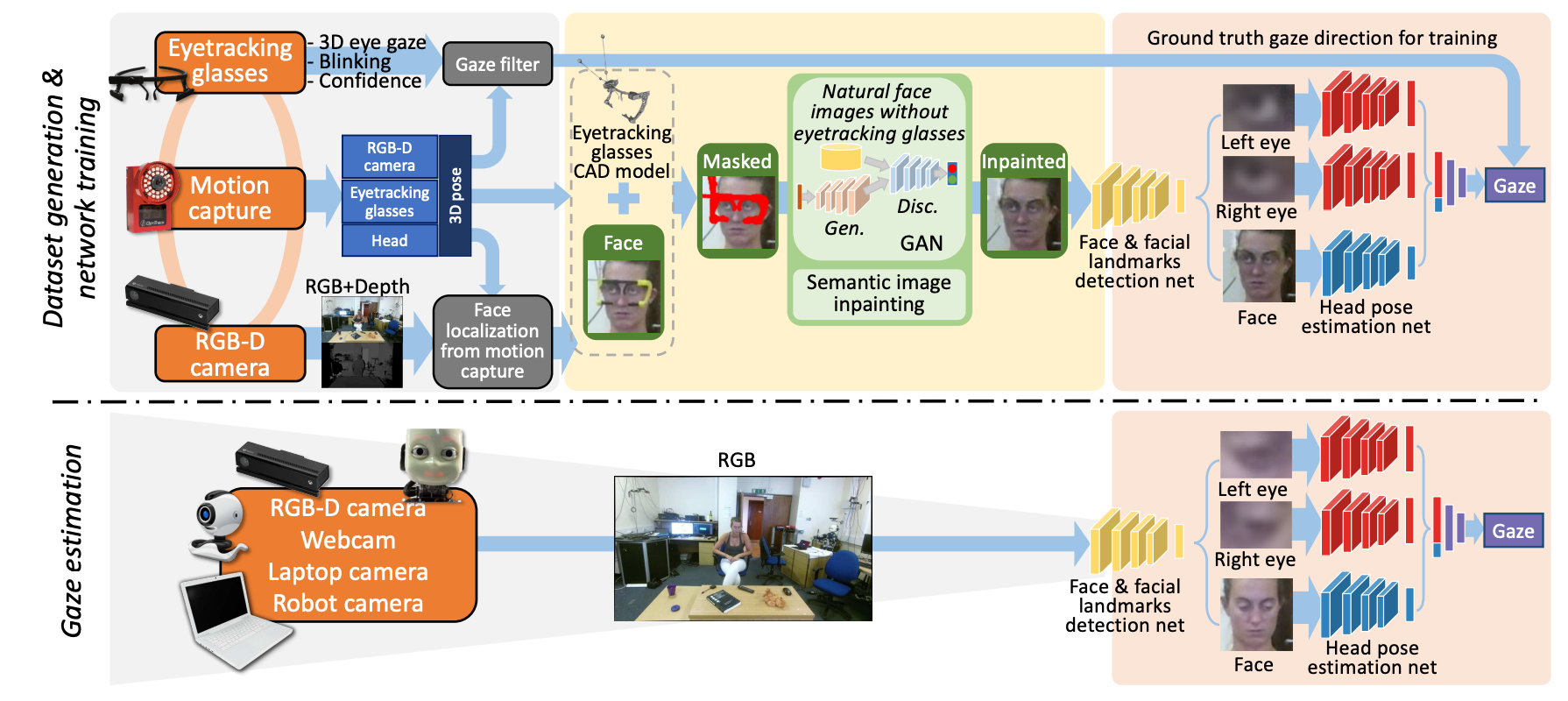}
\caption{An overview of RT-GENE architecture. Courtesy of \cite{fischer2018rt}}
\label{fig:RT-GENE}
\end{figure}

In \cite{fischer2018rt}, Fischer et al. tried to address two limitations of the previous gaze estimation frameworks, which are: hindered ground truth gaze annotation and diminished gaze estimation accuracy as image resolution decreases with distance.
They introduced a novel dataset of varied gaze and head pose images in a natural environment, and also presented a
new real-time algorithm involving appearance-based deep convolutional neural networks with increased capacity to cope with the diverse images in the new dataset. The architecture of this model is shown in 
Fig \ref{fig:RT-GENE}.

\begin{figure}[h]
\centering
\includegraphics[width=0.9\linewidth]{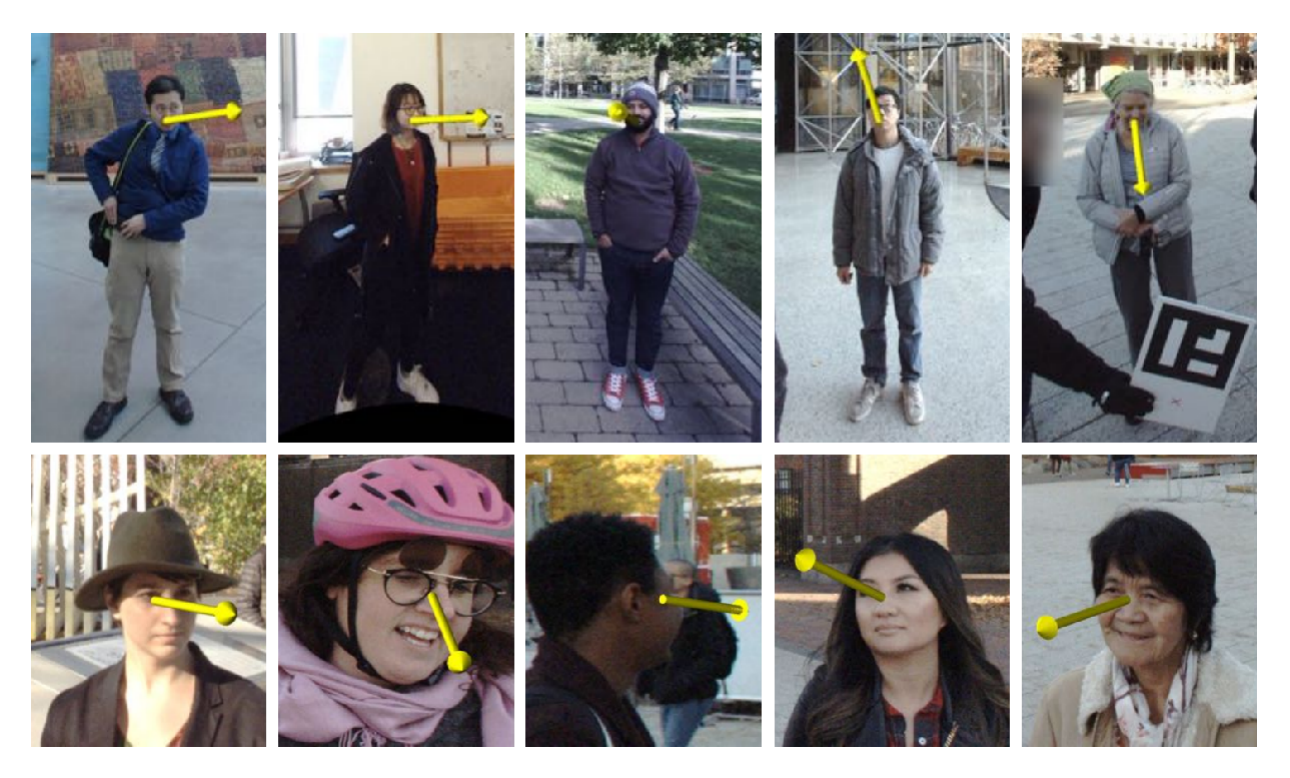}
\caption{Some of the sample images from Gaze360 dataset. Courtesy of \cite{kellnhofer2019gaze360}}
\label{fig:Gaze360}
\end{figure}

In \cite{kellnhofer2019gaze360}, Kellnhofer et al. presented Gaze360, a large-scale gaze-tracking dataset and method for robust 3D gaze estimation in unconstrained images. Their dataset consists of 238 subjects in indoor and outdoor environments with labeled 3D gaze across a wide range of head poses and distances. It was the largest publicly available dataset of its kind by both subject and variety, at the time. Some of the sample images from this dataset are shown in Fig \ref{fig:Gaze360}.
They also proposed a 3D gaze model that extended existing models to include temporal information and to directly output an estimate of gaze uncertainty.

\begin{figure}[h]
\centering
\includegraphics[width=0.99\linewidth]{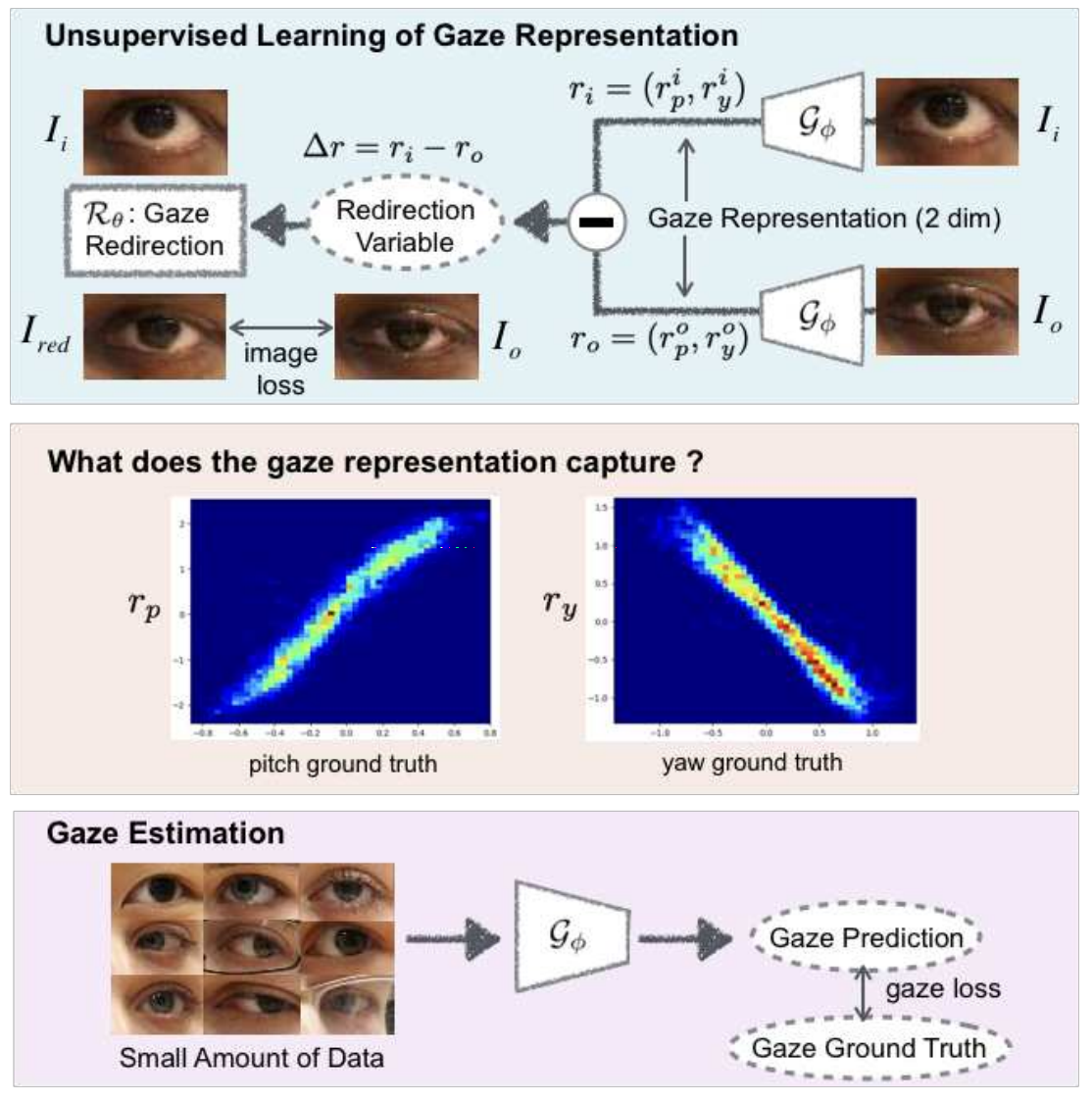}
\caption{The proposed framework for Unsupervised Representation Learning for Gaze Estimation. Courtesy of \cite{yu2020unsupervised}}
\label{fig:gaze_unsuper}
\end{figure}

In \cite{yu2020unsupervised}, Yu and Odobez proposed an effective approach to learn a low dimensional gaze representation without gaze annotations.
The main idea is to rely on a gaze redirection network and use the gaze representation difference of the input and target images (of the redirection network) as the redirection variable. A redirection loss in image domain allows the joint training of both the redirection network and the gaze representation network. In addition, they propose a warping field regularization which not only provides an explicit physical meaning to the gaze representations but also avoids redirection distortions.
The high level architecture of this framework is shown in 
Fig \ref{fig:gaze_unsuper}.

\begin{figure}[h]
\centering
\includegraphics[width=0.99\linewidth]{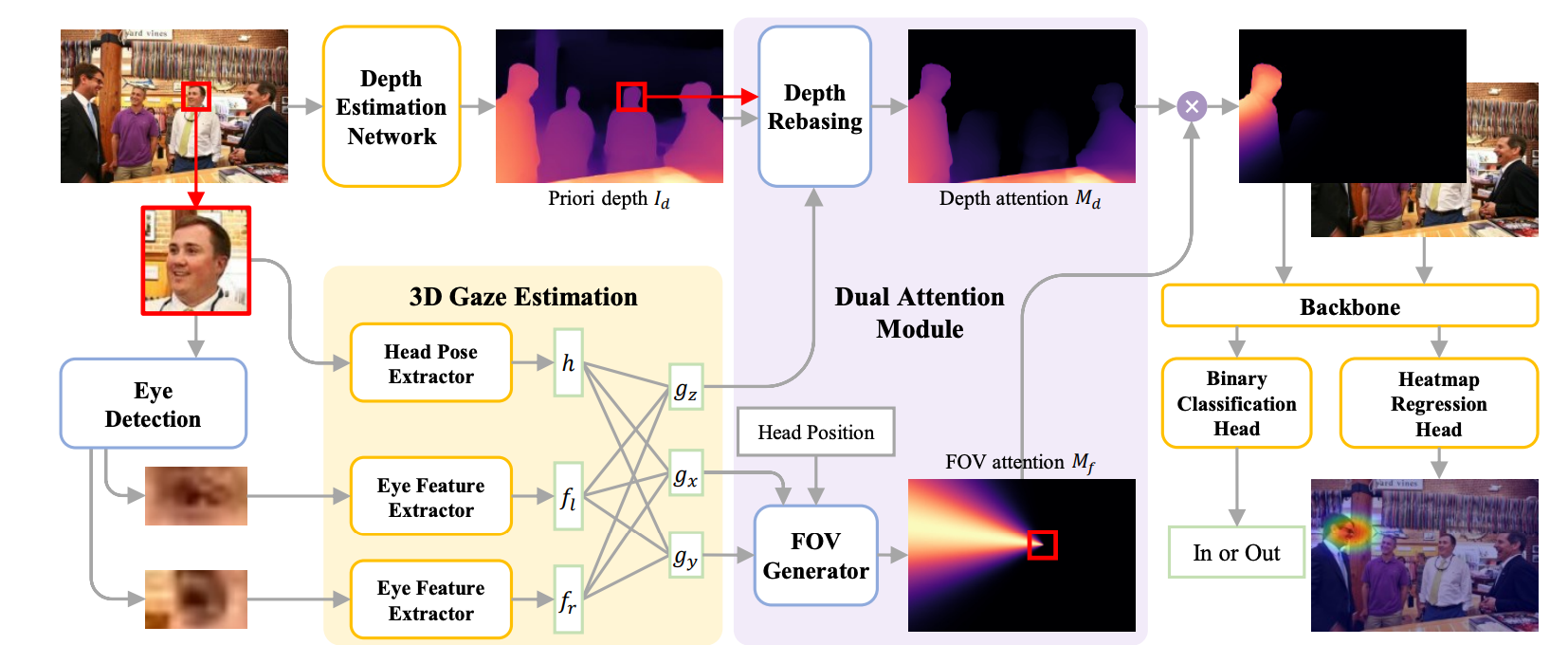}
\caption{The Architecture of Dual Attention Guided Gaze Target Detection. Courtesy of \cite{fang2021dual}}
\label{fig:dual_gaze}
\end{figure}

In \cite{fang2021dual}, Fang et al. proposed a three-stage method to simulate the human gaze inference behavior in 3D space. In the first stage, they introduced a coarse-to-fine strategy to robustly estimate a 3D gaze orientation from the head. The predicted gaze is decomposed into a planar gaze on the image plane and a depth channel gaze. In the second stage, they develop a Dual Attention Module (DAM), which takes the planar gaze to produce the field of view and masks interfering objects regulated by depth information according to the depth-channel gaze. In the third stage, they use the generated dual attention as guidance to perform two sub-tasks: (1) identifying whether the gaze target is inside or out of the image; (2) locating the target if inside.
The architecture of this model is shown in Fig \ref{fig:dual_gaze}.

Some of the other works for eye tracking and gaze estimation includes: Few-shot adaptive gaze estimation \cite{park2019few}, TH-XGaze: A large scale dataset for gaze estimation under extreme head pose and gaze variation \cite{zhang2020eth},towards end-to-end video-based eye-tracking \cite{park2020towards}, weakly-supervised physically unconstrained gaze estimation \cite{kothari2021weakly}.

\subsubsection{Hand Tracking and Pose Estimation}
In \cite{oberweger2015hands}, Oberweger et al. introduced and evaluated several architectures
for Convolutional Neural Networks to predict the 3D
joint locations of a hand given a depth map. 
They introduced a prior on the 3D pose and significantly improved the accuracy and reliability of the predictions. They  also showed how to use context efficiently to deal with ambiguities between fingers.

\begin{figure}[h]
\centering
\includegraphics[width=0.99\linewidth]{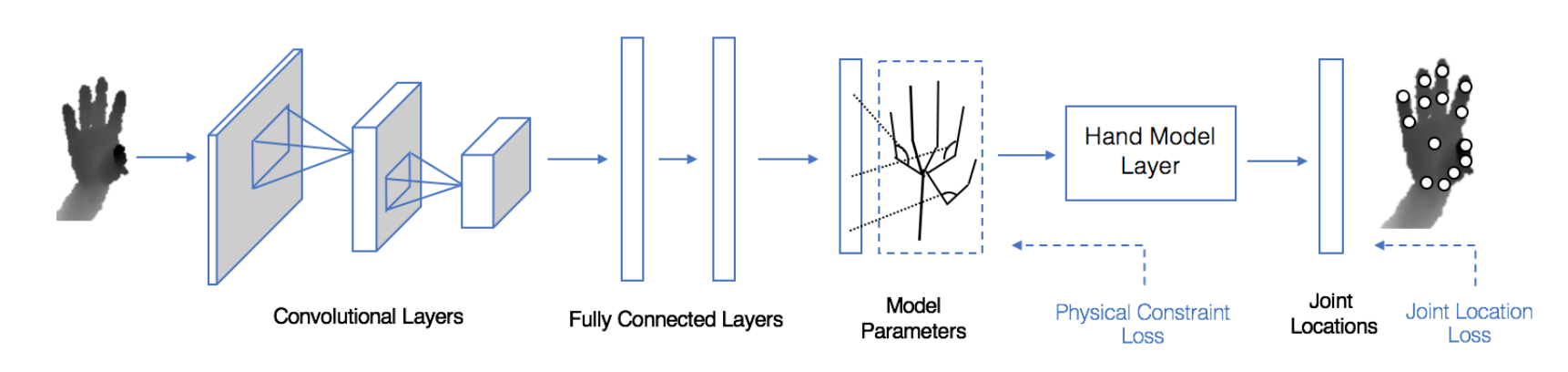}
\caption{The architecture of model based deep hand pose learning. Courtesy of \cite{zhou2016model}}
\label{fig:zhou2016model}
\end{figure}

In \cite{zhou2016model}, Zhou et al. proposed
a model based deep learning approach that adopts
a forward kinematics based layer to ensure the geometric validity of estimated poses. 
After applying standard convolutional and fully connected layers, the hand model pose parameters (mostly joint angles) are predicted. Then a new hand model layer maps the pose parameters to the hand joint locations via a forward kinematic process. The architecture of this framework is shown in Fig \ref{fig:zhou2016model}.

\begin{figure}[h]
\centering
\includegraphics[width=0.99\linewidth]{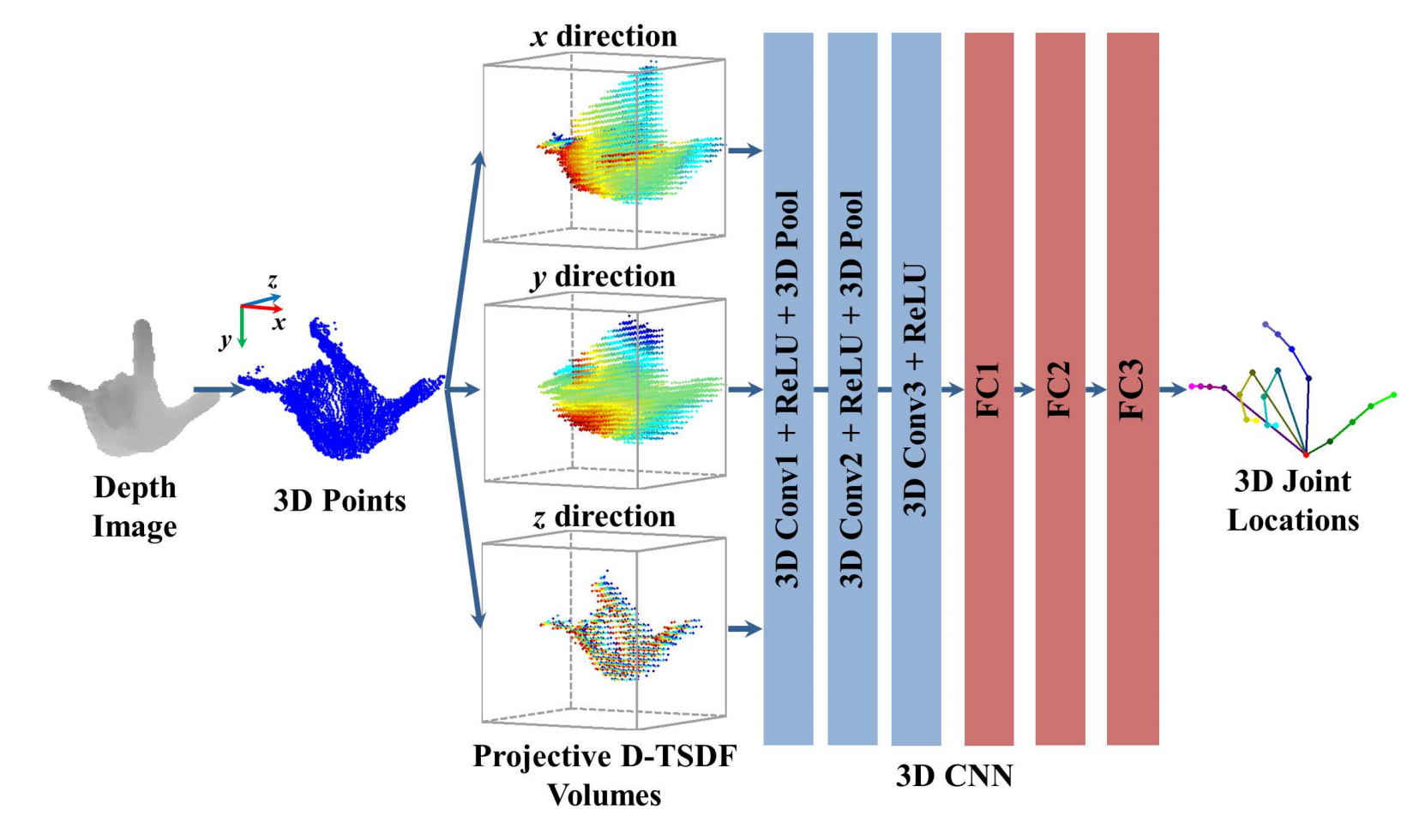}
\caption{The architecture of m3d CNN model for hand pose estimation. Courtesy of \cite{ge20173d}}
\label{fig:hand_3dcnn}
\end{figure}

In \cite{ge20173d}, Ge et al.  proposed a simple, yet effective approach for real-time hand pose estimation from single depth images using 3D CNNs.
Their proposed 3D CNN taking a 3D volumetric representation of the hand depth image as input
can capture the 3D spatial structure of the input and accurately regress full 3D hand pose in a single pass.
The architecture of the proposed 3D CNN by this work is shown in Fig \ref{fig:hand_3dcnn}.

\begin{figure}[h]
\centering
\includegraphics[width=0.99\linewidth]{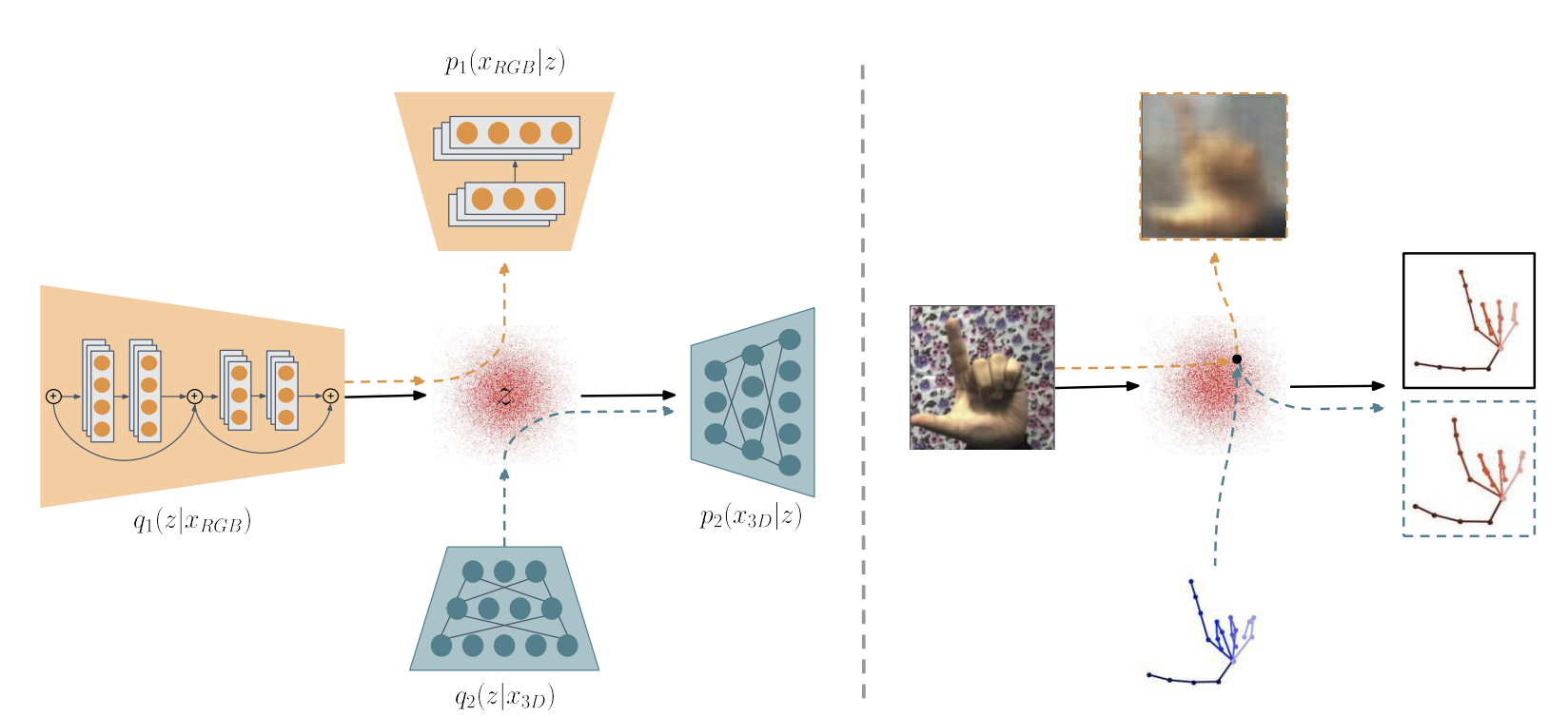}
\caption{Schematic overview of the cross-modal deep variational model. Left: a cross-modal latent space z is learned by training pairs of encoder and decoder q, p networks across multiple modalities (e.g., RGB images to 3D hand poses). Auxiliary encoder-decoder pairs help in regularizing the latent space. Right: The approach allows to embed input samples of one set of modalities (here: RGB, 3D) and to produce consistent and plausible posterior estimates in several different modalities (RGB, 2D and 3D). Courtesy of \cite{spurr2018cross}}
\label{fig:crossmodel}
\end{figure}

In \cite{spurr2018cross}, Spurr et al. proposed a method to learn a statistical hand model represented by a cross-modal trained latent space via a generative deep neural network. They derived an objective function from the variational lower bound of the VAE framework and jointly optimize the resulting cross-modal KL-divergence and the posterior reconstruction objective, naturally admitting a training regime that leads to a coherent latent space across multiple modalities such as RGB images, 2D keypoint  detection or 3D hand configurations.
Additionally, it grants a straightforward way of using semi-supervision. This latent space can be directly used to estimate 3D hand poses from RGB images, outperforming the
state-of-the art in different settings.
The high-level architecture of this framework is shown in 
Fig \ref{fig:crossmodel}.

\begin{figure}[h]
\centering
\includegraphics[width=0.99\linewidth]{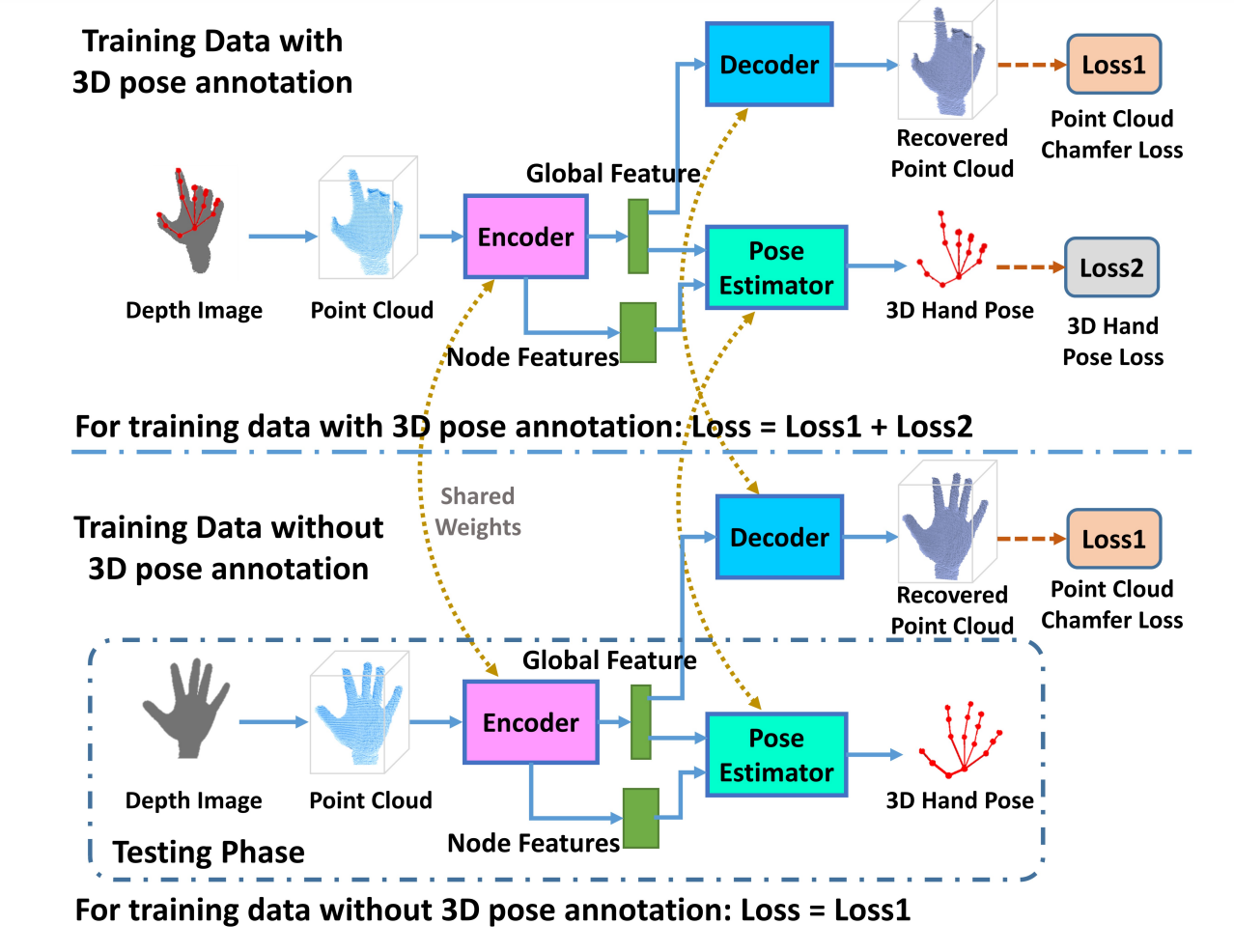}
\caption{Overview of the proposed SO-HandNet framework. Courtesy of \cite{chen2019so}}
\label{fig:SOHandNet}
\end{figure}

In \cite{chen2019so}, inspired by the point cloud
autoencoder presented in self-organizing network (SO-Net)
, Chen et al. proposed SO-HandNet which aimed at making use of the unannotated data to obtain accurate 3D hand pose estimation in a semi-supervised manner. We exploit hand feature encoder (HFE) to extract multi-level features from hand point cloud and then fuse them to regress 3D hand pose
by a hand pose estimator (HPE). We design a hand feature decoder (HFD) to recover the input point cloud from
the encoded feature. The overview of the model architecture of this work is shown in Fig \ref{fig:SOHandNet}.

\begin{figure}[h]
\centering
\includegraphics[width=0.99\linewidth]{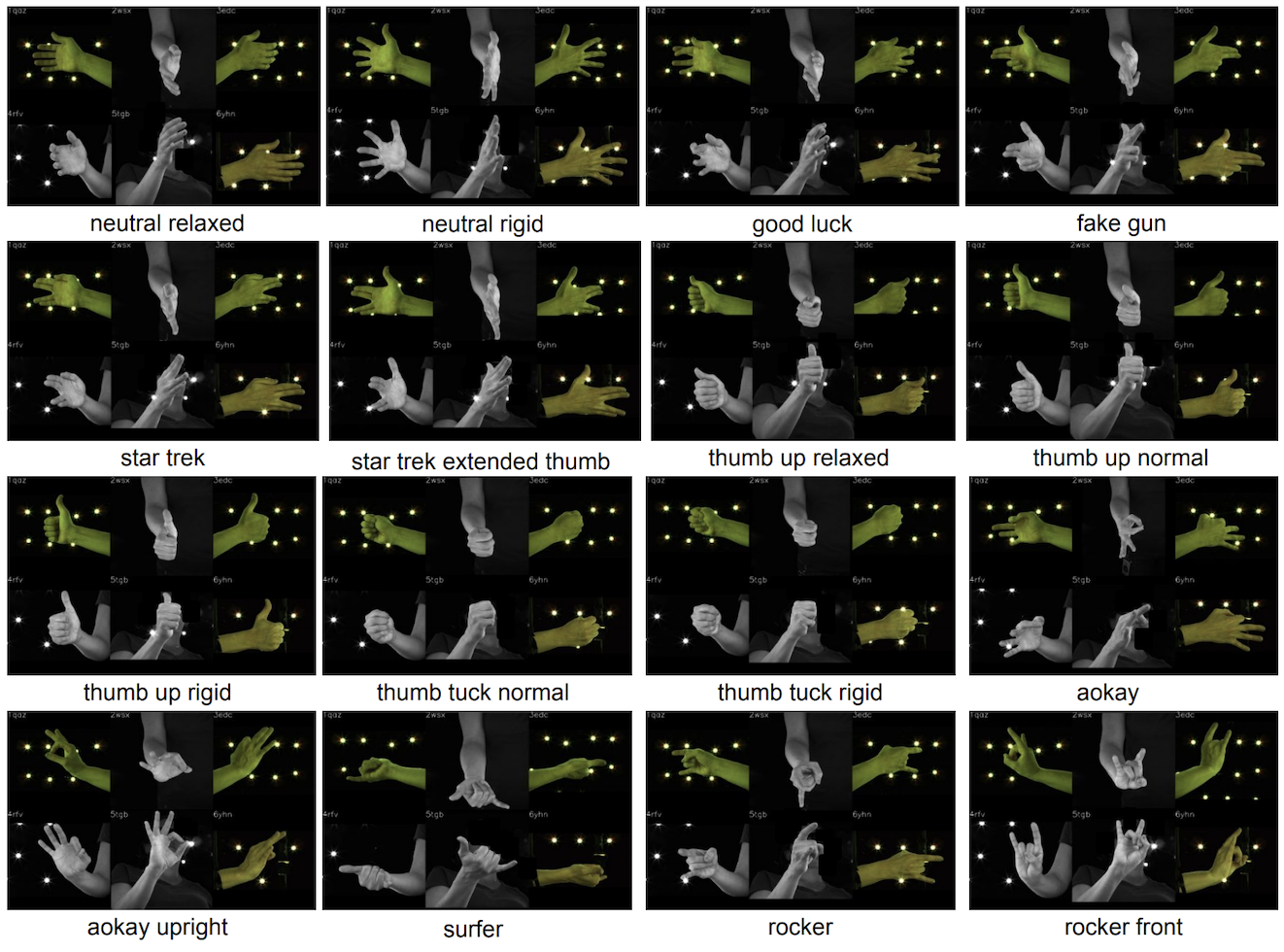}
\caption{Visualization of the single hand PP sequences from InterHand26M dataset. Courtesy of \cite{moon2020interhand2}}
\label{fig:InterHand_data}
\end{figure}

In \cite{moon2020interhand2}, Moon et al. introduced a  a large-scale dataset, called InterHand2.6M, which contains 2.6M labeled single and interacting hand frames under various poses from multiple subjects. They also proposed a baseline network, InterNet, for 3D interacting hand pose estimation from a single RGB image. InterNet simultaneously performs 3D single and interacting hand pose estimation. 
Some of the sample frames from sequences with single hand are shown in Fig \ref{fig:InterHand_data}.

In \cite{caramalau2021active}, Caramalau et al. proposed a Bayesian approximation to a deep learning architecture for 3D hand pose estimation. Through this framework, they explored and analysed the two types of uncertainties that are influenced either by data or by the learning capability. Furthermore, they drew comparisons against the standard estimator over three popular benchmarks.

Some of the other works for hand tracking and pose estimation includes: Spatial attention deep net for hand pose estimation \cite{ye2016spatial}, Deepprior++ \cite{oberweger2017deepprior}, Point-to-point regression pointnet for 3D hand pose estimation \cite{ge2018point}, Hand-transformer: non-autoregressive structured modeling for 3D hand pose estimation \cite{huang2020hand}, 3D Hand Pose Estimation via aligned latent space injection and kinematic losses \cite{stergioulas20213d}.

\subsubsection{Human Pose Estimation and Tracking}
\begin{figure}[h]
\centering
\includegraphics[width=0.99\linewidth]{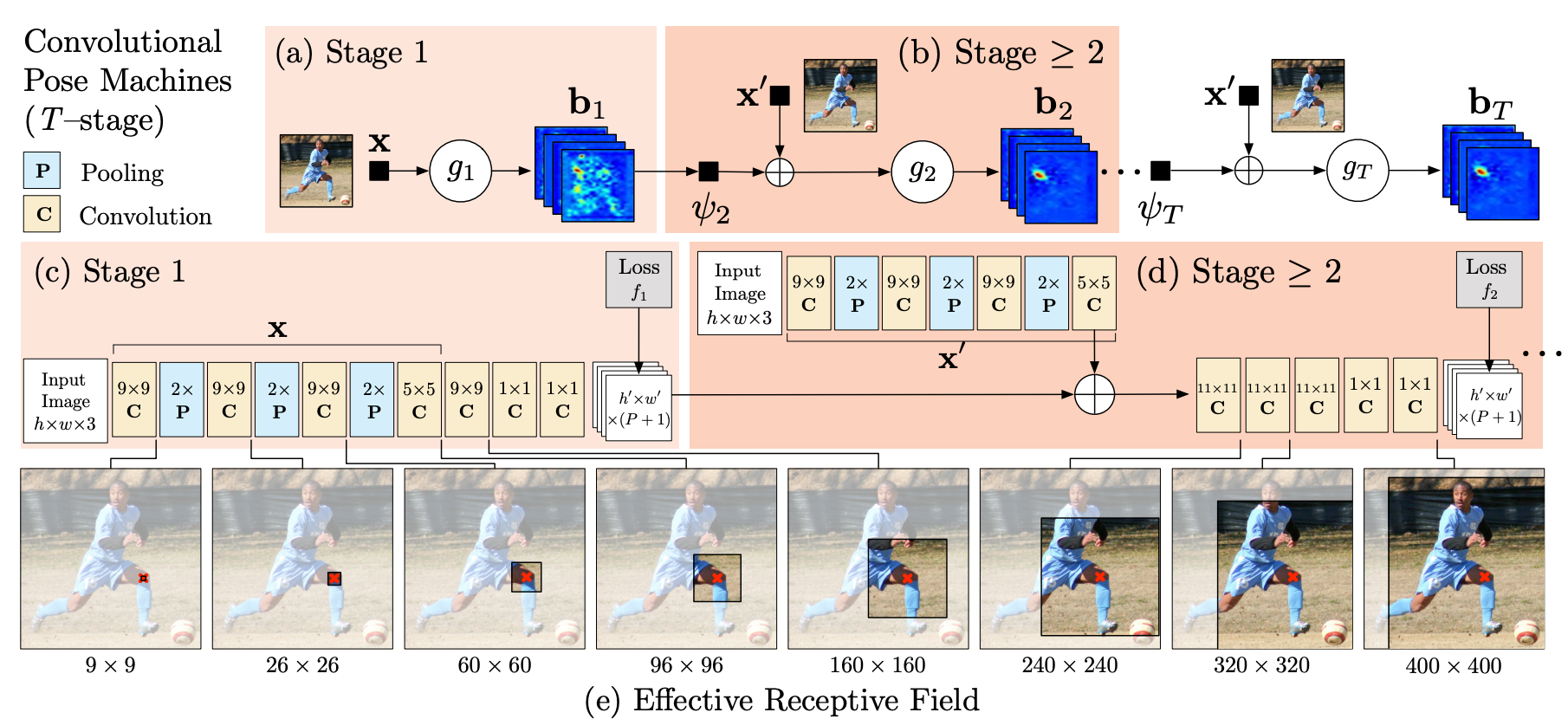}
\caption{Architecture and receptive fields of CPMs. Courtesy of \cite{wei2016convolutional}}
\label{fig:CPM}
\end{figure}

In \cite{wei2016convolutional}, Wei et al. showed a systematic design for how convolutional networks can be incorporated into the pose machine framework for learning image features and image-dependent spatial models for the task of pose estimation.
They implicitly model long-range dependencies between variables in structured prediction tasks such
as articulated pose estimation.
Fig \ref{fig:CPM} shows the high level architecture of the proposed CPM framework.

\begin{figure}[h]
\centering
\includegraphics[width=0.99\linewidth]{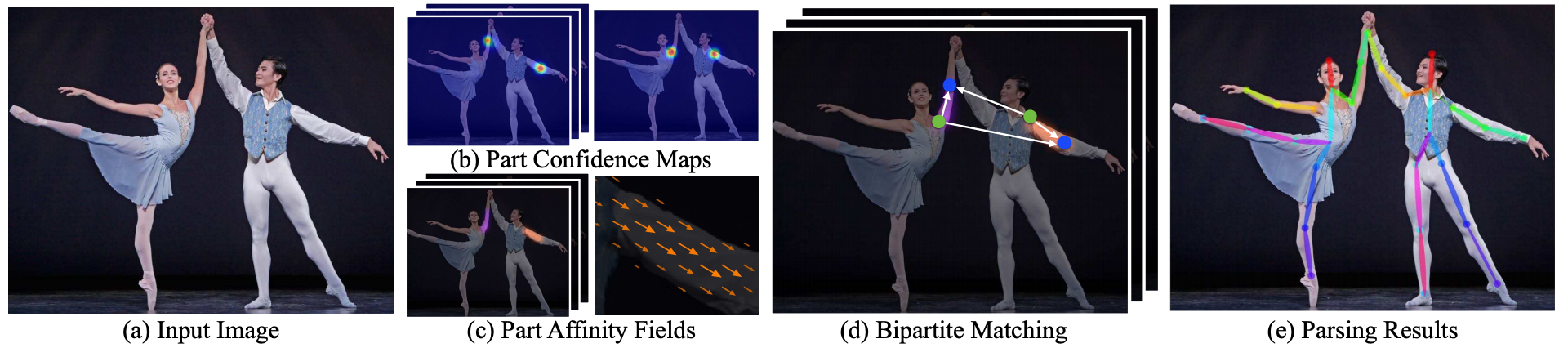}
\caption{Overall pipeline. Our method takes the entire image as the input for a two-branch CNN to jointly predict confidence maps for body part detection, shown in (b), and part affinity fields for parts association, shown in (c). The parsing step performs a set of bipartite matchings to associate body parts candidates (d). We finally assemble them into full body poses for all people in the image (e). Courtesy of \cite{cao2017realtime}}
\label{fig:PAFs_pipeline}
\end{figure}

In \cite{cao2017realtime}, Cao et al. proposed a real-time multi-person 2D pose estimation using part affinity fields.
This approach uses a non-parametric representation, which they referred to as Part Affinity Fields (PAFs), to learn to associate body parts with individuals in the image. The architecture encodes global context, allowing a greedy bottom-up parsing step that maintains high accuracy while achieving real-time performance, irrespective of the number of people in the image.  The architecture is designed to jointly learn part locations and their association via two branches of the same sequential prediction process.
The overall pipeline of this framework is shown in Fig \ref{fig:PAFs_pipeline}.

\begin{figure}[h]
\centering
\includegraphics[width=0.99\linewidth]{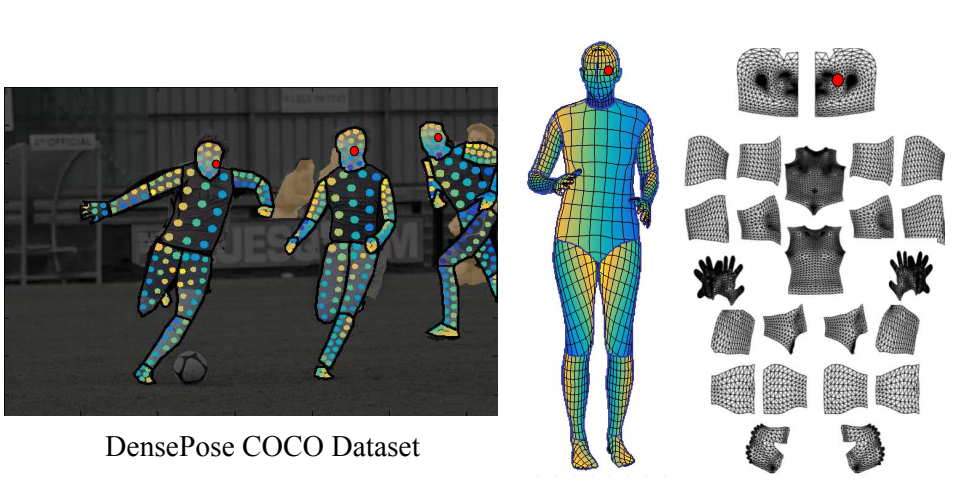}
\caption{DensePose-COCO Dataset annotations.
Right: Partitioning and UV parametrization of the body surface. Courtesy of \cite{guler2018densepose}}
\label{fig:densepose_data}
\end{figure}

In \cite{guler2018densepose}, Guler et al. proposed DensePose, which establishes dense correspondences between an RGB image and a surface-based representation
of the human body. They gathered dense correspondences for 50K persons appearing in the COCO dataset by introducing an efficient annotation pipeline. The annotations of one sample image from this dataset is shown in Fig \ref{fig:densepose_data}.
They then used this dataset to train CNN-based systems that deliver dense correspondence ‘in the wild’, namely in the presence of background, occlusions and scale variations.

In \cite{cao2019openpose}, Pavllo et al. proposed a 3D human pose estimation in video with temporal convolutions and semi-supervised training. 
They demonstrated that 3D poses in video can be effectively estimated with a fully convolutional model based on dilated temporal convolutions over 2D keypoints. They also introduced back-projection, a simple and effective semi-supervised training method that leverages unlabeled video data.
They started with predicted 2D keypoints for unlabeled video, then estimated 3D poses and finally back-project to the input 2D keypoints.

\begin{figure}[h]
\centering
\includegraphics[width=0.99\linewidth]{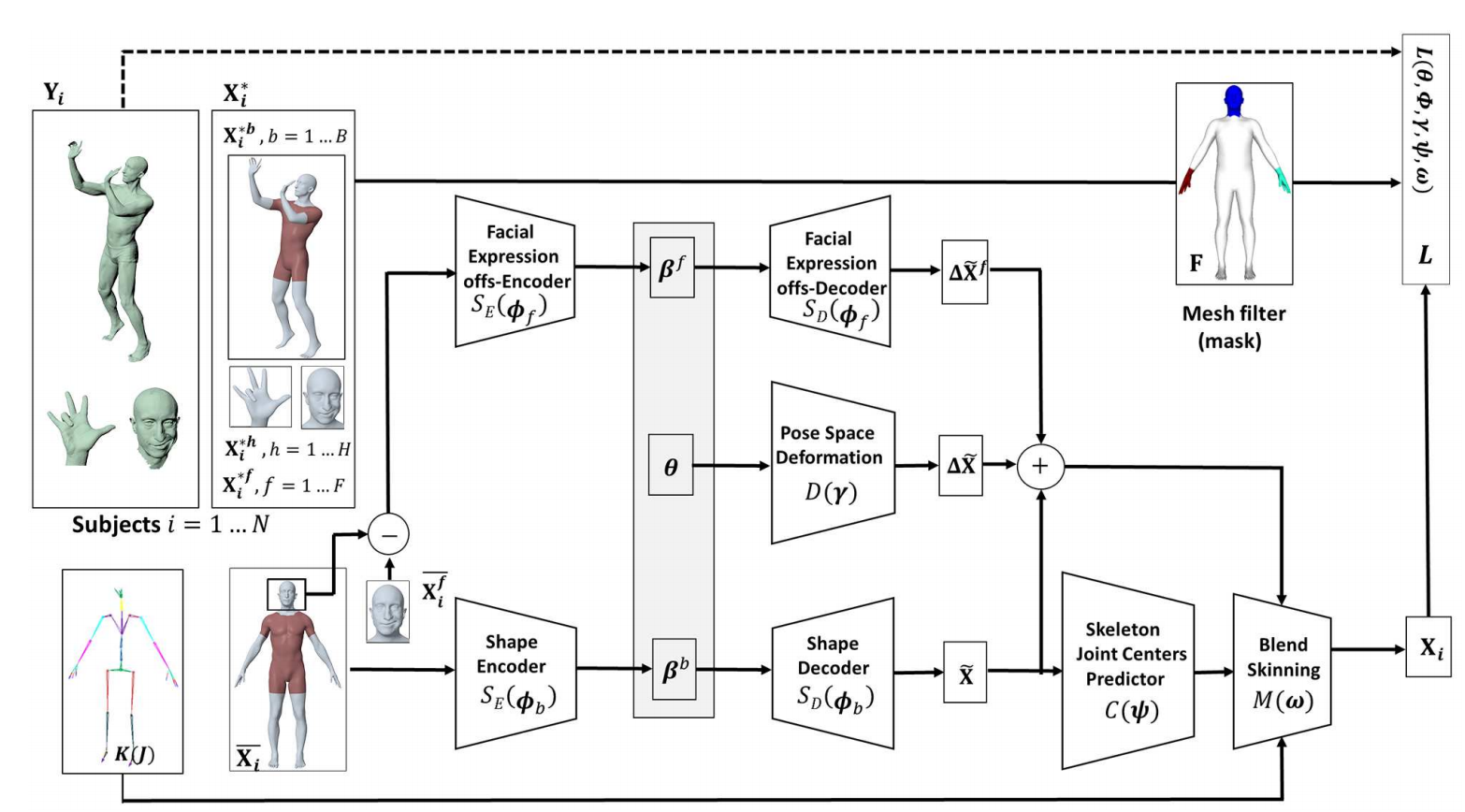}
\caption{Overview of our end-to-end statistical 3D articulated human shape model construction. Courtesy of \cite{xu2020ghum}}
\label{fig:xu2020ghum}
\end{figure}

In \cite{xu2020ghum}, Xu et al. presented a statistical, articulated 3D human shape modeling pipeline, within a fully trainable, modular, deep learning framework. Given high-resolution complete 3D body scans of humans, captured in various poses, together with additional closeups of their head and facial expressions, as well as hand articulation, and given initial, artist designed, gender neutral rigged quad-meshes, they trained all
model parameters including non-linear shape spaces based
on variational auto-encoders, pose-space deformation correctives, skeleton joint center predictors, and blend skinning functions, in a single consistent learning loop.
The models are simultaneously trained with all the 3D dynamic scan data (over 60, 000 diverse human configurations in our new dataset) in order to capture correlations and ensure consistency of various components.
The high-level overview of this framework is shown in Fig  \ref{fig:xu2020ghum}.

In \cite{liu2020keypose}, Liu et al. proposed a Multi-View 3D Labeling and Keypoint Estimation for Transparent Objects, called KeyPose. They forwent using a depth sensor in favor of raw stereo input.
They tried to address two problems:
First, they established an easy method for capturing and labeling 3D keypoints on desktop objects with an RGB camera; Second, they developed a deep neural network, called KeyPose, that learns to accurately predict object poses using 3D keypoints, from stereo input, and works even for transparent objects. 
They also created a dataset of 15 clear objects in five classes, with 48K 3D-keypoint labeled images.

\begin{figure}[h]
\centering
\includegraphics[width=0.99\linewidth]{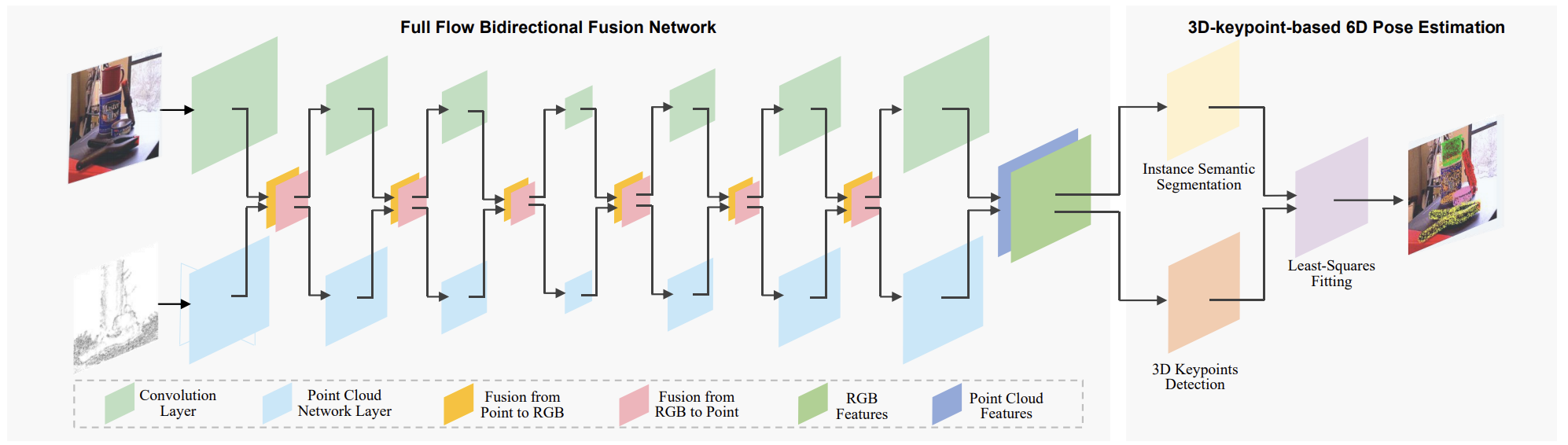}
\caption{The pipeline of FFB6D. A CNN and a point cloud network is utilized for representation learning of RGB image and point cloud respectively. In flow of the two networks, bidirectional fusion modules are added as communicate bridges. The extracted per-point features are then fed into an instance semantic segmentation and a 3D keypoint voting modules to obtain per-object 3D keypoints. Finally, the pose is recovered within a least-squares fitting algorithm. Courtesy of \cite{he2021ffb6d}}
\label{fig:FFB6D_arch}
\end{figure}

In \cite{he2021ffb6d}, He et al. presented FFB6D, a Full Flow Bidirectional fusion network designed for 6D pose estimation from a single RGB-D image. Their key insight is that appearance information in the RGB image and geometry
information from the depth image are two complementary
data sources, and it still remains unknown how to fully
leverage them. Towards this end, FFB6D is proposed, which learns to combine appearance and geometry information for representation learning as well as output representation selection. 
Specifically, at the representation learning
stage, they built bidirectional fusion modules in the full
flow of the two networks, where fusion is applied to each
encoding and decoding layer. In this way, the two networks can leverage local and global complementary information from the other one to obtain better representations.
The high-level overview of FFB6D framework is shown in Fig \ref{fig:FFB6D_arch}.

Some of the other popular frameworks for human pose estimation includes: regional multi-person pose estimation \cite{fang2017rmpe}, simple baselines for human pose estimation and tracking \cite{xiao2018simple}, OpenPose: Realtime Multi-Person 2D Pose Estimation using Part Affinity Fields \cite{cao2019openpose}, SimPoE: Simulated Character Control for 3D Human Pose Estimation \cite{yuan2021simpoe}.



\subsection{Geometry Applications}

Deep learning models developed for vision geometry are important for various AR applications (such as the ones in Games, Museums, Automotive, and Scene Understanding). There are various works developed in this direction. Here we cover some of the prominent works.

\begin{figure}[h]
\centering
\includegraphics[width=0.99\linewidth]{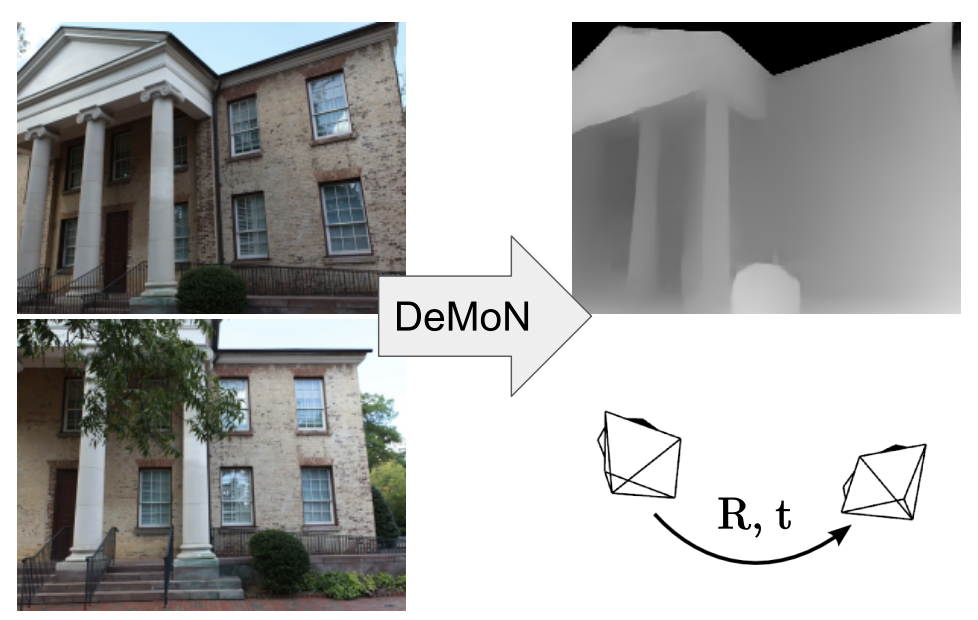}
\caption{Illustration of DeMoN. The input to the network is two
successive images from a monocular camera. The network estimates the depth in the first image and the camera motion. Courtesy of \cite{ummenhofer2017demon}}
\label{fig:DeMoN}
\end{figure}

In \cite{ummenhofer2017demon}, Ummenhofer et al. proposed a depth and motion Network for Learning Monocular Stereo, so called DeMoN. They formulated structure from motion as a learning problem. This network estimates not only depth and motion, but additionally surface normals, optical flow between the images and confidence of the matching.
Fig \ref{fig:DeMoN} shows a sample result of the predicted depth map by DeMon.

\begin{figure}[h]
\centering
\includegraphics[width=0.99\linewidth]{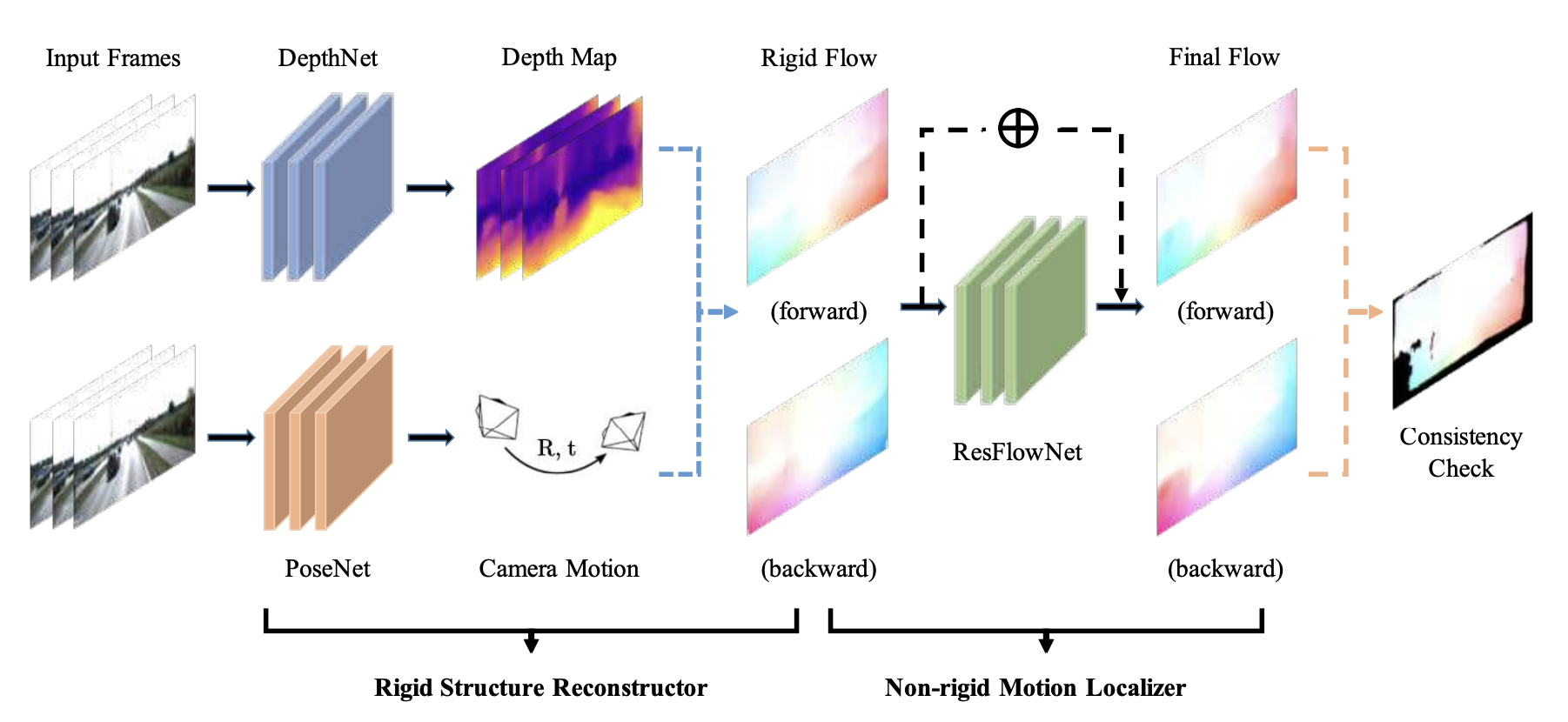}
\caption{The overview of GeoNet Framework. It consists of rigid structure reconstructor for estimating static scene geometry and non-rigid motion localizer for capturing dynamic objects. Courtesy of \cite{yin2018geonet}}
\label{fig:geonet}
\end{figure}

In \cite{yin2018geonet}, Yin et al. proposed GeoNet, a jointly unsupervised learning framework for monocular depth, optical flow and egomotion estimation from videos. The three components are
coupled by the nature of 3D scene geometry, jointly learned
by our framework in an end-to-end manner. 
Fig \ref{fig:geonet} shows the overview of the GeoNet framework.

In \cite{gordon2019depth}, Gordon et al. present a novel method for simultaneously learning depth, ego motion, object motion, and camera intrinsics from monocular videos, using only consistency across
neighboring video frames as a supervision signal. They addressed occlusions geometrically and differentiably, directly
using the depth maps as predicted during training.

In \cite{guizilini20203d}, Guizilini et al. proposed a novel self-supervised monocular depth estimation method combining geometry
with a new deep network, PackNet, learned only from unlabeled monocular videos. Their architecture leverages novel symmetrical packing and unpacking blocks to jointly learn to compress and decompress detail-preserving representations using 3D convolutions.  The 3D inductive bias in PackNet enables it to scale with input resolution and number of parameters without overfitting, generalizing better on
out-of-domain data.

\begin{figure}[h]
\centering
\includegraphics[width=0.99\linewidth]{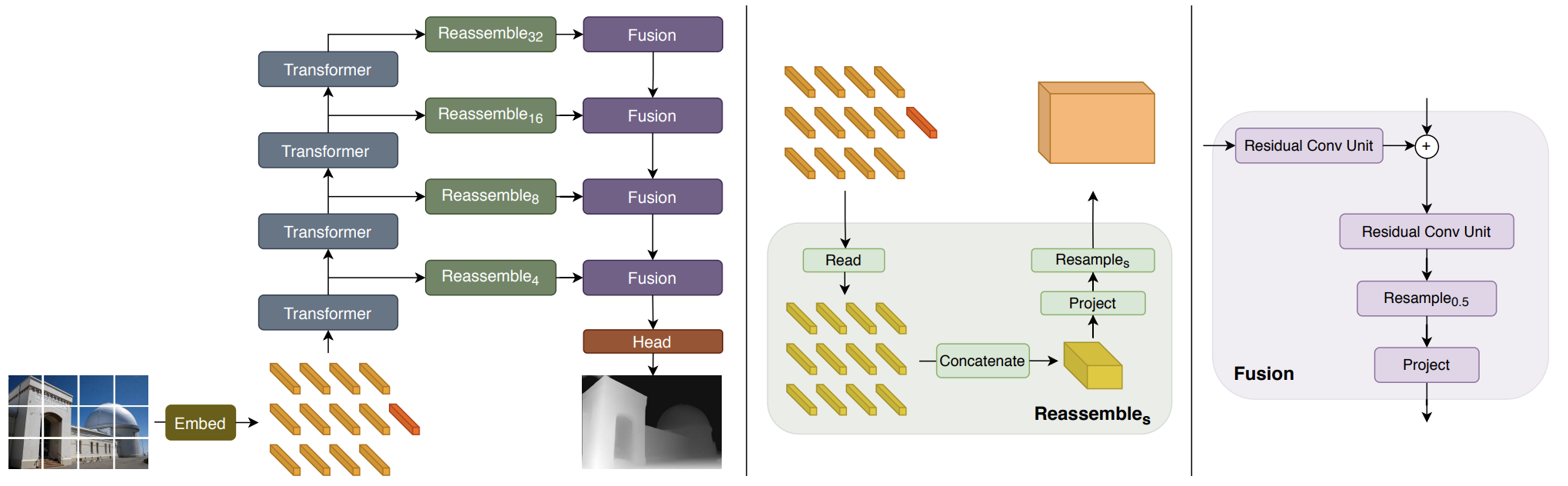}
\caption{The  overview of dense prediction transformer framework. Courtesy of \cite{ranftl2021vision}}
\label{fig:dpt}
\end{figure}

In \cite{ranftl2021vision}, Ranftl et al. introduced dense prediction transformers, an architecture that leverages vision transformers in place of convolutional networks as a backbone, for dense prediction
tasks. They assemble tokens from various stages of the vision transformer into image-like representations at various resolutions and progressively combine them into full resolution predictions using a convolutional decoder.  For monocular depth estimation,
there is an improvement of up to 28\% in relative
performance when compared to a state-of-the-art fully convolutional network. 
Fig \ref{fig:dpt} shows the overview of the proposed dense prediction transformers framework.

Some of the other representative works in this area includes: Unsupervised learning of depth and ego-motion from video \cite{zhou2017unsupervised}, MegaDepth \cite{li2018megadepth}, and TransformerFusion \cite{bozic2021transformerfusion}.

\subsection{Scene Understanding and Reconstruction}
Simultaneous Localization and Mapping (SLAM) denotes the computational technique that creates and updates a map of an unknown space where a robot agent is located, while simultaneously tracking the agent’s location in it. It is a crucial step in many of the AR/MR, and also robotic applications.

In \cite{dai2017scannet}, Dai et al.  introduced ScanNet, an
RGB-D video dataset containing 2.5M views in 1513 scenes
annotated with 3D camera poses, surface reconstructions,
and semantic segmentations. To collect this data, they designed an easy-to-use and scalable RGB-D capture system
that includes automated surface reconstruction and crowd-sourced semantic annotation.
They showed that using this data helps achieve state-of-the-art performance on several 3D scene understanding tasks, including 3D object classification, semantic voxel labeling, and CAD model retrieval.

In \cite{zhang2017mixedfusion}, Zhang et al.  developed an end-to-end system using a depth sensor to scan a scene on the fly. By proposing a Sigmoid-based Iterative Closest Point (S-ICP) method, they decouple the camera motion and the scene motion from the input sequence and segment the scene into static and dynamic parts accordingly. The static part is used to estimate the camera rigid motion, while for the dynamic part, graph node-based motion representation and model-to-depth fitting are applied to reconstruct the scene motions. With the camera and scene motions reconstructed, they further proposed a novel mixed voxel allocation scheme to handle static and
dynamic scene parts with different mechanisms, which helps to gradually fuse a large scene with both static and dynamic objects.

\begin{figure}[h]
\centering
\includegraphics[width=0.99\linewidth]{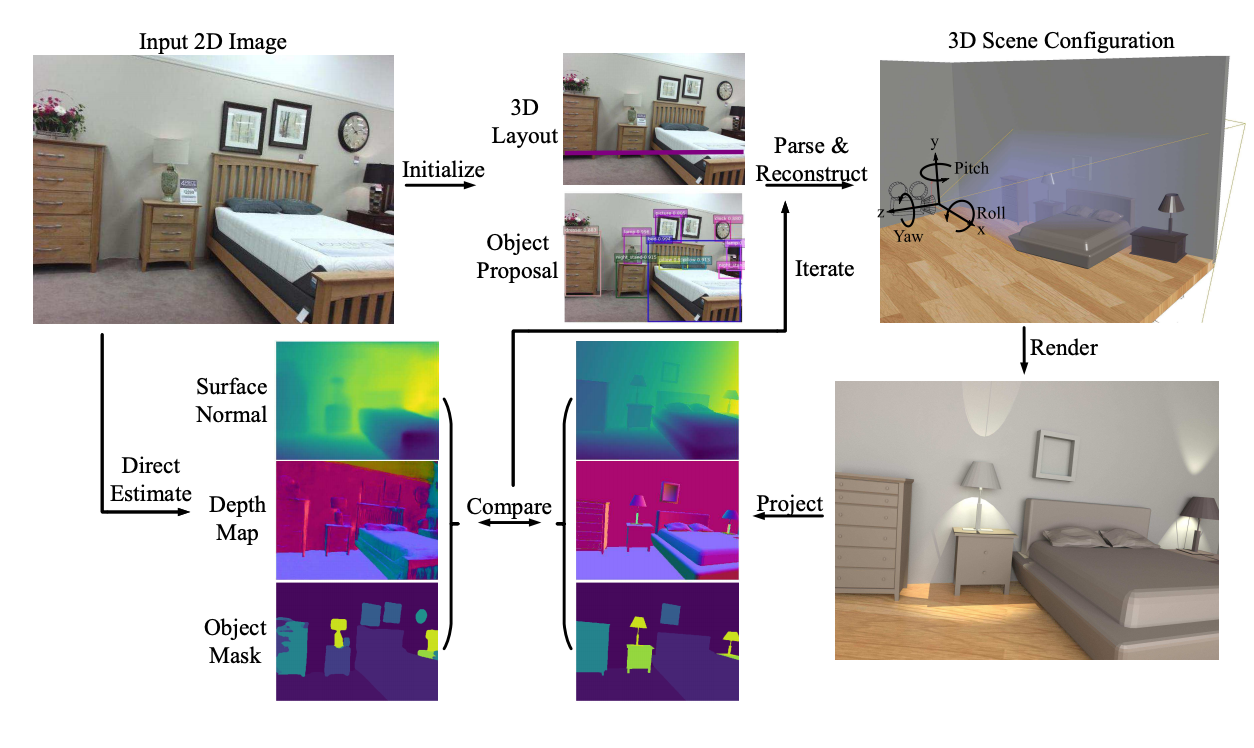}
\caption{Illustration of the proposed holistic 3D indoor scene parsing and reconstruction in an analysis-by synthesis fashion. Courtesy of \cite{huang2018holistic}}
\label{fig:holistic_recons}
\end{figure}

In \cite{huang2018holistic}, Huang et al. proposed a computational framework to jointly parse a single RGB image and reconstruct a holistic 3D configuration composed by a set of CAD models using a stochastic grammar model. Specifically, they introduced a Holistic Scene Grammar (HSG) to represent the 3D scene structure, which characterizes a joint distribution over the functional and geometric space of indoor scenes. The proposed HSG captures three essential and often latent dimensions of the indoor scenes: i) latent human context, describing the affordance and the functionality of a room arrangement, ii) geometric constraints over the scene configurations, and iii) physical constraints that guarantee physically plausible parsing and reconstruction.
They solved this joint parsing and reconstruction problem
in an analysis-by-synthesis fashion, seeking to minimize the differences between the input image and the rendered images generated by our 3D representation, over the space of depth, surface normal, and object segmentation map.
Fig \ref{fig:holistic_recons} illustrates the overview of the proposed holistic 3D indoor scene parsing and reconstruction framework.

\begin{figure}[h]
\centering
\includegraphics[width=0.99\linewidth]{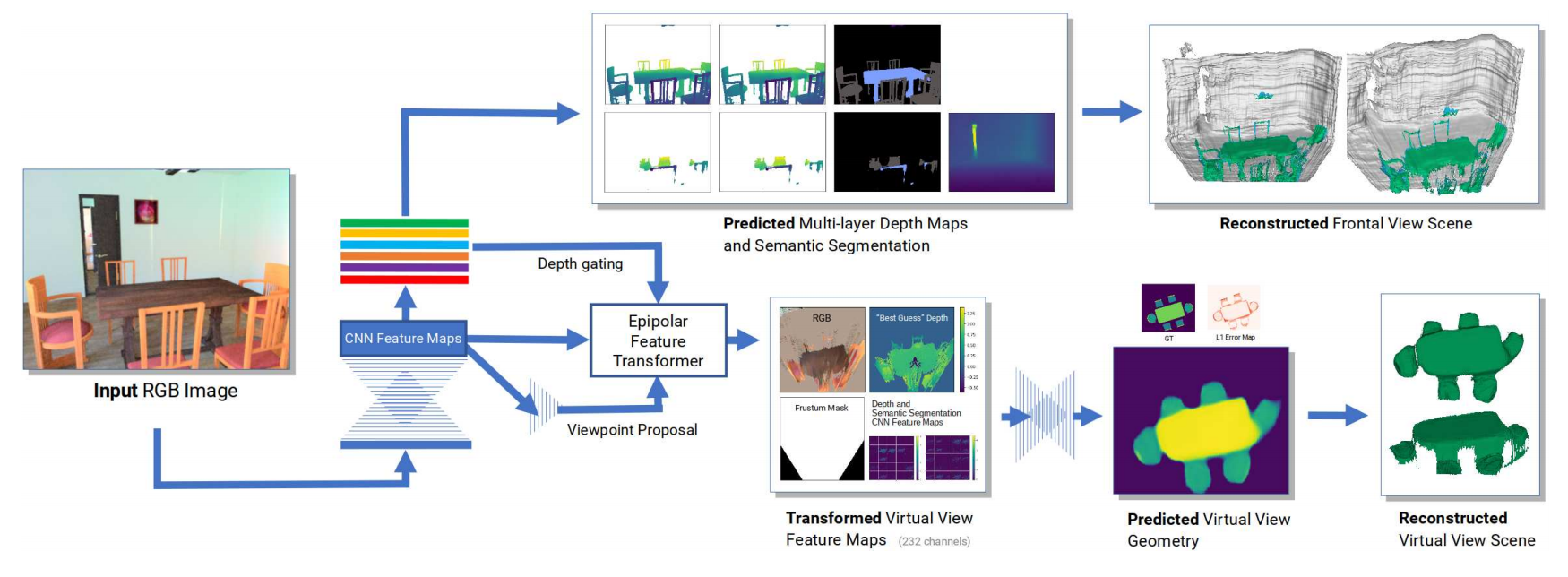}
\caption{Illustration of the proposed 3D scene reconstruction using Multi-layer Depth and Epipolar Transformers. Courtesy of \cite{shin20193d}}
\label{fig:shin20193d}
\end{figure}

In \cite{shin20193d}, Shin et al.  tackled the problem of automatically reconstructing a complete 3D model of a scene from a single RGB image. Their approach utilizes viewer-centered, multi-layer representation of scene geometry adapted from recent methods for single object shape completion. To improve the accuracy of view-centered representations for complex scenes, they introduced a novel “Epipolar Feature Transformer” that transfers convolutional network features from an input view to other virtual camera viewpoints, and thus better covers the 3D scene geometry. 
Unlike previous approaches that first detect and
localize objects in 3D, and then infer object shape using category-specific models, their approach is fully convolutional, end-to-end differentiable, and avoids the resolution and memory limitations of voxel representations.
Fig \ref{fig:shin20193d} illustrates the overview of the proposed framework.

\begin{figure}[h]
\centering
\includegraphics[width=0.99\linewidth]{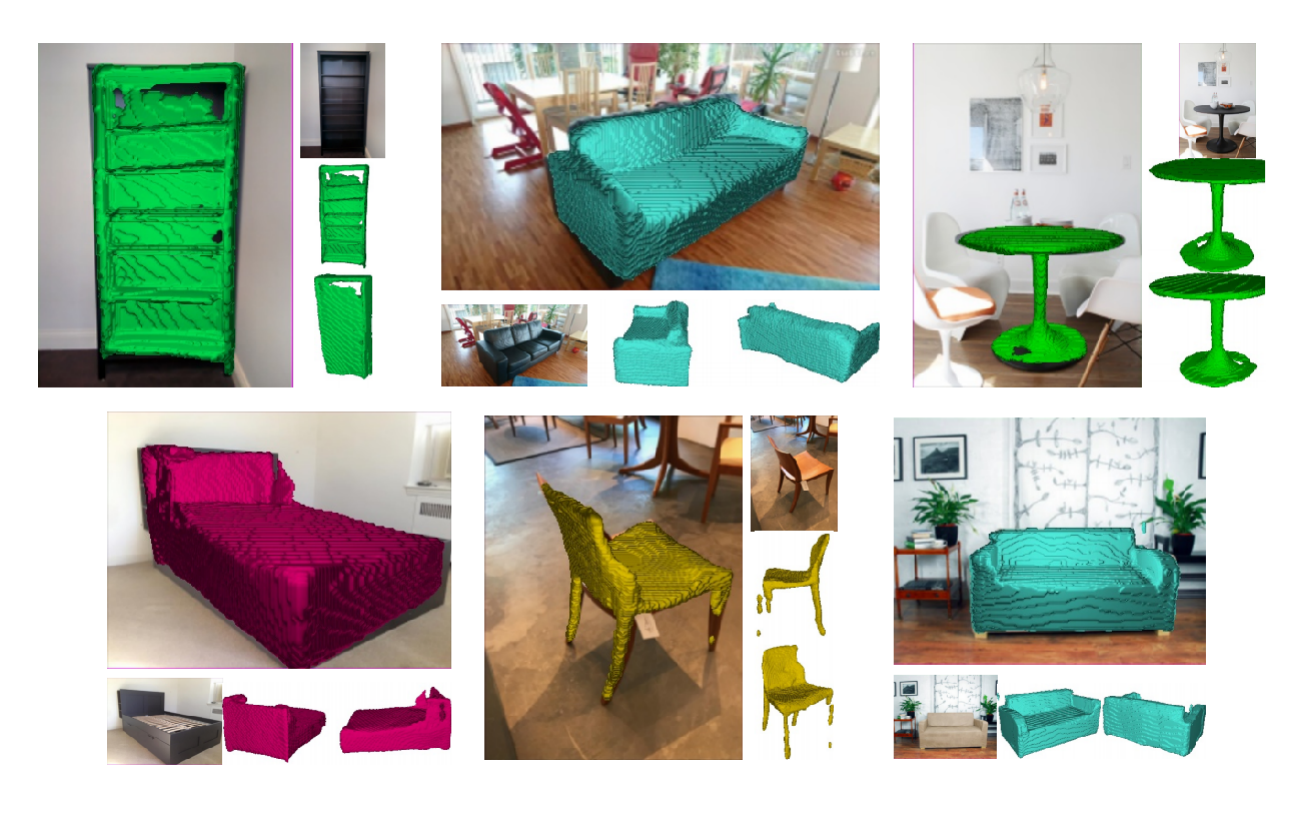}
\caption{Qualitative results of coherent 3D scene reconstruction on Pix3D dataset. Courtesy of \cite{popov2020corenet}}
\label{fig:popov2020corenet}
\end{figure}

In \cite{popov2020corenet}, Popov et al. proposed a coherent 3D scene reconstruction from a single RGB image, using encoder-decoder architectures, along with three extensions:  (1) ray-traced skip connections that propagate local 2D information to the output 3D volume in a physically correct manner; (2) a hybrid 3D volume representation that enables building translation equivariant models, while at the same time encoding fine object details without an excessive memory footprint; (3) a reconstruction loss tailored to capture overall object geometry. 
They reconstruct all objects jointly in one pass, producing a coherent reconstruction, where all objects live in a single consistent 3D coordinate frame relative to the camera and they do not intersect in 3D space.
Some of the sample reconstructed 3D scenes using this framework are shown in  Fig \ref{fig:popov2020corenet}.

\begin{figure}[h]
\centering
\includegraphics[width=0.99\linewidth]{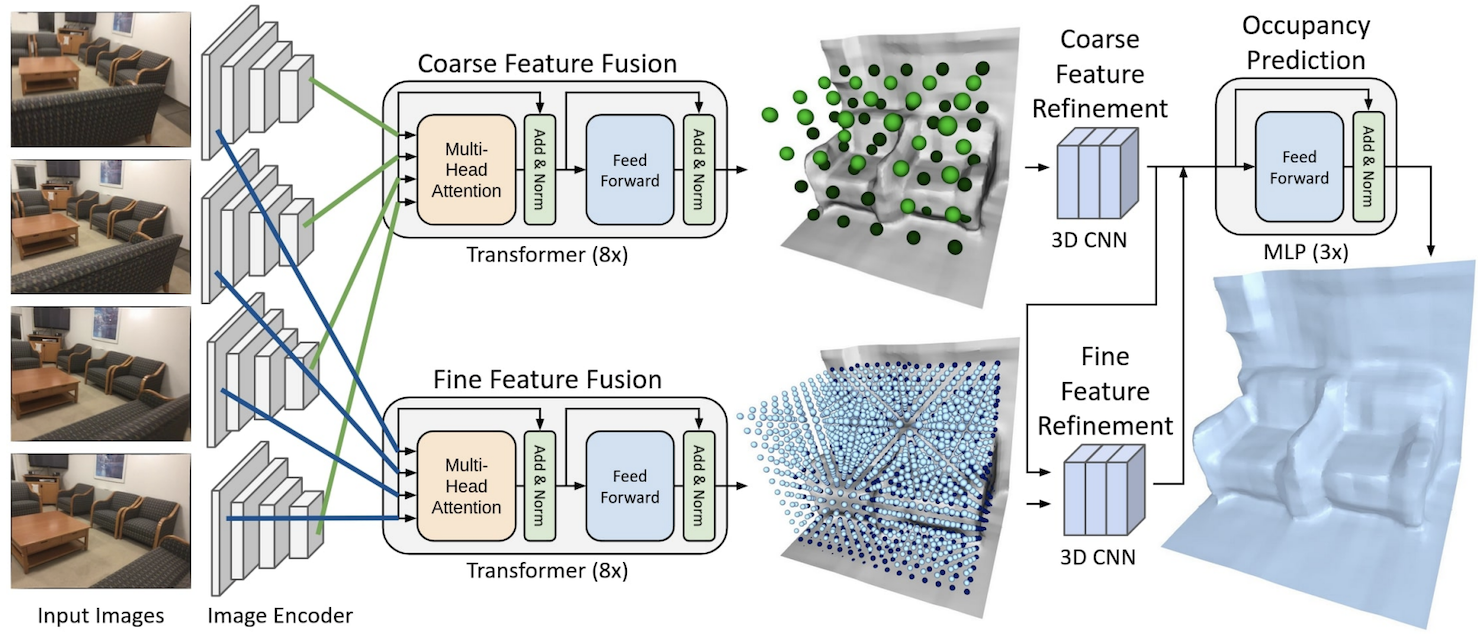}
\caption{The overview of TransformerFusion framework. Courtesy of \cite{bovzivc2021transformerfusion}}
\label{fig:TransformerFusion}
\end{figure}

In \cite{bovzivc2021transformerfusion}, Bovzivc et al. introduced TransformerFusion, a transformer-based 3D scene reconstruction
approach. From an input monocular RGB video, the video frames are processed by a transformer network that fuses the observations into a volumetric feature grid representing the scene; this feature grid is then decoded into an implicit 3D scene representation. Key to their approach is the transformer architecture that enables the network to learn to attend to the most relevant image frames for each 3D location in the scene, supervised only by the scene reconstruction task. Features
are fused in a coarse-to-fine fashion, storing fine-level features only where needed, requiring lower memory storage and enabling fusion at interactive rates. 
The feature grid is then decoded to a higher-resolution scene reconstruction, using an MLP-based surface occupancy prediction from interpolated coarse-to-fine 3D features.
Fig \ref{fig:TransformerFusion} provides the overview of the proposed framework in TransformerFusion.

Some of the other promising works in 3D scene reconstruction includes: CodeSlam \cite{bloesch2018codeslam}, Moulding humans: Non-parametric 3d human shape estimation from single images \cite{gabeur2019moulding}, Atlas: end-to-end 3d scene reconstruction from posed images \cite{murez2020atlas}, From Points to Multi-Object 3D Reconstruction \cite{engelmann2021points}, and VolumeFusion \cite{choe2021volumefusion}.

\section{Future Directions}
\label{sec:future}
Although there has been a huge progress in AR domain in the past few years, several challenges lie ahead. We will next introduce some of the promising research directions that we believe will help in further advancing augmented reality algorithms. 

\subsection{AR in-the-wild}
Many of the current models developed for AR applications work well only under constrained scenarios (such as simple background, or limited occlusion). Developing new models that perform well in general setting and complex environment is an important research area, which can further extend the application of AR models in different areas.
In addition to algorithmic contributions, collecting more complex datasets (with more labeled data in the wild) would be helpful for this purpose.

\subsection{See-through AR}
When (AR) visual effect is overlaid on physical scene, ensuring the realistic feeling for the users/observers is crucial. Even slightest artifact (due to various reasons such as: motion artifacts, quality inconsistency, imperfect segmentation and detection, etc.) could lead to a non-realistic experience for the user. Since human are the main end-user of many of the AR products, subjective tests/metrics could be very useful in assessing these models in early phase, but developing objective metrics to assess how realistic these AR effects/models are (in large-scale sense) is crucial in ensuring good user experience.

\subsection{Realistic 3D Models}
Many of today’s AR models are developed for 2D images, but in order for AR to give people real-world-like feeling, it needs to work well in 3D setting too. Therefore developing AR models for 3D data is crucial (such as realistic human/clothes modeling and manipulation with fine 3D details and textures). There are already some works developed in this direction, but there is still big room for improvement.


\subsection{Security and Privacy in AR setting}
As augmented reality models become more widely used in people's daily life, security and privacy of AR systems are of great importance.
While AR can offer several benefits and new opportunities, making sure users' privacy are taken into account, is very important for these models to become widely trusted.
Hence, developing AR models which have minimal risk of identity thefts and adversarial attacks is imperative.

\subsection{Remote Cooperative AR}
In face-to-face collaboration, people use gesture, gaze and non-verbal cues to communicate in a clear fashion, and in many cases the surrounding environment and objects play a crucial role to providing context and meaning. Physical objects facilitate collaboration both by their appearance, their use as semantic representations, their spatial relationships, and their ability to help focus attention. 
AR system can be used to advance our remote cooperation and collaboration, by taking the surrounding environment of all parties (involved in a discussion or task) into account. However, that  requires more powerful models which can process a lot more information and contexts. 
Co-located AR collaboration can blend the physical and virtual worlds so that real objects can be used to interact with three-dimensional digital content and increase shared understanding.

\subsection{New Sensors for AR (smell, tactile, taste)}
So far, the majority of AR systems are only based on data from visual and depth sensors. But there is no reason for AR to be limited to these sensors, and hopefully that in future there will be  more advanced AR systems which can make use of other types of sensors too, such as smell, taste, tactile (for touch), and beyond.

\subsection{AR Devices in Body}
So far, the main interaction points with AR systems is through cellphones, laptops/PCs, and AR glasses. But developing displays and chips to enable easier interaction with AR systems could be another future direction. It is worth noting that, there are already some works moving in this direction, such as Mojo Lens’ revolutionary design that uses a tiny microLED display to share critical information, and smart sensors (powered by solid-state batteries) built into a Scleral lens that also corrects people’s vision.

\section{Conclusion}
\label{sec:conc}
This work provides a detailed overview of augmented reality, its history, applications, and challenges. It introduces more than twenty applications of AR, and discusses about ten of them in detail. After that, it  provides a survey of some of the recent deep learning based models developed for augmented reality applications, such as for clothing shopping, make-up try on, fitness and workout. Public datasets developed for those tasks are also mentioned in those sections, when available. Given AR’s usefulness, it is continuously applied to new applications . Hence, this paper closes with a discussion about some of the challenges, and possible future directions of AR.

\section*{Acknowledgments}
We would like to thank Iasonas Kokkinos, Qi Pan, Lyric Kaplan, and Liz Markman for reviewing this work, and providing very helpful comments and suggestions.

\bibliographystyle{IEEEtran}
\bibliography{refs}

\begin{thebibliography}{100}
\providecommand{\url}[1]{#1}
\csname url@samestyle\endcsname
\providecommand{\newblock}{\relax}
\providecommand{\bibinfo}[2]{#2}
\providecommand{\BIBentrySTDinterwordspacing}{\spaceskip=0pt\relax}
\providecommand{\BIBentryALTinterwordstretchfactor}{4}
\providecommand{\BIBentryALTinterwordspacing}{\spaceskip=\fontdimen2\font plus
\BIBentryALTinterwordstretchfactor\fontdimen3\font minus
  \fontdimen4\font\relax}
\providecommand{\BIBforeignlanguage}[2]{{%
\expandafter\ifx\csname l@#1\endcsname\relax
\typeout{** WARNING: IEEEtran.bst: No hyphenation pattern has been}%
\typeout{** loaded for the language `#1'. Using the pattern for}%
\typeout{** the default language instead.}%
\else
\language=\csname l@#1\endcsname
\fi
#2}}
\providecommand{\BIBdecl}{\relax}
\BIBdecl

\bibitem{billinghurst2015survey}
M.~Billinghurst, A.~Clark, and G.~Lee, ``A survey of augmented reality,'' 2015.

\bibitem{verge}
\url{https://www.theverge.com/2019/4/4/18294062/snapchat-landmarkers-ar-lenses-filters-eiffel-tower-rainbows}.

\bibitem{4danatomy}
\url{https://www.4danatomy.com/}.

\bibitem{accuvein}
\url{https://www.accuvein.com/}.

\bibitem{loh2020virtual}
C.~L. Loh, ``Virtual fitting room using augmented reality,'' Ph.D.
  dissertation, UTAR, 2020.

\bibitem{borges2019virtual}
A.~d. F.~S. Borges and C.~H. Morimoto, ``A virtual makeup augmented reality
  system,'' in \emph{2019 21st Symposium on Virtual and Augmented Reality
  (SVR)}.\hskip 1em plus 0.5em minus 0.4em\relax IEEE, 2019, pp. 34--42.

\bibitem{nissan}
\url{https://www.nissan-global.com/EN/TECHNOLOGY/OVERVIEW/i2v.html}.

\bibitem{civilization}
\url{https://www.bbc.co.uk/taster/pilots/civilisations-ar}.

\bibitem{hadi2015buy}
M.~Hadi~Kiapour, X.~Han, S.~Lazebnik, A.~C. Berg, and T.~L. Berg, ``Where to
  buy it: Matching street clothing photos in online shops,'' in
  \emph{Proceedings of the IEEE international conference on computer vision},
  2015, pp. 3343--3351.

\bibitem{liu2016deepfashion}
Z.~Liu, P.~Luo, S.~Qiu, X.~Wang, and X.~Tang, ``Deepfashion: Powering robust
  clothes recognition and retrieval with rich annotations,'' in
  \emph{Proceedings of the IEEE conference on computer vision and pattern
  recognition}, 2016, pp. 1096--1104.

\bibitem{dong2017multi}
Q.~Dong, S.~Gong, and X.~Zhu, ``Multi-task curriculum transfer deep learning of
  clothing attributes,'' in \emph{2017 IEEE Winter Conference on Applications
  of Computer Vision (WACV)}.\hskip 1em plus 0.5em minus 0.4em\relax IEEE,
  2017, pp. 520--529.

\bibitem{han2018viton}
X.~Han, Z.~Wu, Z.~Wu, R.~Yu, and L.~S. Davis, ``Viton: An image-based virtual
  try-on network,'' in \emph{Proceedings of the IEEE conference on computer
  vision and pattern recognition}, 2018, pp. 7543--7552.

\bibitem{ge2019deepfashion2}
Y.~Ge, R.~Zhang, X.~Wang, X.~Tang, and P.~Luo, ``Deepfashion2: A versatile
  benchmark for detection, pose estimation, segmentation and re-identification
  of clothing images,'' in \emph{Proceedings of the IEEE/CVF Conference on
  Computer Vision and Pattern Recognition}, 2019, pp. 5337--5345.

\bibitem{han2019clothflow}
X.~Han, X.~Hu, W.~Huang, and M.~R. Scott, ``Clothflow: A flow-based model for
  clothed person generation,'' in \emph{Proceedings of the IEEE/CVF
  International Conference on Computer Vision}, 2019, pp. 10\,471--10\,480.

\bibitem{xie2021wasvton}
Z.~Xie, X.~Zhang, F.~Zhao, H.~Dong, M.~C. Kampffmeyer, H.~Yan, and X.~Liang,
  ``Was-vton: Warping architecture search for virtual try-on network,''
  \emph{arXiv preprint arXiv:2108.00386}, 2021.

\bibitem{guo2020single}
Z.~Guo, X.~Zhang, H.~Mu, W.~Heng, Z.~Liu, Y.~Wei, and J.~Sun, ``Single path
  one-shot neural architecture search with uniform sampling,'' in
  \emph{European Conference on Computer Vision}.\hskip 1em plus 0.5em minus
  0.4em\relax Springer, 2020, pp. 544--560.

\bibitem{neuberger2020image}
A.~Neuberger, E.~Borenstein, B.~Hilleli, E.~Oks, and S.~Alpert, ``Image based
  virtual try-on network from unpaired data,'' in \emph{Proceedings of the
  IEEE/CVF Conference on Computer Vision and Pattern Recognition}, 2020, pp.
  5184--5193.

\bibitem{yang2020towards}
H.~Yang, R.~Zhang, X.~Guo, W.~Liu, W.~Zuo, and P.~Luo, ``Towards
  photo-realistic virtual try-on by adaptively generating-preserving image
  content,'' in \emph{Proceedings of the IEEE/CVF Conference on Computer Vision
  and Pattern Recognition}, 2020, pp. 7850--7859.

\bibitem{li2021toward}
K.~Li, M.~J. Chong, J.~Zhang, and J.~Liu, ``Toward accurate and realistic
  outfits visualization with attention to details,'' in \emph{Proceedings of
  the IEEE/CVF Conference on Computer Vision and Pattern Recognition}, 2021,
  pp. 15\,546--15\,555.

\bibitem{Zhao2021M3DVTONAM}
F.~Zhao, Z.~Xie, M.~C. Kampffmeyer, H.~Dong, S.~Han, T.~Zheng, T.~Zhang, and
  X.~Liang, ``M3d-vton: A monocular-to-3d virtual try-on network,''
  \emph{ArXiv}, vol. abs/2108.05126, 2021.

\bibitem{raj2018swapnet}
A.~Raj, P.~Sangkloy, H.~Chang, J.~Hays, D.~Ceylan, and J.~Lu, ``Swapnet: Image
  based garment transfer,'' in \emph{European Conference on Computer
  Vision}.\hskip 1em plus 0.5em minus 0.4em\relax Springer, 2018, pp. 679--695.

\bibitem{santesteban2019learning}
I.~Santesteban, M.~A. Otaduy, and D.~Casas, ``Learning-based animation of
  clothing for virtual try-on,'' in \emph{Computer Graphics Forum}, vol.~38,
  no.~2.\hskip 1em plus 0.5em minus 0.4em\relax Wiley Online Library, 2019, pp.
  355--366.

\bibitem{gundogdu2019garnet}
E.~Gundogdu, V.~Constantin, A.~Seifoddini, M.~Dang, M.~Salzmann, and P.~Fua,
  ``Garnet: A two-stream network for fast and accurate 3d cloth draping,'' in
  \emph{Proceedings of the IEEE/CVF International Conference on Computer
  Vision}, 2019, pp. 8739--8748.

\bibitem{lazova2019360}
V.~Lazova, E.~Insafutdinov, and G.~Pons-Moll, ``360-degree textures of people
  in clothing from a single image,'' in \emph{2019 International Conference on
  3D Vision (3DV)}.\hskip 1em plus 0.5em minus 0.4em\relax IEEE, 2019, pp.
  643--653.

\bibitem{wu2019m2e}
Z.~Wu, G.~Lin, Q.~Tao, and J.~Cai, ``M2e-try on net: Fashion from model to
  everyone,'' in \emph{Proceedings of the 27th ACM International Conference on
  Multimedia}, 2019, pp. 293--301.

\bibitem{dong2019fw}
H.~Dong, X.~Liang, X.~Shen, B.~Wu, B.-C. Chen, and J.~Yin, ``Fw-gan:
  Flow-navigated warping gan for video virtual try-on,'' in \emph{Proceedings
  of the IEEE/CVF International Conference on Computer Vision}, 2019, pp.
  1161--1170.

\bibitem{jae2019viton}
H.~Jae~Lee, R.~Lee, M.~Kang, M.~Cho, and G.~Park, ``La-viton: a network for
  looking-attractive virtual try-on,'' in \emph{Proceedings of the IEEE/CVF
  International Conference on Computer Vision Workshops}, 2019, pp. 0--0.

\bibitem{hsiao2019fashion}
W.-L. Hsiao, I.~Katsman, C.-Y. Wu, D.~Parikh, and K.~Grauman, ``Fashion++:
  Minimal edits for outfit improvement,'' in \emph{Proceedings of the IEEE/CVF
  International Conference on Computer Vision}, 2019, pp. 5047--5056.

\bibitem{patel2020tailornet}
C.~Patel, Z.~Liao, and G.~Pons-Moll, ``Tailornet: Predicting clothing in 3d as
  a function of human pose, shape and garment style,'' in \emph{Proceedings of
  the IEEE/CVF Conference on Computer Vision and Pattern Recognition}, 2020,
  pp. 7365--7375.

\bibitem{hsiao2020vibe}
W.-L. Hsiao and K.~Grauman, ``Vibe: Dressing for diverse body shapes,'' in
  \emph{Proceedings of the IEEE/CVF Conference on Computer Vision and Pattern
  Recognition}, 2020, pp. 11\,059--11\,069.

\bibitem{ren2021cloth}
B.~Ren, H.~Tang, F.~Meng, R.~Ding, L.~Shao, P.~H. Torr, and N.~Sebe, ``Cloth
  interactive transformer for virtual try-on,'' \emph{arXiv preprint
  arXiv:2104.05519}, 2021.

\bibitem{choi2021viton}
S.~Choi, S.~Park, M.~Lee, and J.~Choo, ``Viton-hd: High-resolution virtual
  try-on via misalignment-aware normalization,'' in \emph{Proceedings of the
  IEEE/CVF Conference on Computer Vision and Pattern Recognition}, 2021, pp.
  14\,131--14\,140.

\bibitem{ge2021parser}
Y.~Ge, Y.~Song, R.~Zhang, C.~Ge, W.~Liu, and P.~Luo, ``Parser-free virtual
  try-on via distilling appearance flows,'' in \emph{Proceedings of the
  IEEE/CVF Conference on Computer Vision and Pattern Recognition}, 2021, pp.
  8485--8493.

\bibitem{yang2021ct}
F.~Yang and G.~Lin, ``Ct-net: Complementary transfering network for garment
  transfer with arbitrary geometric changes,'' in \emph{Proceedings of the
  IEEE/CVF Conference on Computer Vision and Pattern Recognition}, 2021, pp.
  9899--9908.

\bibitem{liu2016makeup}
S.~Liu, X.~Ou, R.~Qian, W.~Wang, and X.~Cao, ``Makeup like a superstar: Deep
  localized makeup transfer network,'' \emph{arXiv preprint arXiv:1604.07102},
  2016.

\bibitem{alashkar2017examples}
T.~Alashkar, S.~Jiang, S.~Wang, and Y.~Fu, ``Examples-rules guided deep neural
  network for makeup recommendation,'' in \emph{Proceedings of the AAAI
  Conference on Artificial Intelligence}, vol.~31, no.~1, 2017.

\bibitem{li2018beautygan}
T.~Li, R.~Qian, C.~Dong, S.~Liu, Q.~Yan, W.~Zhu, and L.~Lin, ``Beautygan:
  Instance-level facial makeup transfer with deep generative adversarial
  network,'' in \emph{Proceedings of the 26th ACM international conference on
  Multimedia}, 2018, pp. 645--653.

\bibitem{chang2018pairedcyclegan}
H.~Chang, J.~Lu, F.~Yu, and A.~Finkelstein, ``Pairedcyclegan: Asymmetric style
  transfer for applying and removing makeup,'' in \emph{Proceedings of the IEEE
  conference on computer vision and pattern recognition}, 2018, pp. 40--48.

\bibitem{gu2019ladn}
Q.~Gu, G.~Wang, M.~T. Chiu, Y.-W. Tai, and C.-K. Tang, ``Ladn: Local
  adversarial disentangling network for facial makeup and de-makeup,'' in
  \emph{Proceedings of the IEEE/CVF International Conference on Computer
  Vision}, 2019, pp. 10\,481--10\,490.

\bibitem{jiang2020psgan}
W.~Jiang, S.~Liu, C.~Gao, J.~Cao, R.~He, J.~Feng, and S.~Yan, ``Psgan: Pose and
  expression robust spatial-aware gan for customizable makeup transfer,'' in
  \emph{Proceedings of the IEEE/CVF Conference on Computer Vision and Pattern
  Recognition}, 2020, pp. 5194--5202.

\bibitem{nguyen2021lipstick}
T.~Nguyen, A.~T. Tran, and M.~Hoai, ``Lipstick ain't enough: Beyond color
  matching for in-the-wild makeup transfer,'' in \emph{Proceedings of the
  IEEE/CVF Conference on Computer Vision and Pattern Recognition}, 2021, pp.
  13\,305--13\,314.

\bibitem{cao2019makeup}
C.~Cao, F.~Lu, C.~Li, S.~Lin, and X.~Shen, ``Makeup removal via bidirectional
  tunable de-makeup network,'' \emph{IEEE Transactions on Multimedia}, vol.~21,
  no.~11, pp. 2750--2761, 2019.

\bibitem{liu2019face}
X.~Liu, R.~Wang, C.-F. Chen, M.~Yin, H.~Peng, S.~Ng, and X.~Li, ``Face
  beautification: Beyond makeup transfer,'' \emph{arXiv preprint
  arXiv:1912.03630}, 2019.

\bibitem{chen2019beautyglow}
H.-J. Chen, K.-M. Hui, S.-Y. Wang, L.-W. Tsao, H.-H. Shuai, and W.-H. Cheng,
  ``Beautyglow: On-demand makeup transfer framework with reversible generative
  network,'' in \emph{Proceedings of the IEEE/CVF Conference on Computer Vision
  and Pattern Recognition}, 2019, pp. 10\,042--10\,050.

\bibitem{velusamy2020fabsoften}
S.~Velusamy, R.~Parihar, R.~Kini, and A.~Rege, ``Fabsoften: Face beautification
  via dynamic skin smoothing, guided feathering, and texture restoration,'' in
  \emph{Proceedings of the IEEE/CVF Conference on Computer Vision and Pattern
  Recognition Workshops}, 2020, pp. 530--531.

\bibitem{kips2020gan}
R.~Kips, P.~Gori, M.~Perrot, and I.~Bloch, ``Ca-gan: Weakly supervised color
  aware gan for controllable makeup transfer,'' in \emph{European Conference on
  Computer Vision}.\hskip 1em plus 0.5em minus 0.4em\relax Springer, 2020, pp.
  280--296.

\bibitem{zhu2017unpaired}
J.-Y. Zhu, T.~Park, P.~Isola, and A.~A. Efros, ``Unpaired image-to-image
  translation using cycle-consistent adversarial networks,'' in
  \emph{Proceedings of the IEEE international conference on computer vision},
  2017, pp. 2223--2232.

\bibitem{yi2017dualgan}
Z.~Yi, H.~Zhang, P.~Tan, and M.~Gong, ``Dualgan: Unsupervised dual learning for
  image-to-image translation,'' in \emph{Proceedings of the IEEE international
  conference on computer vision}, 2017, pp. 2849--2857.

\bibitem{choi2018stargan}
Y.~Choi, M.~Choi, M.~Kim, J.-W. Ha, S.~Kim, and J.~Choo, ``Stargan: Unified
  generative adversarial networks for multi-domain image-to-image
  translation,'' in \emph{Proceedings of the IEEE conference on computer vision
  and pattern recognition}, 2018, pp. 8789--8797.

\bibitem{huang2018multimodal}
X.~Huang, M.-Y. Liu, S.~Belongie, and J.~Kautz, ``Multimodal unsupervised
  image-to-image translation,'' in \emph{Proceedings of the European conference
  on computer vision (ECCV)}, 2018, pp. 172--189.

\bibitem{karras2019style}
T.~Karras, S.~Laine, and T.~Aila, ``A style-based generator architecture for
  generative adversarial networks,'' in \emph{Proceedings of the IEEE/CVF
  Conference on Computer Vision and Pattern Recognition}, 2019, pp. 4401--4410.

\bibitem{he2019attgan}
Z.~He, W.~Zuo, M.~Kan, S.~Shan, and X.~Chen, ``Attgan: Facial attribute editing
  by only changing what you want,'' \emph{IEEE transactions on image
  processing}, vol.~28, no.~11, pp. 5464--5478, 2019.

\bibitem{choi2020stargan}
Y.~Choi, Y.~Uh, J.~Yoo, and J.-W. Ha, ``Stargan v2: Diverse image synthesis for
  multiple domains,'' in \emph{Proceedings of the IEEE/CVF Conference on
  Computer Vision and Pattern Recognition}, 2020, pp. 8188--8197.

\bibitem{karras2020analyzing}
T.~Karras, S.~Laine, M.~Aittala, J.~Hellsten, J.~Lehtinen, and T.~Aila,
  ``Analyzing and improving the image quality of stylegan,'' in
  \emph{Proceedings of the IEEE/CVF Conference on Computer Vision and Pattern
  Recognition}, 2020, pp. 8110--8119.

\bibitem{wu2021stylespace}
Z.~Wu, D.~Lischinski, and E.~Shechtman, ``Stylespace analysis: Disentangled
  controls for stylegan image generation,'' in \emph{Proceedings of the
  IEEE/CVF Conference on Computer Vision and Pattern Recognition}, 2021, pp.
  12\,863--12\,872.

\bibitem{karras2021alias}
T.~Karras, M.~Aittala, S.~Laine, E.~H{\"a}rk{\"o}nen, J.~Hellsten, J.~Lehtinen,
  and T.~Aila, ``Alias-free generative adversarial networks,'' \emph{arXiv
  preprint arXiv:2106.12423}, 2021.

\bibitem{krafka2016eye}
K.~Krafka, A.~Khosla, P.~Kellnhofer, H.~Kannan, S.~Bhandarkar, W.~Matusik, and
  A.~Torralba, ``Eye tracking for everyone,'' in \emph{Proceedings of the IEEE
  conference on computer vision and pattern recognition}, 2016, pp. 2176--2184.

\bibitem{zhang2017s}
X.~Zhang, Y.~Sugano, M.~Fritz, and A.~Bulling, ``It's written all over your
  face: Full-face appearance-based gaze estimation,'' in \emph{Proceedings of
  the IEEE Conference on Computer Vision and Pattern Recognition Workshops},
  2017, pp. 51--60.

\bibitem{fischer2018rt}
T.~Fischer, H.~J. Chang, and Y.~Demiris, ``Rt-gene: Real-time eye gaze
  estimation in natural environments,'' in \emph{Proceedings of the European
  Conference on Computer Vision (ECCV)}, 2018, pp. 334--352.

\bibitem{kellnhofer2019gaze360}
P.~Kellnhofer, A.~Recasens, S.~Stent, W.~Matusik, and A.~Torralba, ``Gaze360:
  Physically unconstrained gaze estimation in the wild,'' in \emph{Proceedings
  of the IEEE/CVF International Conference on Computer Vision}, 2019, pp.
  6912--6921.

\bibitem{yu2020unsupervised}
Y.~Yu and J.-M. Odobez, ``Unsupervised representation learning for gaze
  estimation,'' in \emph{Proceedings of the IEEE/CVF Conference on Computer
  Vision and Pattern Recognition}, 2020, pp. 7314--7324.

\bibitem{fang2021dual}
Y.~Fang, J.~Tang, W.~Shen, W.~Shen, X.~Gu, L.~Song, and G.~Zhai, ``Dual
  attention guided gaze target detection in the wild,'' in \emph{Proceedings of
  the IEEE/CVF Conference on Computer Vision and Pattern Recognition}, 2021,
  pp. 11\,390--11\,399.

\bibitem{park2019few}
S.~Park, S.~D. Mello, P.~Molchanov, U.~Iqbal, O.~Hilliges, and J.~Kautz,
  ``Few-shot adaptive gaze estimation,'' in \emph{Proceedings of the IEEE/CVF
  International Conference on Computer Vision}, 2019, pp. 9368--9377.

\bibitem{zhang2020eth}
X.~Zhang, S.~Park, T.~Beeler, D.~Bradley, S.~Tang, and O.~Hilliges,
  ``Eth-xgaze: A large scale dataset for gaze estimation under extreme head
  pose and gaze variation,'' in \emph{European Conference on Computer
  Vision}.\hskip 1em plus 0.5em minus 0.4em\relax Springer, 2020, pp. 365--381.

\bibitem{park2020towards}
S.~Park, E.~Aksan, X.~Zhang, and O.~Hilliges, ``Towards end-to-end video-based
  eye-tracking,'' in \emph{European Conference on Computer Vision}.\hskip 1em
  plus 0.5em minus 0.4em\relax Springer, 2020, pp. 747--763.

\bibitem{kothari2021weakly}
R.~Kothari, S.~De~Mello, U.~Iqbal, W.~Byeon, S.~Park, and J.~Kautz,
  ``Weakly-supervised physically unconstrained gaze estimation,'' in
  \emph{Proceedings of the IEEE/CVF Conference on Computer Vision and Pattern
  Recognition}, 2021, pp. 9980--9989.

\bibitem{oberweger2015hands}
M.~Oberweger, P.~Wohlhart, and V.~Lepetit, ``Hands deep in deep learning for
  hand pose estimation,'' \emph{arXiv preprint arXiv:1502.06807}, 2015.

\bibitem{zhou2016model}
X.~Zhou, Q.~Wan, W.~Zhang, X.~Xue, and Y.~Wei, ``Model-based deep hand pose
  estimation,'' \emph{arXiv preprint arXiv:1606.06854}, 2016.

\bibitem{ge20173d}
L.~Ge, H.~Liang, J.~Yuan, and D.~Thalmann, ``3d convolutional neural networks
  for efficient and robust hand pose estimation from single depth images,'' in
  \emph{Proceedings of the IEEE Conference on Computer Vision and Pattern
  Recognition}, 2017, pp. 1991--2000.

\bibitem{spurr2018cross}
A.~Spurr, J.~Song, S.~Park, and O.~Hilliges, ``Cross-modal deep variational
  hand pose estimation,'' in \emph{Proceedings of the IEEE Conference on
  Computer Vision and Pattern Recognition}, 2018, pp. 89--98.

\bibitem{chen2019so}
Y.~Chen, Z.~Tu, L.~Ge, D.~Zhang, R.~Chen, and J.~Yuan, ``So-handnet:
  Self-organizing network for 3d hand pose estimation with semi-supervised
  learning,'' in \emph{Proceedings of the IEEE/CVF International Conference on
  Computer Vision}, 2019, pp. 6961--6970.

\bibitem{moon2020interhand2}
G.~Moon, S.-I. Yu, H.~Wen, T.~Shiratori, and K.~M. Lee, ``Interhand2. 6m: A
  dataset and baseline for 3d interacting hand pose estimation from a single
  rgb image,'' in \emph{Computer Vision--ECCV 2020: 16th European Conference,
  Glasgow, UK, August 23--28, 2020, Proceedings, Part XX 16}.\hskip 1em plus
  0.5em minus 0.4em\relax Springer, 2020, pp. 548--564.

\bibitem{caramalau2021active}
R.~Caramalau, B.~Bhattarai, and T.-K. Kim, ``Active learning for bayesian 3d
  hand pose estimation,'' in \emph{Proceedings of the IEEE/CVF Winter
  Conference on Applications of Computer Vision}, 2021, pp. 3419--3428.

\bibitem{ye2016spatial}
Q.~Ye, S.~Yuan, and T.-K. Kim, ``Spatial attention deep net with partial pso
  for hierarchical hybrid hand pose estimation,'' in \emph{European conference
  on computer vision}.\hskip 1em plus 0.5em minus 0.4em\relax Springer, 2016,
  pp. 346--361.

\bibitem{oberweger2017deepprior}
M.~Oberweger and V.~Lepetit, ``Deepprior++: Improving fast and accurate 3d hand
  pose estimation,'' in \emph{Proceedings of the IEEE international conference
  on computer vision Workshops}, 2017, pp. 585--594.

\bibitem{ge2018point}
L.~Ge, Z.~Ren, and J.~Yuan, ``Point-to-point regression pointnet for 3d hand
  pose estimation,'' in \emph{Proceedings of the European conference on
  computer vision (ECCV)}, 2018, pp. 475--491.

\bibitem{huang2020hand}
L.~Huang, J.~Tan, J.~Liu, and J.~Yuan, ``Hand-transformer: non-autoregressive
  structured modeling for 3d hand pose estimation,'' in \emph{European
  Conference on Computer Vision}.\hskip 1em plus 0.5em minus 0.4em\relax
  Springer, 2020, pp. 17--33.

\bibitem{stergioulas20213d}
A.~Stergioulas, T.~Chatzis, D.~Konstantinidis, K.~Dimitropoulos, and P.~Daras,
  ``3d hand pose estimation via aligned latent space injection and kinematic
  losses,'' in \emph{Proceedings of the IEEE/CVF Conference on Computer Vision
  and Pattern Recognition}, 2021, pp. 1730--1739.

\bibitem{wei2016convolutional}
S.-E. Wei, V.~Ramakrishna, T.~Kanade, and Y.~Sheikh, ``Convolutional pose
  machines,'' in \emph{Proceedings of the IEEE conference on Computer Vision
  and Pattern Recognition}, 2016, pp. 4724--4732.

\bibitem{cao2017realtime}
Z.~Cao, T.~Simon, S.-E. Wei, and Y.~Sheikh, ``Realtime multi-person 2d pose
  estimation using part affinity fields,'' in \emph{Proceedings of the IEEE
  conference on computer vision and pattern recognition}, 2017, pp. 7291--7299.

\bibitem{guler2018densepose}
R.~A. G{\"u}ler, N.~Neverova, and I.~Kokkinos, ``Densepose: Dense human pose
  estimation in the wild,'' in \emph{Proceedings of the IEEE conference on
  computer vision and pattern recognition}, 2018, pp. 7297--7306.

\bibitem{cao2019openpose}
Z.~Cao, G.~Hidalgo, T.~Simon, S.-E. Wei, and Y.~Sheikh, ``Openpose: realtime
  multi-person 2d pose estimation using part affinity fields,'' \emph{IEEE
  transactions on pattern analysis and machine intelligence}, vol.~43, no.~1,
  pp. 172--186, 2019.

\bibitem{xu2020ghum}
H.~Xu, E.~G. Bazavan, A.~Zanfir, W.~T. Freeman, R.~Sukthankar, and
  C.~Sminchisescu, ``Ghum \& ghuml: Generative 3d human shape and articulated
  pose models,'' in \emph{Proceedings of the IEEE/CVF Conference on Computer
  Vision and Pattern Recognition}, 2020, pp. 6184--6193.

\bibitem{liu2020keypose}
X.~Liu, R.~Jonschkowski, A.~Angelova, and K.~Konolige, ``Keypose: Multi-view 3d
  labeling and keypoint estimation for transparent objects,'' in
  \emph{Proceedings of the IEEE/CVF conference on computer vision and pattern
  recognition}, 2020, pp. 11\,602--11\,610.

\bibitem{he2021ffb6d}
Y.~He, H.~Huang, H.~Fan, Q.~Chen, and J.~Sun, ``Ffb6d: A full flow
  bidirectional fusion network for 6d pose estimation,'' in \emph{Proceedings
  of the IEEE/CVF Conference on Computer Vision and Pattern Recognition}, 2021,
  pp. 3003--3013.

\bibitem{fang2017rmpe}
H.-S. Fang, S.~Xie, Y.-W. Tai, and C.~Lu, ``Rmpe: Regional multi-person pose
  estimation,'' in \emph{Proceedings of the IEEE international conference on
  computer vision}, 2017, pp. 2334--2343.

\bibitem{xiao2018simple}
B.~Xiao, H.~Wu, and Y.~Wei, ``Simple baselines for human pose estimation and
  tracking,'' in \emph{Proceedings of the European conference on computer
  vision (ECCV)}, 2018, pp. 466--481.

\bibitem{yuan2021simpoe}
Y.~Yuan, S.-E. Wei, T.~Simon, K.~Kitani, and J.~Saragih, ``Simpoe: Simulated
  character control for 3d human pose estimation,'' in \emph{Proceedings of the
  IEEE/CVF Conference on Computer Vision and Pattern Recognition}, 2021, pp.
  7159--7169.

\bibitem{ummenhofer2017demon}
B.~Ummenhofer, H.~Zhou, J.~Uhrig, N.~Mayer, E.~Ilg, A.~Dosovitskiy, and
  T.~Brox, ``Demon: Depth and motion network for learning monocular stereo,''
  in \emph{Proceedings of the IEEE Conference on Computer Vision and Pattern
  Recognition}, 2017, pp. 5038--5047.

\bibitem{yin2018geonet}
Z.~Yin and J.~Shi, ``Geonet: Unsupervised learning of dense depth, optical flow
  and camera pose,'' in \emph{Proceedings of the IEEE conference on computer
  vision and pattern recognition}, 2018, pp. 1983--1992.

\bibitem{gordon2019depth}
A.~Gordon, H.~Li, R.~Jonschkowski, and A.~Angelova, ``Depth from videos in the
  wild: Unsupervised monocular depth learning from unknown cameras,'' in
  \emph{Proceedings of the IEEE/CVF International Conference on Computer
  Vision}, 2019, pp. 8977--8986.

\bibitem{guizilini20203d}
V.~Guizilini, R.~Ambrus, S.~Pillai, A.~Raventos, and A.~Gaidon, ``3d packing
  for self-supervised monocular depth estimation,'' in \emph{Proceedings of the
  IEEE/CVF Conference on Computer Vision and Pattern Recognition}, 2020, pp.
  2485--2494.

\bibitem{ranftl2021vision}
R.~Ranftl, A.~Bochkovskiy, and V.~Koltun, ``Vision transformers for dense
  prediction,'' in \emph{Proceedings of the IEEE/CVF International Conference
  on Computer Vision}, 2021, pp. 12\,179--12\,188.

\bibitem{zhou2017unsupervised}
T.~Zhou, M.~Brown, N.~Snavely, and D.~G. Lowe, ``Unsupervised learning of depth
  and ego-motion from video,'' in \emph{Proceedings of the IEEE conference on
  computer vision and pattern recognition}, 2017, pp. 1851--1858.

\bibitem{li2018megadepth}
Z.~Li and N.~Snavely, ``Megadepth: Learning single-view depth prediction from
  internet photos,'' in \emph{Proceedings of the IEEE Conference on Computer
  Vision and Pattern Recognition}, 2018, pp. 2041--2050.

\bibitem{bozic2021transformerfusion}
A.~Bozic, P.~Palafox, J.~Thies, A.~Dai, and M.~Nie{\ss}ner,
  ``Transformerfusion: Monocular rgb scene reconstruction using transformers,''
  \emph{Advances in Neural Information Processing Systems}, vol.~34, 2021.

\bibitem{dai2017scannet}
A.~Dai, A.~X. Chang, M.~Savva, M.~Halber, T.~Funkhouser, and M.~Nie{\ss}ner,
  ``Scannet: Richly-annotated 3d reconstructions of indoor scenes,'' in
  \emph{Proceedings of the IEEE conference on computer vision and pattern
  recognition}, 2017, pp. 5828--5839.

\bibitem{zhang2017mixedfusion}
H.~Zhang and F.~Xu, ``Mixedfusion: Real-time reconstruction of an indoor scene
  with dynamic objects,'' \emph{IEEE transactions on visualization and computer
  graphics}, vol.~24, no.~12, pp. 3137--3146, 2017.

\bibitem{huang2018holistic}
S.~Huang, S.~Qi, Y.~Zhu, Y.~Xiao, Y.~Xu, and S.-C. Zhu, ``Holistic 3d scene
  parsing and reconstruction from a single rgb image,'' in \emph{Proceedings of
  the European conference on computer vision (ECCV)}, 2018, pp. 187--203.

\bibitem{shin20193d}
D.~Shin, Z.~Ren, E.~B. Sudderth, and C.~C. Fowlkes, ``3d scene reconstruction
  with multi-layer depth and epipolar transformers,'' in \emph{Proceedings of
  the IEEE/CVF International Conference on Computer Vision}, 2019, pp.
  2172--2182.

\bibitem{popov2020corenet}
S.~Popov, P.~Bauszat, and V.~Ferrari, ``Corenet: Coherent 3d scene
  reconstruction from a single rgb image,'' in \emph{European Conference on
  Computer Vision}.\hskip 1em plus 0.5em minus 0.4em\relax Springer, 2020, pp.
  366--383.

\bibitem{bovzivc2021transformerfusion}
A.~Bo{\v{z}}i{\v{c}}, P.~Palafox, J.~Thies, A.~Dai, and M.~Nie{\ss}ner,
  ``Transformerfusion: Monocular rgb scene reconstruction using transformers,''
  \emph{arXiv preprint arXiv:2107.02191}, 2021.

\bibitem{bloesch2018codeslam}
M.~Bloesch, J.~Czarnowski, R.~Clark, S.~Leutenegger, and A.~J. Davison,
  ``Codeslam—learning a compact, optimisable representation for dense visual
  slam,'' in \emph{Proceedings of the IEEE conference on computer vision and
  pattern recognition}, 2018, pp. 2560--2568.

\bibitem{gabeur2019moulding}
V.~Gabeur, J.-S. Franco, X.~Martin, C.~Schmid, and G.~Rogez, ``Moulding humans:
  Non-parametric 3d human shape estimation from single images,'' in
  \emph{Proceedings of the IEEE/CVF International Conference on Computer
  Vision}, 2019, pp. 2232--2241.

\bibitem{murez2020atlas}
Z.~Murez, T.~van As, J.~Bartolozzi, A.~Sinha, V.~Badrinarayanan, and
  A.~Rabinovich, ``Atlas: End-to-end 3d scene reconstruction from posed
  images,'' in \emph{Computer Vision--ECCV 2020: 16th European Conference,
  Glasgow, UK, August 23--28, 2020, Proceedings, Part VII 16}.\hskip 1em plus
  0.5em minus 0.4em\relax Springer, 2020, pp. 414--431.

\bibitem{engelmann2021points}
F.~Engelmann, K.~Rematas, B.~Leibe, and V.~Ferrari, ``From points to
  multi-object 3d reconstruction,'' in \emph{Proceedings of the IEEE/CVF
  Conference on Computer Vision and Pattern Recognition}, 2021, pp. 4588--4597.

\bibitem{choe2021volumefusion}
J.~Choe, S.~Im, F.~Rameau, M.~Kang, and I.~S. Kweon, ``Volumefusion: Deep depth
  fusion for 3d scene reconstruction,'' in \emph{Proceedings of the IEEE/CVF
  International Conference on Computer Vision}, 2021, pp. 16\,086--16\,095.

\end{thebibliography}

\textbf{Legal Disclaimer:}
This paper provides an overview of prominent machine learning models developed for augmented reality. It is not representative of Snap Inc’s practices, processes, techniques, and perspectives.

\end{document}